\documentclass[11pt,a4paper]{article}
\usepackage{a4wide}
\usepackage{graphicx}
\usepackage[fleqn]{amsmath}
\usepackage{amssymb}
\usepackage{amsthm}
\usepackage{bm}
\usepackage{url}
\usepackage{color}
\newcommand{\bbbr}{\mathbb{R}}
\theoremstyle{plain}
\newtheorem{definition}{Definition}
\newtheorem{proposition}{Proposition}
\newtheorem{theorem}{Theorem}
\allowdisplaybreaks
\begin{document}
\title{Amoeba Techniques for Shape and Texture Analysis}
\author{Martin Welk\\ University for Health Sciences, Medical Informatics and Technology (UMIT),\\ Eduard-Walln\"ofer-Zentrum 1, 6060 Hall/Tyrol, Austria\\ \url{martin.welk@umit.at} }
\date{June 10, 2015}
\maketitle

\begin{abstract}
Morphological amoebas are image-adaptive structuring
elements for morphological and other local image filters introduced
by Lerallut et al. Their construction is based on combining spatial
distance with contrast information into an image-dependent metric.
Amoeba filters show interesting parallels to image filtering
methods based on partial differential equations (PDEs), which can be
confirmed by asymptotic equivalence results. In computing amoebas,
graph structures are generated that hold information about local
image texture.
This paper reviews and summarises the work of the author and his
coauthors on morphological amoebas, particularly their relations to
PDE filters and texture analysis. It presents some extensions and
points out directions for future investigation on the subject.

\bigskip \noindent%
 \end{abstract}

\section{Introduction}
\label{sec:intro}

Mathematical morphology \cite{Matheron-Book67,Serra-Book82,Serra-Book88}
has developed since the 1960s as a powerful theoretical framework from
which versatile instruments for shape analysis in images can be derived,
such as for structure-preserving denoising or shape simplification
\cite{Heijmans-Book94}.
The fundamental building blocks of classical mathematical morphology are
non-linear local image filters like dilation, erosion, and median filters.
They rely on aggregating intensities within a neighbourhood of any given
pixel by e.g.\ maximum, minimum, and median operations. The selection of
neighbourhoods for processing is classically done by shifting a sliding
window of fixed size and shape across the image. In the context of morphology,
this sliding window is known as structuring element.

More recently, concepts for adaptivity have been developed generally in
image filtering and also specifically in morphology
\cite{Verly-TIP93,BragaNeto-ismm96}. One recent concept for adaptive
morphology are morphological amoebas introduced by Lerallut et al.\
\cite{Lerallut-ismm05}. These are space-variant structuring elements
constructed from a combination of spatial distance measurement with
local contrast measurement via an amoeba metric.

In earlier work by the author of the present paper and his coauthors,
properties of amoeba filters and their relations to image filters based
on partial differential equations (PDEs) were investigated
\cite{Welk-ssvm13,Welk-Aiep14,Welk-JMIV11}. As an application to image
segmentation, an amoeba-based active contour method was designed
\cite{Welk-ssvm11,Welk-ssvm13,Welk-JMIV15}. Recently, a combination of
edge-weighted graphs generated in the computation of amoebas with graph
indices was used to introduce a new class of texture descriptors
\cite{Welk-qgt14} which are currently under further investigation.
This paper reviews and summarises the results from these works.
Directions of ongoing research on this topic are sketched.

With focus on giving a comprehensive overview of the theory that has
been developed in various earlier publications, the (mostly lengthy) proofs
of the results are omitted here and referred to the respective original
sources. Nevertheless, the main principles underlying the proofs are shortly
outlined.
Although amoeba filtering of multi-channel images has been addressed to some
extent in \cite{Welk-Aiep14}, this aspect of the topic presents itself in
a stage too early for a summarised presentation, and is therefore not
included in the present paper.

In the following the structure of the paper is detailed,
highlighting contributions that are novel in this presentation.

Section~\ref{sec:amoebas} introduces the concept of morphological amoebas
as image-adaptive structuring elements in the space-discrete as well as the
space-continuous setting. To ease bridging to the graph techniques discussed
later in Section~\ref{sec:texture}, the presentation in the discrete case
emphasises the modelling of discrete images by neighbourhood graphs and
uses standard terminology from graph theory, thereby following
\cite{Welk-qgt14}. The presentation of the space-continuous case is
similar to that e.g.\ in \cite{Welk-Aiep14}.

The application of amoebas in image filtering is the topic of
Section~\ref{sec:filters}. Median filters, morphological dilation and erosion
are presented together with their relationship to PDE image filters,
reproducing herein results from
\cite{Welk-ssvm13,Welk-JMIV15,Welk-Aiep14,Welk-JMIV11}.
Regarding the association between amoeba metrics on the discrete filtering side
and edge-stopping functions occurring in the corresponding PDEs, the current
work adds to the previously considered exemplary $L^1$ and $L^2$ (Euclidean)
amoeba metrics as a third simple case the $L^\infty$ (maximum) amoeba metric
and states explicitly the corresponding edge-stopping function.
Moreover, the amoeba variants of morphological opening and closing
are included in the description for the first time. For dilation, erosion,
opening and closing filters, the presentation here emphasises the
algebraic background including max-plus/min-plus convolution and conjugacy
of structure elements.

Section~\ref{sec:seg} considers the application of
amoeba techniques to devise basic algorithms for unsupervised segmentation
of grey-value images,
namely the \emph{amoeba active contours (AAC)} first
introduced in \cite{Welk-ssvm11} and further investigated in
\cite{Welk-ssvm13,Welk-JMIV15}.
Results from \cite{Welk-JMIV15} on the
relation between AAC and geodesic active contours are reported.

In image filtering by nonlinear PDEs, one often computes the nonlinearities
not from the input images themselves but from Gaussian pre-smoothed versions
of these, in order to reduce noise sensitivity of filters and to improve
numerical stability. This is also the case with self-snakes and active
contour PDEs; note that the self-snakes PDE is even ill-posed without such
pre-smoothing.
Section~\ref{sec:presmoothing} investigates the effect of pre-smoothing in
the self-snakes PDE using perturbation analysis on a synthetic example;
furthermore, it discusses how a comparable stabilisation can be achieved in
the amoeba median filter framework. The analysis presented in this section
relies on previous work in \cite{Welk-ssvm13,Welk-Aiep14} in which
oscillatory perturbations aligned with the gradient direction were studied,
and extends it by including also perturbations aligned with the level line
direction.

Section~\ref{sec:texture} is devoted to a different direction of application
of amoeba ideas. Noticing that the computation of discrete amoeba structuring
elements is intimately related with graph structures -- a weighted
neighbourhood graph, weighted and unweighted Dijkstra search trees -- in
the neighbourhood of each pixel, one can try to extract local texture
information from these graphs. Quantitative graph theory \cite{Dehmer-Book12}
offers a variety
of graph indices for generating quantitative information from graph structures.
The presentation of the construction of texture descriptors from amoebas and
graph indices in this section follows \cite{Welk-qgt14}. Compared to the
large set of descriptors covered in \cite{Welk-qgt14}, only a few
representatives are shown here, complementing their mathematical description
by a visualised example.
Extending the previous work on texture discrimination in \cite{Welk-qgt14},
the present paper also shows a first example of the new texture descriptors
in texture segmentation by using the descriptors as components of an input
image for multi-channel GAC segmentation.

\section{Morphological Amoebas}
\label{sec:amoebas}

Well-known local image filters such as the mean filter, median filter,
morphological dilation or erosion consist of two steps: a sliding-window
\emph{selection} step, and the \emph{aggregation} of selected input values
by taking e.g.\ the arithmetic mean, median, maximum or minimum.
A strategy to improve the sensitivity of such filters to
important image structures is to modify the selection step by using
spatially adaptive neighbourhoods instead of a fixed sliding window.
The general idea is to give preference in the selection
to neighbouring image locations with similar intensities, and thus to
reduce the flow of grey-value information across high contrast steps
or slopes in the filter process.

First introduced by Lerallut et al.\ \cite{Lerallut-ismm05,Lerallut-IVC07}
as structuring elements for adaptive morphology,
morphological amoebas are a specific type of such spatially adaptive
neighbourhoods. Their construction relies on the combination of
spatial distance in the image domain with grey-value contrast into
a modified metric on the image.

\subsection{Edge-weighted Neighbourhood Graph}
\label{subsec:nbhgraph}

To define morphological amoebas on discrete images, we start by
considering edge-weighted graphs based on the image grid.

\begin{definition}
Let $f$ be a discrete image. Construct an edge-weighted graph
$G_w(f):=(V,E,w)$ with vertex set $V$, edge set $E$ and weights $w$
as follows. The vertex set $V$ is formed by all pixels of $f$.
Two vertices $i$, $j$ are connected, $\{i,j\}\in E$, if and only if
pixels $i$, $j$ are neighbours under a suitably chosen neighbourhood
notion.
To define the edge weights $w_{i,j}$ for an edge
$\{i,j\}\in E$, consider the
corresponding pixel locations $\bm{p}_i$ and $\bm{p}_j$ as well as the
intensities $f_i$ and $f_j$, and set $w_{i,j}$ to
\begin{equation}
w_{ij} := \varphi \bigl(
\lVert\bm{p}_i-\bm{p}_j\rVert_2, \beta\,\lvert f_i-f_j\rvert\bigr)
\label{Gwfweights-generic}
\end{equation}
where $\lVert\bm{p}_i-\bm{p}_j\rVert_2$ denotes Euclidean distance in the
image plane, $\beta>0$ is a contrast scale parameter weighting between
spatial and tonal distances,
and $\varphi$ is a norm on $\bbbr^2$ which can be rewritten as
\begin{align}
\varphi(s,t) & =\begin{cases}
\lvert t\rvert\cdot\nu(\lvert s/t\rvert)\;,&t>0\;,\\
\lvert s\rvert\;,&t=0
\end{cases}
\label{ametphi}
\end{align}
with a monotonically increasing function $\nu:\bbbr^+_0\to\bbbr^+$
(by continuity, $\nu(0)=1$).

The edge-weighted graph $G_w(f)$ is called \emph{neighbourhood graph} of $f$.
\end{definition}

In this definition, neighbourhood can be understood as a
4-neighbourhood, as done in \cite{Lerallut-ismm05}, or as an 8-neighbourhood
as in \cite{Welk-qgt14,Welk-Aiep14,Welk-JMIV11}. The latter choice gets
somewhat closer to a Euclidean measurement of spatial distances in the image
plane and is therefore also considered the default in the present work.

As to the norm function $\nu$, the setting $\nu(z)\equiv\nu_1(z)=1+z$
corresponds to the $L^1$ metric also used in \cite{Lerallut-ismm05} that gives
\begin{equation}
w_{ij}=\lVert\bm{p}_i-\bm{p}_j\rVert_2+\beta\,\lvert f_i-f_j\rvert\;,
\end{equation}
whereas $\nu(z)\equiv\nu_2(z)=\sqrt{1+z^2}$ entails a Euclidean ($L^2$)
metric in which the edge weights are obtained by the Pythagorean sum
\begin{equation}
w_{ij}=\sqrt{\lVert\bm{p}_i-\bm{p}_j\rVert_2^2+\beta^2\lvert f_i-f_j\rvert^2}
\;.
\end{equation}
A straightforward generalisation is
\begin{equation}
\nu_p(z)=(1+z^p)^{1/p} \quad\text{for}~p\ge1\;,
\label{nup}
\end{equation}
which in the limit $p\to+\infty$ also includes
$\nu_\infty(z)=\max\{1,z\}$ and the corresponding edge weight
\begin{equation}
w_{ij}=\max\left\{\lVert\bm{p}_i-\bm{p}_j\rVert_2,\beta\,
\lvert f_i-f_j\rvert \right\}\;.
\end{equation}

\subsection{Discrete Amoeba Metric}
\label{subsec:amoebametric-disc}

We use the edge-weighted neighbourhood graph to define
the discrete amoeba metric on image $f$.

\begin{definition}
Let a discrete image $f$ be given. Let $G_w(f)$ be its neighbourhood graph
with edge weights given by \eqref{Gwfweights-generic}.
Define for two pixels $i$ and $j$ their distance $d(i,j)$ as the
minimal total weight (length) among all paths between $i$ and $j$
in $G_w(f)$. Then $d$ is called \emph{(discrete) amoeba metric} on $f$.

The metric $d$ is called $L^p$ amoeba metric, $1\le p<\infty$,
if it is derived from \eqref{nup}, or $L^\infty$ amoeba metric if it is
obtained from $\nu(z)=\max\{1,z\}$. The $L^2$ amoeba metric is also
called Euclidean amoeba metric.
\end{definition}

\begin{definition}
In a discrete image $f$ with amoeba metric $d$,
an \emph{amoeba structuring element} (short: \emph{amoeba})
$\mathcal{A}_\varrho(i)\equiv\mathcal{A}_\varrho(f;i)$ with amoeba
radius $\varrho$ and reference point at pixel $i$ is a discrete
$\varrho$-ball around pixel $i$ in the amoeba metric,
i.e.\ the set of all vertices within a distance $\varrho$ from $i$,
\begin{equation}
\mathcal{A}_\varrho(i):=\{j~|~d(i,j)\le\varrho\}\;.
\end{equation}
\end{definition}

The derivation of amoebas from a metric with a global radius parameter
$\varrho$ has an interesting consequence: for two pixels $i$, $j$, one has
\begin{equation}
i\in\mathcal{A}_\varrho(j)\quad\Leftrightarrow\quad j\in\mathcal{A}_\varrho(i)
\;,
\label{mutu}
\end{equation}
which is helpful in the design of some morphological filters.

\subsection{Computation of Discrete Amoebas}
\label{subsec:amoebacomp-disc}

To compute amoebas in a discrete image, one has to search the neighbourhood
of each given reference pixel $i$ in order to identify the pixels $j$ with
amoeba distance $d(i,j)\le\varrho$.
Given that the edge weights $w_{i,j}$ in $G_w(f)$ are nonnegative,
this can be achieved by running Dijkstra's shortest path algorithm
\cite{Dijkstra-NUMA59} on $G_w(f)$ starting at pixel $i$. As this algorithm
enumerates neighbour pixels in order of increasing path weight, it can
be stopped as soon as a pixel $j$ with $d(i,j)>\varrho$ is visited.

Moreover, by the construction of the amoeba distance it is clear that
the Euclidean distance in the image domain is a lower bound for the amoeba
distance between pixels. Therefore the Dijkstra algorithm for the start vertex
$i$ can be run on the subgraph of $G_f(w)$ that contains just the pixels from
the Euclidean $\varrho$-neighbourhood of $i$.

\subsection{Amoebas on Continuous Domains}
\label{subsec:amoebas-cont}

Even superficial inspection of results obtained by some amoeba filters
indicates that they have striking similarities to image processing methods
based on partial differential equations (PDEs). This observation has been
substantiated in \cite{Welk-JMIV15,Welk-Aiep14,Welk-JMIV11} by studying
space-continuous versions of amoeba filters; the results proven there allow
to interpret amoeba filters as time steps of explicit discretisations
for suitable PDEs.

To devise space-continuous versions of amoeba filters, one has to translate
first the notion of amoeba metric to the space-continuous setting. Once this
is done, the definition of an amoeba as a $\varrho$-ball around a reference
point is straightforward.

The amoeba metric for a space-continuous greyvalue image -- a real-valued
function $f$ over a connected compact image domain $\varOmega\subset\bbbr^n$ --
can be stated by assigning to each two given points $\bm{p},\bm{q}\in\varOmega$
as their distance the minimum of a path integral between
$\bm{p}$ and $\bm{q}$. Just like the edge weights in the discrete
amoeba construction, the integrand of the path integral is obtained by
applying a suitable norm $\varphi$ to the spatial metric (the Euclidean curve
element of the path) and the greyvalue metric (the standard metric on the
real domain), such that the amoeba distance reads as
\begin{align}
d(\bm{p},\bm{q}) &= \min\limits_{\bm{c}} \int\limits_0^1\varphi\bigl(
\lVert\bm{c}'(t)\rVert_2,\beta\,\lvert (f\circ \bm{c})'(t)\rvert\bigr)
\,\mathrm{d}t
= \min\limits_{\bm{c}} \int\limits_0^1\varphi\bigl(
\lVert\bm{c}'(t)\rVert_2,
\beta\,\lvert \bm{\nabla}f^{\mathrm{T}}\bm{c}'(t)\rvert
\bigr)\,\mathrm{d}t
\label{ametcont}
\end{align}
where $\bm{c}$ runs over all regular curves $\bm{c}:[0,1]\to\varOmega$ with
$\bm{c}(0)=\bm{p}$, $\bm{c}(1)=\bm{q}$, and $\varphi$ can be chosen as in
the discrete case.

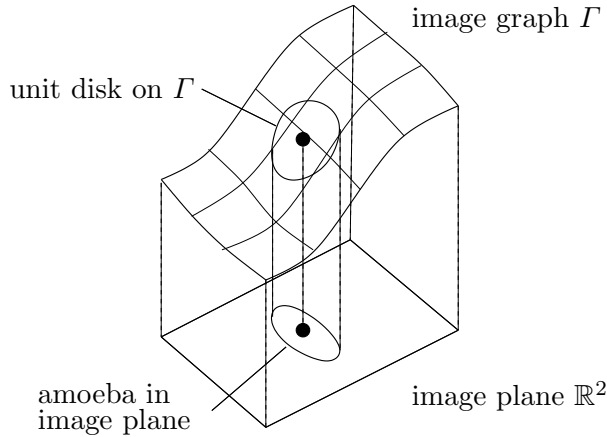
\begin{figure}[t]
\unitlength0.001\textwidth
\begin{picture}(500,360)
\put (140,  0){
\includegraphics[width=253\unitlength]{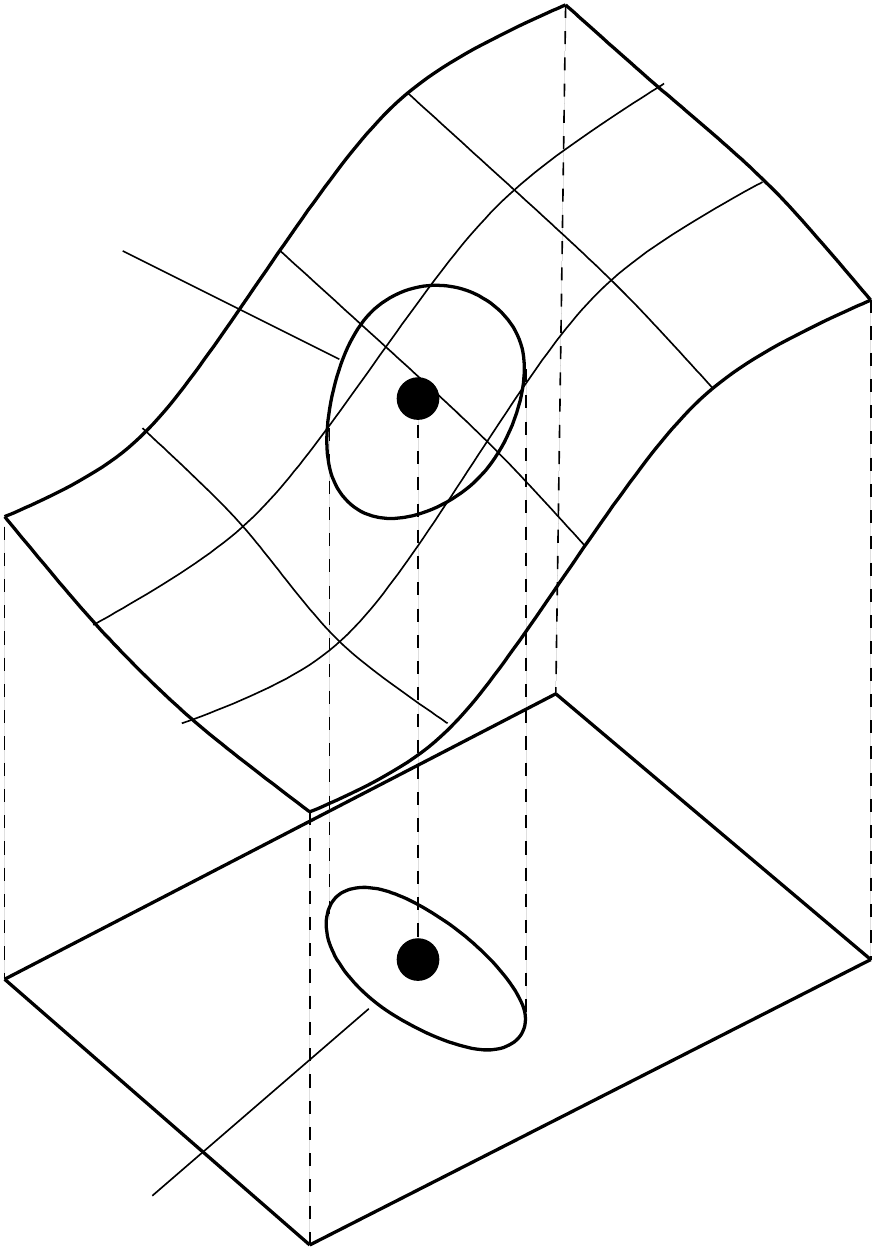}}
\put ( 22,280) {unit disk on $\varGamma$}
\put ( 46, 22) {amoeba in}
\put ( 46,  0) {image plane}
\put (360,340) {image graph $\varGamma$}
\put (360, 20) {image plane $\bbbr^2$}
\end{picture}
\caption{\label{fig:am3d}Amoeba as projection of the unit disk on the image
graph to the image plane.}
\end{figure}

Let us associate to the function $f:\bbbr^2\supset\varOmega\to\bbbr$ its
(vertically rescaled) \emph{graph,} the manifold
$\varGamma:=\{(x,y,\beta\,f(x,y))~|~(x,y)\in\varOmega\}\subset\bbbr^3$.
Then we see that the amoeba distance $d(\pm{p},\pm{q})$ between two points
$\pm{p}$, $\pm{q}$ in the image domain $\varOmega$ can be interpreted as
a distance $\hat{d}(\bm{p}',\bm{q}')$ on $\varGamma$.
The points $\bm{p}',\bm{q}'\in\varGamma$ herein are given by
$\bm{p}':=(\bm{p},f(\bm{p}))$, $\bm{q}':=(\bm{q},f(\bm{q}))$.
To define the metric $\hat{d}$ on $\varGamma$, consider a metric $\tilde{d}$ in
the surrounding space $\bbbr^3$ that
combines Euclidean metric in the $x$-$y$-plane with
the standard metric in $z$-direction via the function $\varphi$ from
\eqref{ametphi} that appeared already in the original construction of the
amoeba metric.
Using $\tilde{d}$ in $\bbbr^3$,
the metric $\hat{d}$ is obtained as its induced metric
on the submanifold $\varGamma\subset\bbbr^3$.
Figure~\ref{fig:am3d} illustrates that the amoeba structuring element
is then the projection of a unit disk on $\varGamma$ back to the image plane.

\begin{figure}[t]
\unitlength1.1mm
\begin{picture}(138, 39)
\put(3,0){\includegraphics[height=39\unitlength]
    {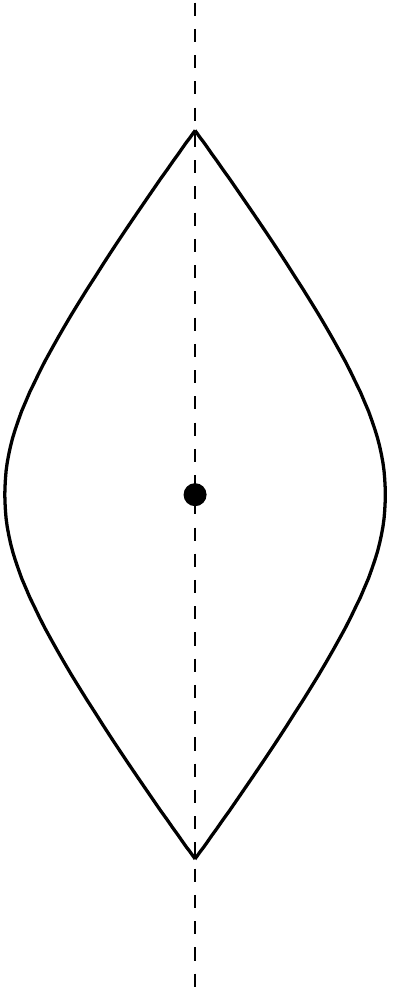}}
\put(5,0){(a)}
\put(25,0){\includegraphics[height=39\unitlength]
    {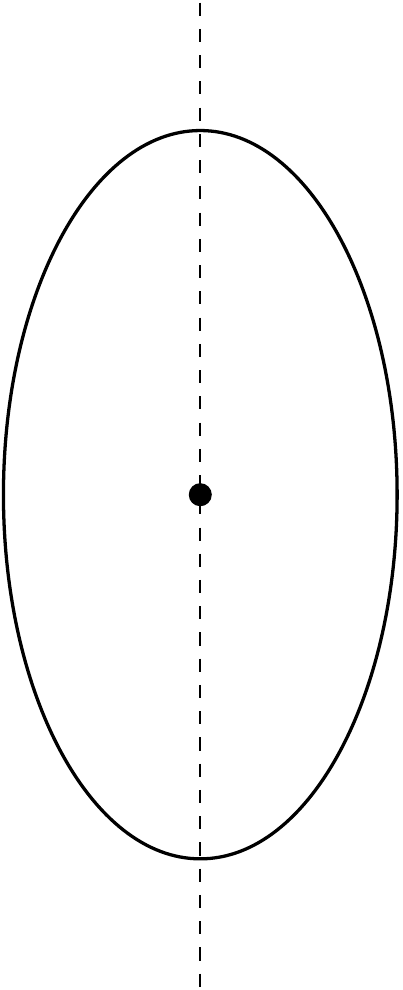}}
\put(27,0){(b)}
\put(47,0){\includegraphics[height=39\unitlength]
    {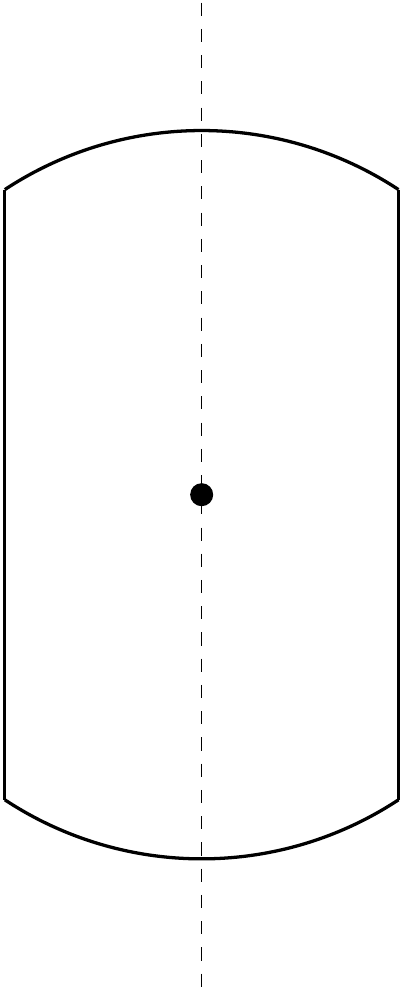}}
\put(49,0){(c)}
\put(75,0){\includegraphics[height=39\unitlength]
    {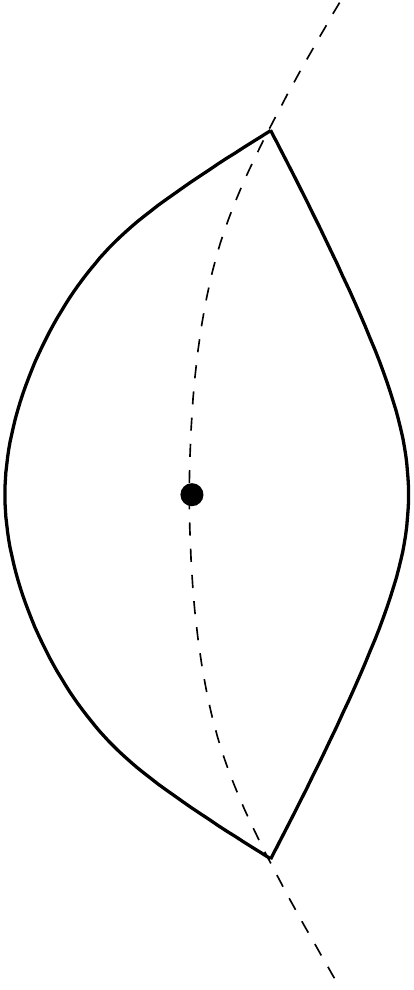}}
\put(77,0){(d)}
\put(97,0){\includegraphics[height=39\unitlength]
    {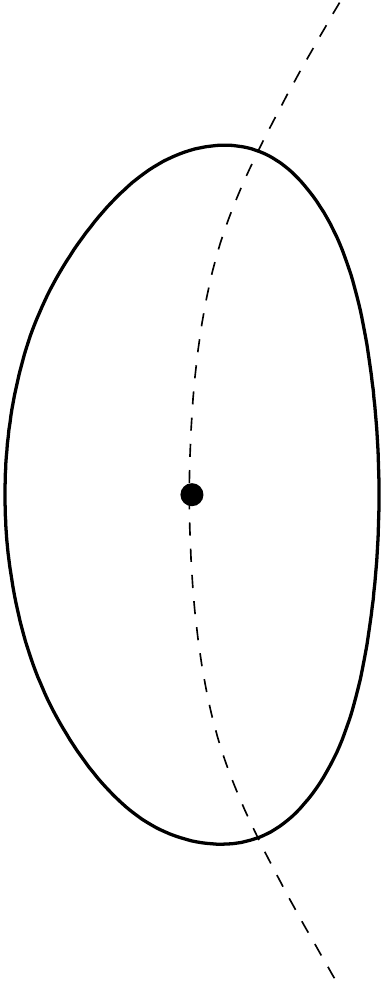}}
\put(99,0){(e)}
\put(119,0){\includegraphics[height=39\unitlength]
    {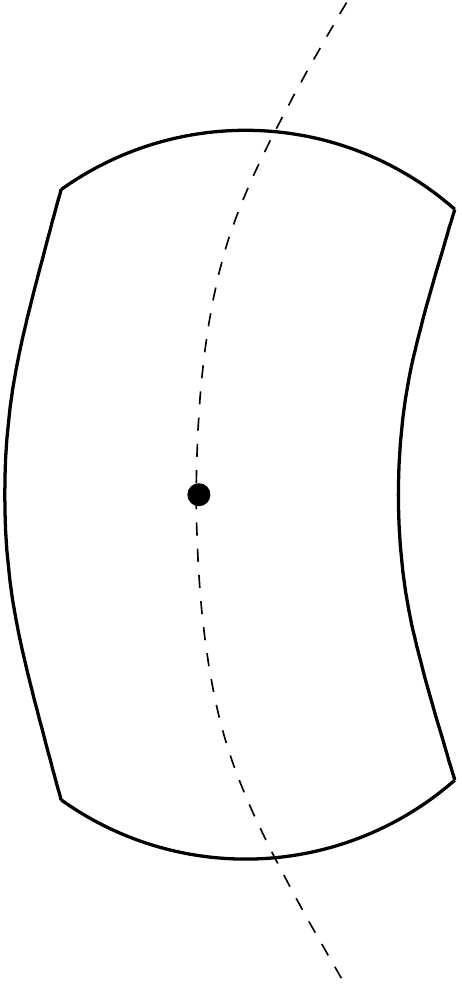}}
\put(121,0){(f)}
\end{picture}
\caption{\label{fig:amshapes}Typical shapes of amoebas in the
continuous domain for different amoeba metrics. The left group of three,
(a--c), shows
amoebas on an image with equidistant straight level lines,
the right group of three, (d--f), 
shows amoebas on curved level lines (schematic).
Frames (a) and (d) show $L^1$ amoeba metric, (b) and (e) show Euclidean
amoeba metric, (c) and (f) show the maximum amoeba
metric. Each amoeba is shown with its reference point (bold) and 
level line through the reference point (dashed). 
}
\end{figure}
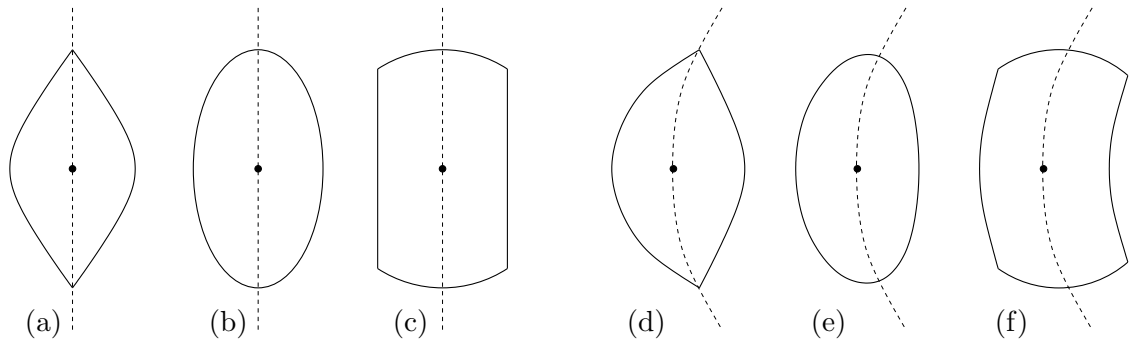

Figure~\ref{fig:amshapes} shows typical amoeba shapes in smooth
image regions for the three exemplary amoeba metrics exposed in
Section~\ref{subsec:amoebametric-disc}.

\section{Amoeba-Based Image Filters}
\label{sec:filters}

To obtain applicable image filters, the amoeba procedure described
above is used as a selection step and
needs to be complemented by some aggregation step.
We consider here standard choices of aggregation operators from
classical local filters; introducing also modifications into this
part of the filtering procedure is left as a possible direction
for future research. Moreover, keeping close to the original context
in which amoebas were developed, we focus on morphological operators.
Here, morphological operators are characterised by their invariance
under arbitrary monotonically increasing transformations of the
intensities, see e.g.\ \cite{Maragos-icip09}, which means that also
median and quantiles belong to this class.

\subsection{Median}
\label{subsec:median}

A median filter aggregates the intensity values of the selected
pixels by taking their median.
In the non-adaptive, sliding-window
setting this filter can be traced back to Tukey~\cite{Tukey-Book71},
and since then it has gained high popularity as a simple
denoising filter that preserves discontinuities (edges) and its robustness
with respect to some types of noise.
Median filtering can be iterated. Unlike average filters, the
median filter on a discrete image possesses non-trivial steady
states, so-called root signals \cite{Eckhardt-JMIV03}, that depend on
the filter window. The smaller the filter window, the faster the
iterated median filtering process locks in at a root signal.

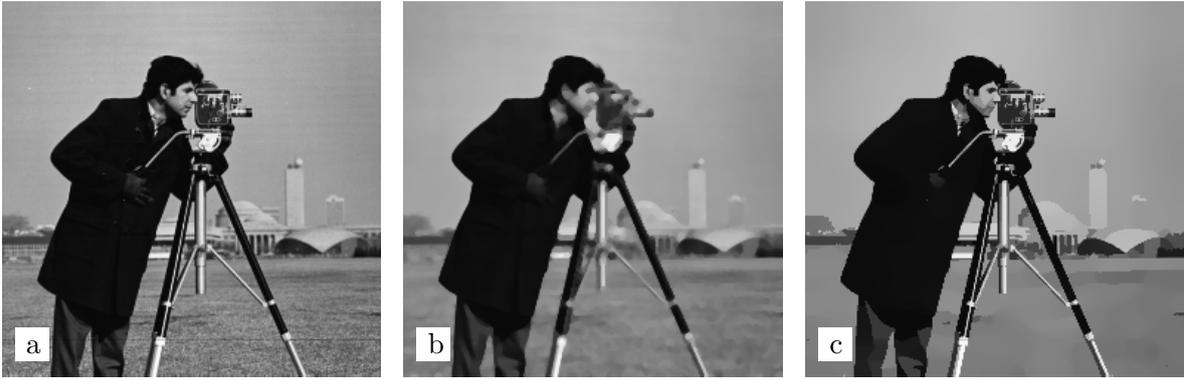
\begin{figure}[t]
\unitlength0.01\textwidth
\begin{picture}(100,32)
\put(0,0){\includegraphics[width=32\unitlength]
    {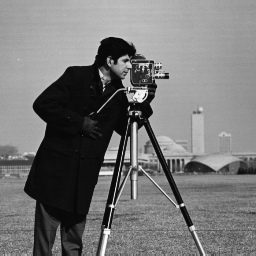}}
\put(34,0){\includegraphics[width=32\unitlength]
    {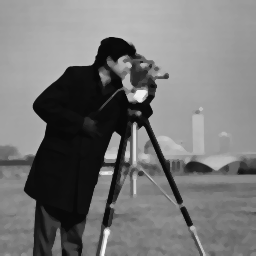}}
\put(68,0){\includegraphics[width=32\unitlength]
    {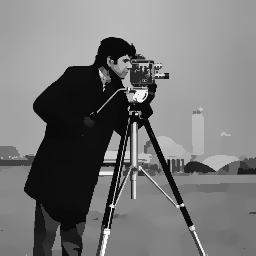}}
\put( 1.1,2){\colorbox{white}{\rule{0pt}{.6em}\hbox to.6em{\kern.1em\smash{a}}}}
\put(35.1,2){\colorbox{white}{\rule{0pt}{.6em}\hbox to.6em{\kern.1em\smash{b}}}}
\put(69.1,2){\colorbox{white}{\rule{0pt}{.6em}\hbox to.6em{\kern.1em\smash{c}}}}
\end{picture}
\caption{\label{fig:iamf}Non-adaptive and amoeba median filtering.
\textbf{(a)} Original image. --
\textbf{(b)} Filtered by 5 iterations of standard median filtering
with a discrete disk of radius $2$ as structuring element. --
\textbf{(c)} Filtered by 5 iterations of amoeba median filtering
with Euclidean amoeba metric, $\beta=0.2$, $\varrho=7$.
}
\end{figure}

Despite the nice preservation of edges, the non-adaptive median
filter involves a displacement of curved edges in inward direction
and rounding of corners that is often undesired. Amoeba median
filtering greatly reduces this effect. Figure~\ref{fig:iamf}
demonstrates this by an example.

\subsubsection{PDE Approximation}

As noticed already in 1997 by Guichard and
Morel \cite{Guichard-sana97},
the overall robust denoising effect and the characteristic
corner-rounding behaviour of standard median filtering resemble
the properties of the well-known (mean) curvature motion PDE
\cite{Alvarez-SINUM92}. Further analysis confirmed
this observation by proving an asymptotic relationship between
the two filters, as set forth in the following proposition.

\begin{proposition}[Guichard and Morel, 1997 \cite{Guichard-sana97}]
\label{prop:imfpde}
For a smooth function $u:\varOmega\to\bbbr$, one iteration of
median filtering with a $\varrho$-ball as structuring element
approximates for $\varrho\to0$ a time step of size $\tau=\varrho^2/6$ of
the curvature motion PDE \cite{Alvarez-SINUM92}
\begin{equation}
\label{mcm}
u_t = \lvert\bm{\nabla}u\rvert\,\mathrm{div}\left(
\frac{\bm{\nabla}u}{\lvert\bm{\nabla}u
\rvert}\right) \;.
\end{equation}
\end{proposition}

This seminal result motivates the investigation
of relations between amoeba and PDE filters whose results are
reviewed in the further course of the present paper.

Just like amoeba median filtering differs from standard
median filtering by an adaptation procedure that suppresses
smoothing across edges, the curvature motion equation \eqref{mcm}
has a counterpart in which also the flow across edges is suppressed.
This so-called self-snakes filter \cite{Sapiro-icip96} allows
curvature-based image smoothing and simplification, preserves and
even enhances edges, while at the same time avoiding to shift them,
as curvature motion does. It turns out that indeed amoeba median
filtering is connected to self-snakes by a similar asymptotic
relationship as that of Proposition~\ref{prop:imfpde}, as follows.

\begin{theorem}[\cite{Welk-Aiep14,Welk-JMIV11}]
\label{thm:iamfpde}
For a smooth function $u:\varOmega\to\bbbr$, one iteration of
amoeba median filtering with amoeba radius $\varrho$ approximates
for $\varrho\to0$ a time step of size $\tau=\varrho^2/6$ of
the self-snakes PDE \cite{Sapiro-icip96}
\begin{equation}
\label{ssn}
u_t = \lvert\bm{\nabla}u\rvert\,\mathrm{div}\left(
g\bigl(\lvert\bm{\nabla}u\rvert\bigr)\,\frac{\bm{\nabla}u}{\lvert\bm{\nabla}u
\rvert}\right)
\end{equation}
where $g:\bbbr_0^+\to\bbbr_0^+$ is a decreasing edge-stopping function
that depends on the amoeba metric being used.
\end{theorem}

Proofs for Theorem~\ref{thm:iamfpde} have been given in
\cite{Welk-Aiep14,Welk-JMIV11}. While these proofs are not reproduced in
detail here, it is of interest to describe the two different strategies
that are used in these proofs. These approaches form also the basis for
the further amoeba---PDE asymptotics results presented
in Section~\ref{subsec:dilero}.

\subsubsection{Proof Strategies}
\label{subsubsec:iamfpdeproof}

The crucial observation for all median filter---PDE equivalence results
since Gui\-chard and Morel's proof of Proposition~\ref{prop:imfpde}
in \cite{Guichard-sana97} is that the median of a smooth function
$u$ within a given compact structuring element $\mathcal{A}$ is the function
value whose corresponding level line divides the structuring element into two
parts of equal area. Herein it is assumed that each value of $u$ within the
structuring element is associated with a unique level line segment inside
$\mathcal{A}$, which is satisfied for sufficiently small fixed or amoeba
structuring elements whose reference point $\bm{x}_0$ is not an extremum of
$u$, and therefore acceptable when studying the limit $\varrho\to0$.

\begin{figure}[t]
\unitlength0.001\textwidth
\begin{picture}(630,350)
\put( 30,0){\includegraphics[height=350\unitlength]{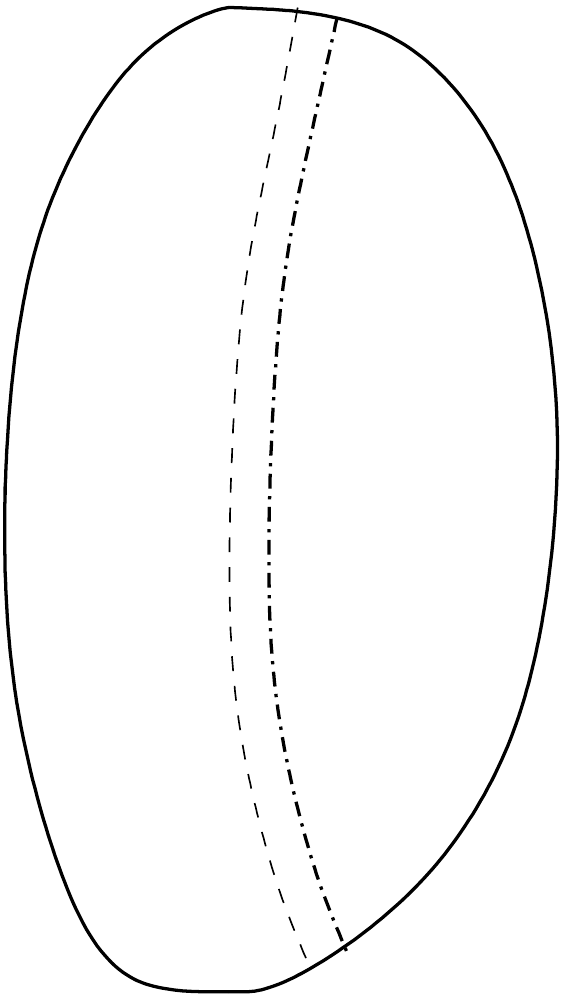}}
\put(119,165){$\bullet$}
\put(131,160){$\bm{x}_0$}
\put(250,0){(a)}
\put(360,0){\includegraphics[height=350\unitlength]
  {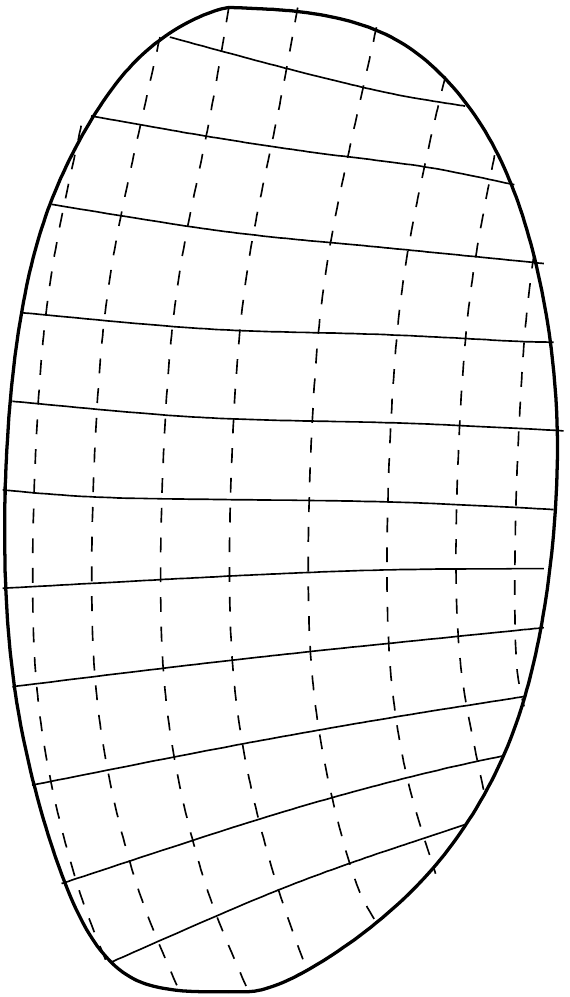}}
\put(426,156){\colorbox{white}{\rule{0pt}{.8ex}\!\!\smash{$\bm{x}_0$}\!\!}}
\put(449,167){$\bullet$}
\put(580,0){(b)}
\end{picture}
\caption{\label{fig:bisect}
\textbf{(a)} Amoeba with reference point $\bm{x}_0$,
level line through $\bm{x}_0$ (dot-dashed) and bisecting level line
(dashed), schematic. --
\textbf{(b)} Amoeba with curvilinear coordinate system
formed by level lines (dashed) and gradient flow lines (solid).}
\end{figure}
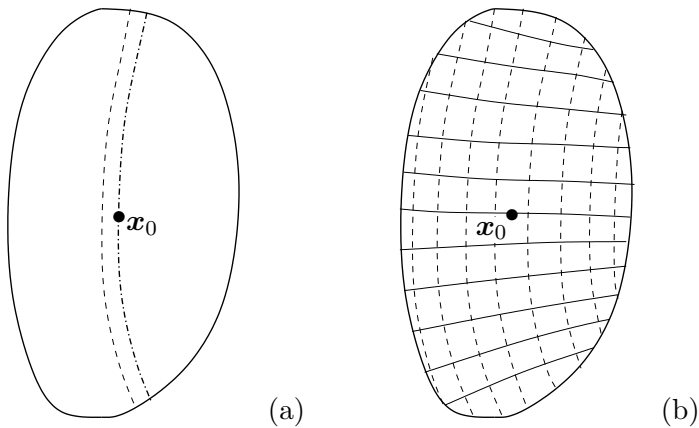

The amount by which a single median filtering step
changes the function value at the reference point $\bm{x}_0$ of the
structuring element then corresponds, up to multiplication with
$\lvert\bm{\nabla}u\rvert$, to the distance between the area-bisecting
level line and the level line through $\bm{x}_0$, see the illustration in
Figure~\ref{fig:bisect}(a). The two approaches
discussed in the following differ in the way how they measure the area
of the structuring elements and parts thereof.

\paragraph{Proof Strategy I.}

The first strategy has been followed in \cite{Welk-JMIV11} to prove
Theorem~\ref{thm:iamfpde} for the entire class of amoeba metrics discussed
in Section~\ref{sec:amoebas} above, see also the more detailed version
in \cite[Section 4.1.1]{Welk-Aiep14}.
It is close to the approach from
\cite{Guichard-sana97} in that it develops the smooth function $u$ around
the reference point $\bm{x}_0$ into a Taylor expansion up to second order.
The Taylor expansion is then used to approximate, for an amoeba
$\mathcal{A}=\mathcal{A}_\varrho(\bm{x}_0)$, three items: first,
the range of function values occurring within $\mathcal{A}$, i.e.\ the
minimum $\min_\mathcal{A}u$ and maximum $\max_\mathcal{A}u$, second,
the length $L(z)$ of the level line segment for each
$z\in[\min_\mathcal{A}u,\max_\mathcal{A}u]$, and third, the density
$\delta(z)$ of level lines around each $z$, which equals the steepness of
the slope of $u$ near the level line of $z$.

Integrating the lengths of level lines over function values, weighted with
their reciprocal densities, yields the area of $\mathcal{A}$, i.e.\
\begin{equation}
\mathrm{Area}(\mathcal{A}) = 
\int\limits_{\min_\mathcal{A} u}^{\max_\mathcal{A} u}
\frac{L(z)}{\delta(z)}\,\mathrm{d}z\;.
\label{Aarea-cartesian}
\end{equation}
As this integral effectively runs over level lines, splitting the integration
interval exactly corresponds to cutting $\mathcal{A}$ at some level line.
The calculation of the desired median of $u$ within $\mathcal{A}$
is then achieved by determining a suitable splitting point in the integration
interval so that the integrals on both sub-intervals become equal.

Summarising, this strategy describes the amoeba shape in terms of a
curvilinear coordinate system aligned with the gradient and level
line directions at $\bm{x}_0$, in which the level lines take the role of
coordinate lines, compare Figure~\ref{fig:bisect}(b).

\paragraph{Proof Strategy II.}

The second strategy abandons the consideration of the individual
level lines within $\mathcal{A}$; the only level line that is explicitly
studied is the one through $\bm{x}_0$ itself.
Instead of the distorted Cartesian
coordinate system one uses polar coordinates to describe the shape of the
amoeba. This approach has first been used in \cite{Welk-ssvm13} in the
context of amoeba active contours (see Section~\ref{subsec:aac}),
and again in
\cite[Section 4.1.2]{Welk-Aiep14},
both times restricted to the Euclidean amoeba metric.
It has been extended to cover the full generality of amoeba metrics under
consideration in \cite{Welk-JMIV15}, again for amoeba active contours.

\begin{figure}[t]
\setlength{\unitlength}{0.001\textwidth}
\begin{picture}(1000,350)
\put( 60,  0){\includegraphics[height=350\unitlength]{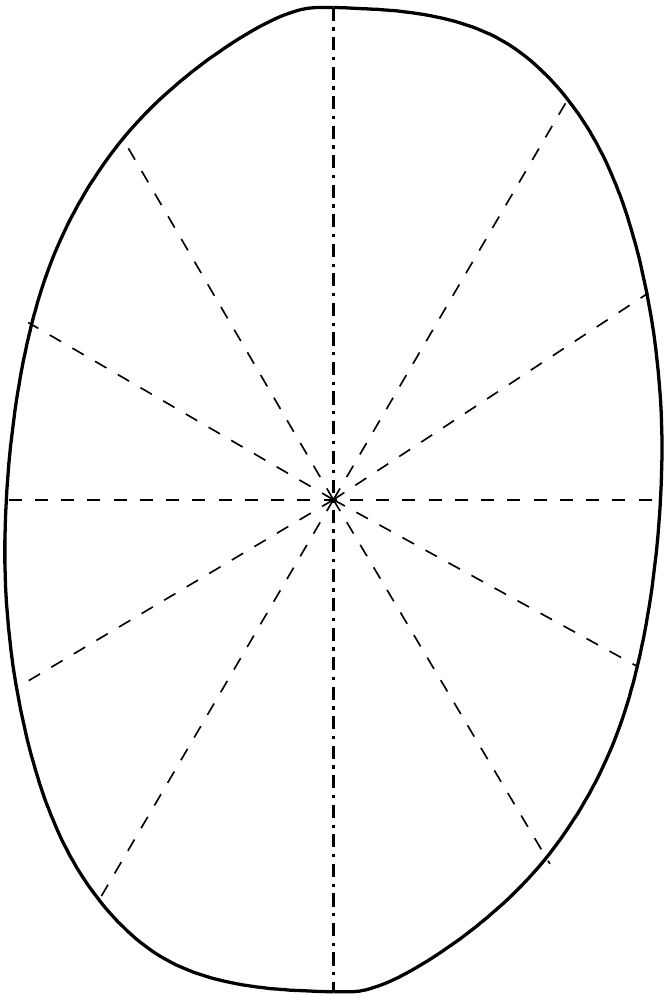}}
\put(171,168){$\bullet$}
\put(145,163){\colorbox{white}{\rule{0pt}{.8ex}\!\!\smash{$\bm{x}_0$}\!\!}}
\put( 40,  0){(a)}
\put(420,  0){\includegraphics[height=350\unitlength]{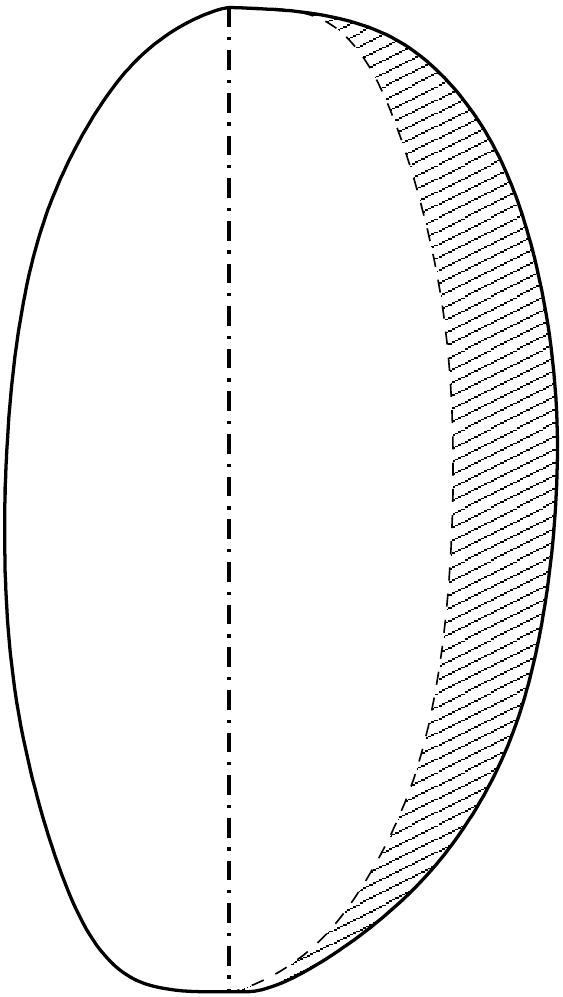}}
\put(494,168){$\bullet$}
\put(467,160){$\bm{x}_0$}
\put(500,263){\line(1,0){10}}
\put(512,260){{\tiny straight}}
\put(512,246){{\tiny level}}
\put(512,232){{\tiny line}}
\put(585,190){\colorbox{white}{\rule{0pt}{1ex}\!\!\smash{$\varDelta_1$}\!\!}}
\put(596,290){\line(1,0){12}}
\put(612,301){{\tiny asymmetric}}
\put(612,287){{\tiny amoeba}}
\put(400,  0){(b)}
\put(740,  0){\includegraphics[height=350\unitlength]{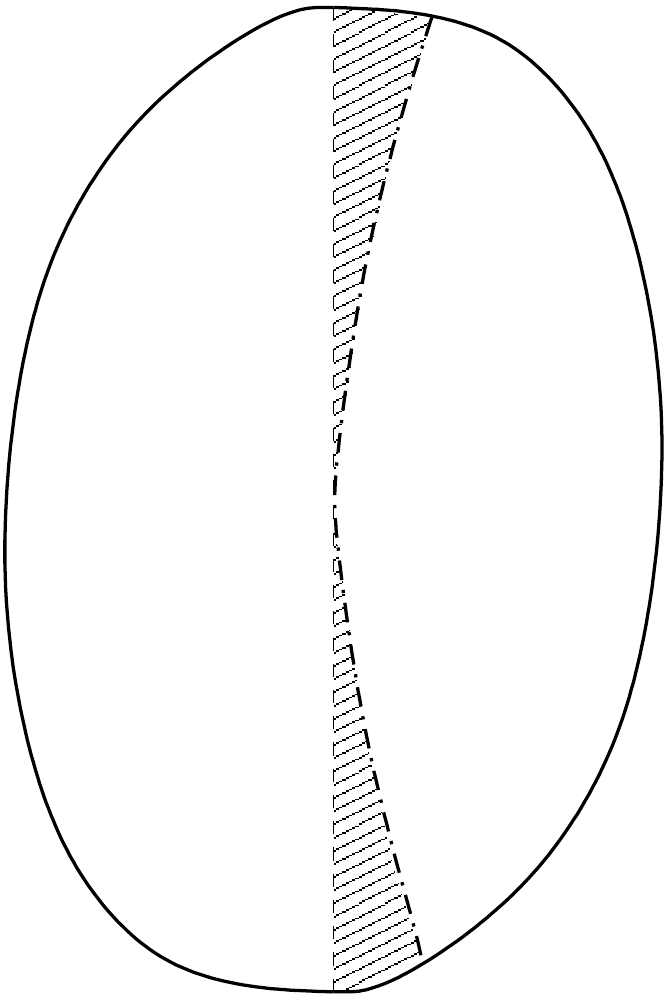}}
\put(851,168){$\bullet$}
\put(822,160){$\bm{x}_0$}
\put(870,263){\line(1,0){10}}
\put(882,260){{\tiny curved}}
\put(882,246){{\tiny level}}
\put(882,232){{\tiny line}}
\put(855,30){\colorbox{white}{\rule{0pt}{1ex}\!\!\smash{$\varDelta_2$}\!\!}}
\put(953,290){\line(1,0){12}}
\put(969,301){{\tiny symmetric}}
\put(969,287){{\tiny amoeba}}
\put(720,0){(c)}
\end{picture}
\caption{\label{fig:delta}%
\textbf{(a)} Amoeba with straight level line (dot-dashed) through its
reference point $\bm{x}_0$ and further radial lines (dashed)
of a polar coordinate system centred at $\bm{x}_0$.
\textbf{(b)} Area difference $\varDelta_1$ in an asymmetric amoeba with
straight level lines. The hashed region is enclosed between the right
arc of the amoeba contour and the point-mirrored copy of its left arc. --
\textbf{(c)} Area difference $\varDelta_2$ in a symmetric amoeba with
curved level lines. --
(b), (c) from \cite{Welk-ssvm13}.}
\end{figure}
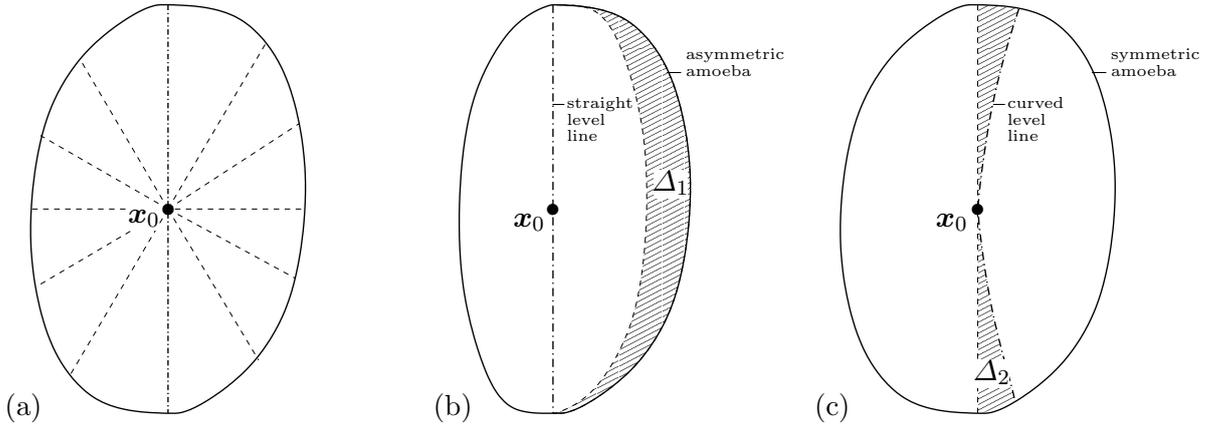

Writing the outline of $\mathcal{A}$ as a function $a(\alpha)$
of the polar angle $\alpha\in[0,2\pi]$, the amoeba's area is stated by
the standard integral for areas enclosed by function graphs in polar
coordinates as
\begin{equation}
\mathrm{Area}(\mathcal{A}) 
= \frac12\int\limits_0^{2\pi}a(\alpha)^2\,\mathrm{d}\alpha\;.
\label{Aarea-polar}
\end{equation}
Unlike for \eqref{Aarea-cartesian}, splitting this integral yields
areas of sectors instead of segments; however, if the level line through
$\bm{x}_0$ happens to be a straight line, splitting up the integral
\eqref{Aarea-polar} at the pair of opposite angles corresponding to the
level line direction yields the areas of two segments into which
$\mathcal{A}$ is cut by that level line, compare Figure~\ref{fig:delta}(a).

Provided that $\mathcal{A}$ is symmetric (w.r.t.\ point reflection at the
reference point), the two segments are of equal area, making in this
case the median equal to $u(\bm{x}_0)$.
Deviations from this situation that make the median differ from $u(\bm{x}_0)$
can be separated into two contributions: first, the asymmetry of the amoeba;
second, the curvature of the level lines. Cross-effects of the two
contributions influence
only higher order terms that can be neglected in the asymptotic analysis;
thus the two sources can be studied independently. In approximating
the area difference $\varDelta_1$ caused by the asymmetric amoeba shape,
one can assume that the level lines are straight, see
Figure~\ref{fig:delta}(b), while the level line curvature effect $\varDelta_2$
can be studied under the assumption that $\mathcal{A}$ has symmetric shape,
see Figure~\ref{fig:delta}(c).

Finally, the combined effect $\varDelta_1+\varDelta_2$ must be compensated
by a parallel shift of the level line through $\bm{x}_0$, compare again
Figure~\ref{fig:bisect}(a). From the shift the median, and thus
the right-hand side of the PDE approximated by the amoeba filter,
can be derived.

\subsubsection{Amoeba Metrics and Edge-Stopping Functions}

It remains to specify the relation between amoeba metric and edge-stopping
function mentioned in Theorem~\ref{thm:iamfpde}. In
\cite{Welk-JMIV11,Welk-Aiep14}, the following representation of $g$
in terms of the function $\nu$ defining the amoeba metric has been proven.
\begin{equation}
g(z) = \frac{3}{\beta^2s^2\nu^3(1/(\beta z))} \int\limits_0^1
\xi^2 \sqrt{\nu^{-2}\left(\frac1\xi\,\nu\left(\frac1{\beta z}\right)\right)
-\frac1{\beta^2z^2}\,}\,\mathrm{d}\xi \;,
\label{nu-g}
\end{equation}
where $\nu^{-2}(z)$ is short for $(\nu^{-1}(z))^2$, i.e.\ the square of the
inverse of $\nu$, and $\nu^3(z)$ for the cube $(\nu(z))^3$.

In the case of the Euclidean amoeba metric, $\nu(z)=\sqrt{1+z^2}$, the
expression \eqref{nu-g} simplifies to
\begin{equation}
g(z) \equiv
g_2(z) = \frac{1}{1+\beta^2z^2}\;,
\label{pmdiffusivity}
\end{equation}
which is, up to the substitution $\lambda=1/\beta$, the Perona-Malik
diffusivity \cite{Perona-PAMI90} that is also one of the common choices
for $g$ in the self-snakes equation.

When using the $L^1$ amoeba metric, $\nu(z)=1+z$, the integral in \eqref{nu-g}
can be numerically evaluated, and one obtains an edge-stopping function
$g(s)\equiv g_1(s)$
that differs from \eqref{pmdiffusivity} in that it decreases away from
$g(0)=1$ already with nonvanishing negative slope, thus reacting more
sensitive to even small image contrasts.

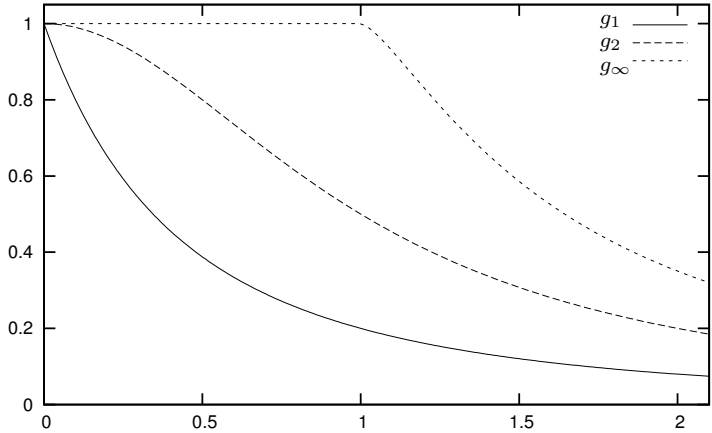
\begin{figure}[t]
\unitlength0.01\textwidth
\begin{picture} (63, 38)
\put(-2, 0.0){\includegraphics[width=65\unitlength]{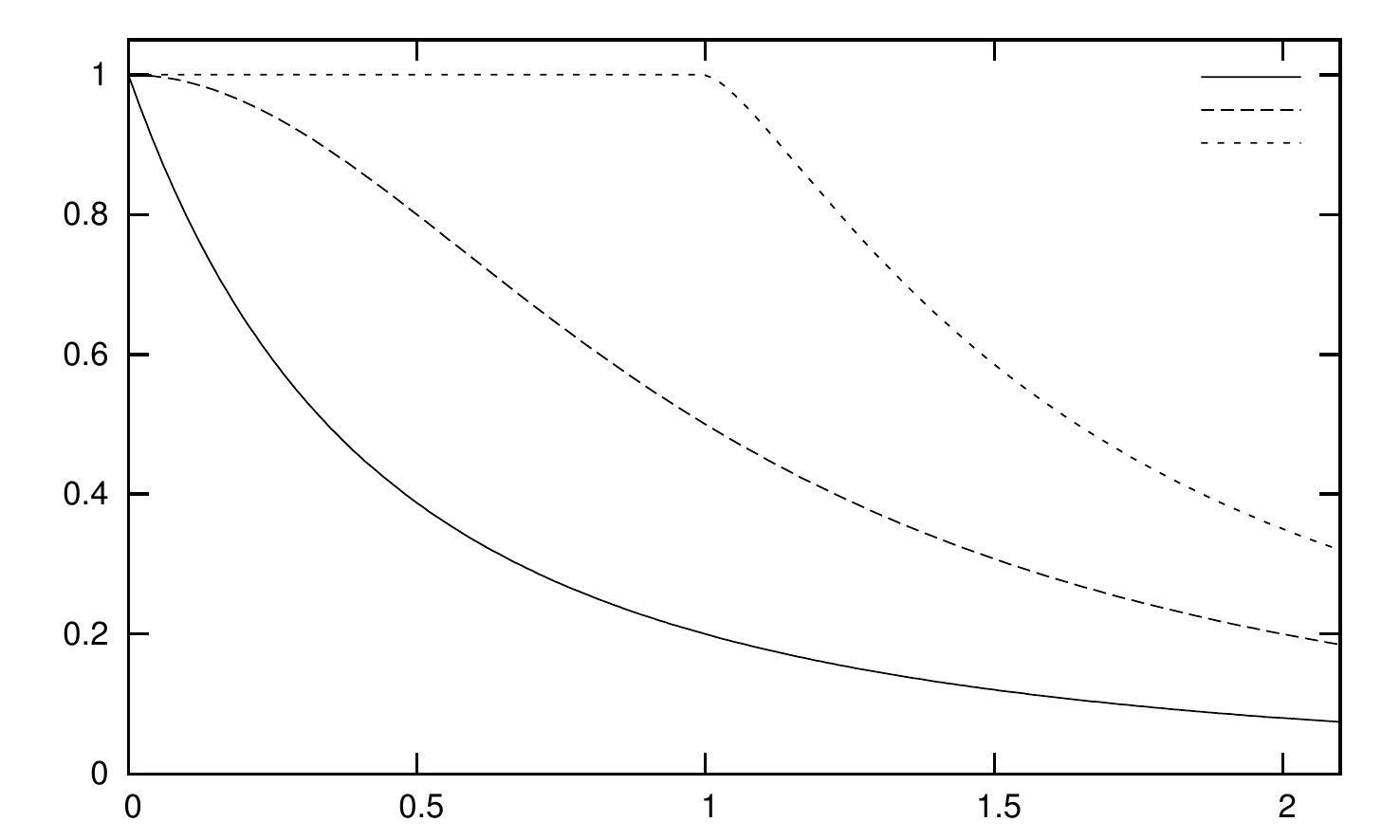}}
\put(51,35.5){$\scriptstyle g_1$}
\put(51,33.5){$\scriptstyle g_2$}
\put(51,31.5){$\scriptstyle g_\infty$}
\end{picture}
\caption{\label{fig:edgestoppers}Edge-stopping functions $g_1$, $g_2$ and
$g_\infty$ associated to $L^1$, Euclidean and $L^\infty$ amoeba metrics,
respectively. Throughout these metrics, the contrast scale $\beta$ has been
set to $1$.}
\end{figure}

Finally, for the $L^\infty$ amoeba metric, $\nu(z)=\max\{1,z\}$, it is again
possible to state $g$ in closed form,
\begin{equation}
g(z) \equiv
g_\infty(z) = \begin{cases} 1\;, & \beta z\le 1\;, \\
1-\left(1-\frac1{\beta^2z^2}\right)^{3/2}\;, & \beta z>1
\end{cases}
\label{g-linf}
\end{equation}
which shows that $g_\infty$ is completely insensitive to image contrasts up
to $z=1/\beta$ and then starts decreasing with a kink.
All three edge-stopping functions are depicted in
Figure~\ref{fig:edgestoppers}.

\subsection{Dilation and Erosion}
\label{subsec:dilero}

The two most fundamental operations of mathematical morphology,
dilation and erosion, use as aggregation step the maximum and minimum
of intensities, respectively. This can naturally be done also in
combination with an amoeba-based pixel selection step.

We point out that the standard dilation of an image $u$
with fixed structuring element $S$ can be written as
\begin{equation}
(u\oplus S)(i) = \max\limits_{j\in i+S} u(j) 
= \max\limits_{j\in\varOmega} \bigl(u(j)+\omega^-_S(i-j)\bigr)\;,
\label{dil1}
\end{equation}
where $\omega^-_S$ denotes the function
\begin{equation}
\omega^-_S(k) = \begin{cases} 0\;,&-k\in S\;,\\-\infty\;,&\text{else.}\end{cases}
\end{equation}
The last term in \eqref{dil1} allows an interesting interpretation in
terms of the max-plus algebra \cite{Baccelli-Book92,Quadrat-icm95}, an
algebraic structure on $\bbbr\cup\{+\infty,-\infty\}$
in which the maximum operation takes the role of
addition in the usual algebra of real numbers, while addition takes the
role of multiplication. It is evident that \eqref{dil1} is nothing
else but a convolution of $u$ and $\omega^-_S$
in the max-plus algebra, see \cite{Maragos-JMIV05}.

In writing erosion in an analogous way, we follow a convention frequently
used in the literature by using instead of the structuring element $S$
the conjugate structuring element $S^*$,
which comes down geometrically to a point reflection on the origin, $S^*=-S$.
The advantage of this convention is that
subsequent definitions like those for opening and closing become simpler
\cite{Heijmans-CVGIP90}, compare Section~\ref{subsec:opclo}.

Defining then $\omega^+_{S^*}$ as zero on $S$, but $+\infty$ outside,
erosion is stated as
\begin{equation}
(u\ominus S)(i) = \min\limits_{j\in j+S^*} u(j)
= \min\limits_{j\in\varOmega} \bigl(u(j)+\omega^+_{S^*}(i-j)\bigr) 
= \min\limits_{j\in\varOmega} \bigl(u(j)+\omega^+_{S}(j-i)\bigr) 
\;,
\label{ero1}
\end{equation}
which can be interpreted again as a convolution of $u$ and $\omega^+_{S^*}$
in the min-plus algebra \cite{Maragos-JMIV05}.

Abandoning the fixed window and using a family
$\mathcal{S}:=\{i\mapsto S(i)~|~i\in\varOmega\}$
of structuring elements $S(i)$ located at pixel $i$,
one can write amoeba dilation as
\begin{gather}
(u\oplus \mathcal{S})(i)
= \max\limits_{j\in\varOmega} \bigl(u(j)+\omega^-_{\mathcal{S}}(i,j)\bigr)\;,
\label{dil2}
\\
\omega^-_{\mathcal{S}}(i,j) = \begin{cases} 0\;,&j\in S(i)\;,\\
-\infty\;,&\text{else.}\end{cases}
\end{gather}
Just as the last term in \eqref{dil1}
is a max-plus convolution, the right-hand side \eqref{dil2} is the
max-plus analogon
of a (discretised) integral operator. Herein,
$\omega^-_{\mathcal{S}}(i,j)$ acts as the max-plus counterpart of just
the same type of integral kernel that appears as point-spread
function in space-variant image deconvolution models.

Similarly, amoeba erosion becomes a min-plus integral operator with a
min-plus kernel
$\omega^+_{\mathcal{S}^*}(i,j)\equiv\omega^+_{\mathcal{S}}(j,i)$.
Generally, conjugate structuring elements in the space-variant case are
given by
\begin{equation}
S^*(i) = \{ j\in\varOmega~|~ i\in S(j)\}\;.
\end{equation}
Interestingly, if $\mathcal{S}$ is made up by amoebas
$S(i)\equiv\mathcal{A}_\varrho(i)$,
there is no difference whether the conjugate structuring elements
$\mathcal{S}^*$ or standard structuring elements $\mathcal{S}$ are used in
erosion: property
\eqref{mutu} of the amoebas entails
$\omega^{\pm}_{\mathcal{S}}(j,i)=\omega^{\pm}_{\mathcal{S}}(i,j)$ for all
$i,j\in\varOmega$, or equivalently
\begin{equation}
\mathcal{A}_\varrho^*(i)\equiv\mathcal{A}_\varrho(i) \;.
\label{amoe-selfconj}
\end{equation}
We will denote this property as \emph{self-conjugacy} of amoebas.

\begin{figure}[t]
\includegraphics[width=0.30\textwidth]{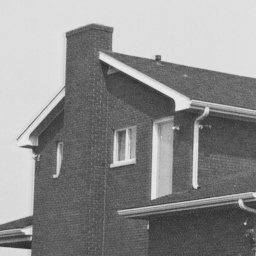}
\caption{\label{fig:morphinit}Grey-scale image ($256\times256$ pixels)
used to demonstrate non-adaptive and amoeba-based morphological filters.}
\end{figure}

\begin{figure}[t]
\unitlength0.01\textwidth
\begin{picture}(100,23.5)
\put( 0,0)
{\includegraphics[width=23.5\unitlength]{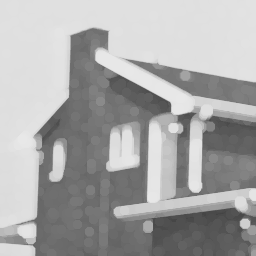}}
\put(25.5,0)
{\includegraphics[width=23.5\unitlength]{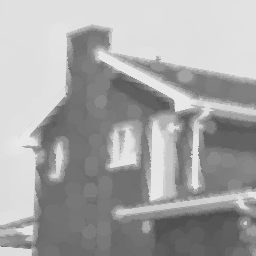}}
\put(51, 0)
{\includegraphics[width=23.5\unitlength]{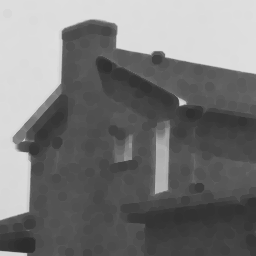}}
\put(76.5, 0)
{\includegraphics[width=23.5\unitlength]{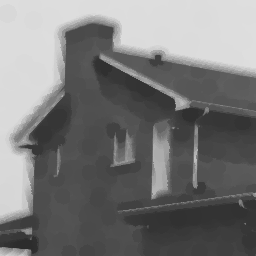}}
\put( 1.1,2){\colorbox{white}{\rule{0pt}{.7em}\hbox to.7em{\kern.1em\smash{a}}}}
\put(26.6,2){\colorbox{white}{\rule{0pt}{.7em}\hbox to.7em{\kern.1em\smash{b}}}}
\put(52.1,2){\colorbox{white}{\rule{0pt}{.7em}\hbox to.7em{\kern.1em\smash{c}}}}
\put(77.6,2){\colorbox{white}{\rule{0pt}{.7em}\hbox to.7em{\kern.1em\smash{d}}}}
\end{picture}
\caption{\label{fig:dilero}Morphological dilation and erosion, 
non-adaptive and amoeba-based, of the test image from 
Figure~\ref{fig:morphinit}.
\textbf{(a)} Non-adaptive morphological dilation 
with disk of radius $\varrho=5$ as structuring element. --
\textbf{(b)} Amoeba dilation 
with Euclidean amoeba metric, $\beta=0.1$, $\varrho=10$. --
\textbf{(c)} Non-adaptive morphological erosion
with structuring element as in (a). --
\textbf{(d)} Amoeba erosion 
with amoeba parameters as in (b).
}
\end{figure}
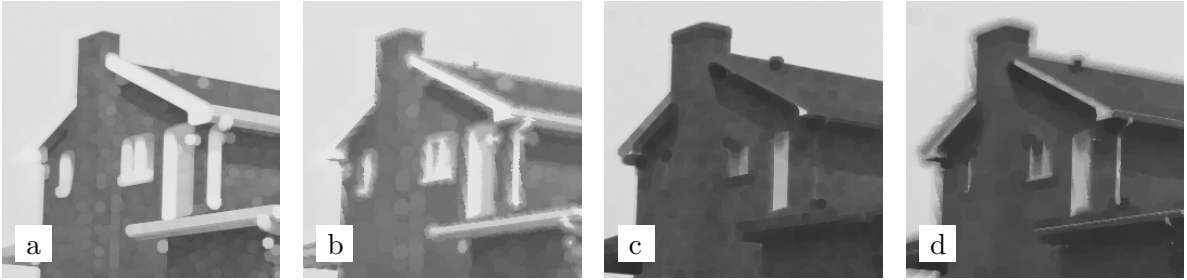

Figure~\ref{fig:dilero}
shows the results of non-adaptive and amoeba dilation and erosion of
an example image depicted in Figure~\ref{fig:morphinit}.
Non-adaptive as well as amoeba-based
dilation extend bright image details, but it can be seen that the
spreading of bright image parts is stopped at strong edges; similarly
for the propagation of dark details by erosion.

\subsubsection{PDE Approximation}

It is a well-known fact that Hamilton-Jacobi PDEs
\begin{equation}
u_t=\pm\lvert\bm{\nabla}u\vert
\label{hamjac-basecase}
\end{equation}
describe dilation (``$+$'' case) and
erosion (``$-$'') of continuous-scale images or level-set functions $u$
in the sense that evolution of an initial image $u(t=0)=f$ by
\eqref{hamjac-basecase} up to time $T=\varrho$ yields the dilation or erosion
of $f$ with a Euclidean ball-shaped structuring element of radius $\varrho$.
It can therefore be expected that amoeba dilation and erosion, too, should
be related to hyperbolic PDEs resembling \eqref{hamjac-basecase}. The following
result from \cite{Welk-Aiep14} confirms this intuition.

\begin{theorem}[\cite{Welk-Aiep14}]
For a smooth function $u:\varOmega\to\bbbr$, one step of amoeba dilation
or amoeba erosion with amoeba radius $\varrho$ and Euclidean amoeba metric
approximates for $\varrho\to0$
a time step of size $\tau=\varrho$ of an explicit time discretisation of
the Hamilton-Jacobi-type PDE
\begin{equation}
u_t=\pm\frac{\lvert\bm{\nabla}u\rvert}{\sqrt{1+\beta^2\,\lvert\bm{\nabla}u
\rvert^2}} \;,
\label{hamjac-amoeba}
\end{equation}
where the ``$+$'' sign applies for dilation, and ``$-$'' for erosion.
\end{theorem}

The proof of this result can be found in \cite{Welk-Aiep14}; it is based on
Proof Strategy I from Section~\ref{subsubsec:iamfpdeproof}.

Note that unlike in Theorem~\ref{thm:iamfpde} the time step size here
depends linearly, not quadratically, on $\varrho$. In \cite{Welk-Aiep14}
the theorem is formulated slightly more general to cover also amoeba
$\alpha$-quantile filters that interpolate in a natural way between
median filtering ($\alpha=1/2$), dilation ($\alpha=1$) and erosion
($\alpha=0$). As a result of the different order of decay of $\tau$
for $\varrho\to0$, it comes as no surprise that for
$\alpha\ne1/2$ always the advection behaviour of the Hamilton-Jacobi equation
\eqref{hamjac-amoeba} dominates over the parabolic equation \eqref{mcm},
thus turning quantile filters into ``slower'' approximations to the
same PDE.

\subsection{Opening and Closing}
\label{subsec:opclo}

In mathematical morphology, the opening of an image $f$ with (fixed)
structuring element $S$ is defined as the concatenation of an erosion
followed by a dilation with $S$. In case $S$ is not point-symmetric
it is essential that, as mentioned in Section~\ref{subsec:dilero},
the conjugate structuring element $S^*$ is used in the erosion step.
Opening therefore reads as
\begin{equation}
(f\circ S)(i) = \bigl((f\ominus S)\oplus S\bigr)(i) 
= 
\max\limits_{j\in\varOmega}~
\min\limits_{k\in\varOmega}~
\bigl(f(k) + \omega_{S^*}^+(j-k) + \omega_{S}^-(i-j)\bigr)\;.
\end{equation}
Analogously, closing is defined as dilation followed by erosion,
\begin{equation}
(f\bullet S)(i) = \bigl((f\oplus S)\ominus S\bigr)(i) 
= 
\min\limits_{j\in\varOmega}~
\max\limits_{k\in\varOmega}~
\bigl(f(k) + \omega_{S}^-(j-k) + \omega_{S^*}^+(i-j)\bigr)\;.
\end{equation}

Again, it is straightforward to turn these operations into adaptive
variants by using amoeba structuring elements.
Amoeba opening and closing of image
$f$ with amoebas of radius $\varrho$ are given as
\begin{align}
f\circ \mathcal{S}_\varrho(f) &= 
\bigl(f\ominus \mathcal{S}_\varrho(f)\bigr)\oplus \mathcal{S}_\varrho(f)\;,\\
f\bullet \mathcal{S}_\varrho(f) &= 
\bigl(f\oplus \mathcal{S}_\varrho(f)\bigr) \ominus \mathcal{S}_\varrho(f)
\end{align}
where $\mathcal{S}_\varrho(f)=
\{i \mapsto \mathcal{A}_\varrho(f;i)~|~i\in\varOmega\}$.

It is worth noticing that the difficulty about using the conjugate set of
structuring elements for erosion disappears here due to the
self-conjugacy \eqref{amoe-selfconj} of the amoeba structuring element set.

As it is essential to
use the same set of structuring elements in the dilation and erosion step,
both steps must be carried out with the amoebas obtained from the original
image. The underlying principle
is that in the second step (dilation for opening or erosion
for closing) each pixel should influence exactly those pixels which have
influenced it in the first step before.
As a consequence, e.g.\ amoeba opening is not exactly the same as amoeba
erosion followed by amoeba dilation -- this sequence would be understood by
default as recalculating amoebas after the erosion step, i.e.\
\begin{align}
\bigl(f\ominus \mathcal{S}_\varrho(f)\bigr)\oplus 
\mathcal{S}_\varrho\bigl(f\ominus\mathcal{S}_\varrho(f)\bigr)\;,
\end{align}
which is inappropriate for an opening operation.

\begin{figure}[t]
\unitlength0.01\textwidth
\begin{picture}(100,23.5)
\put( 0,0)
{\includegraphics[width=23.5\unitlength]{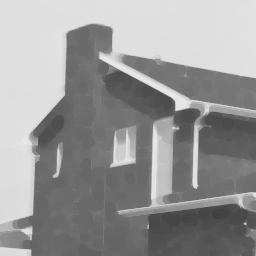}}
\put(25.5,0)
{\includegraphics[width=23.5\unitlength]{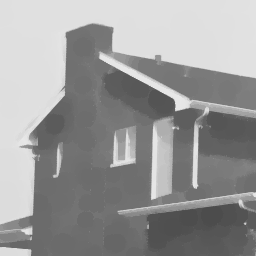}}
\put(51, 0)
{\includegraphics[width=23.5\unitlength]{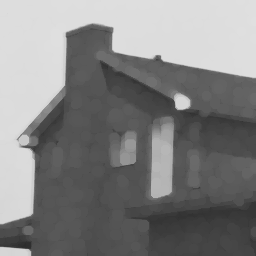}}
\put(76.5, 0)
{\includegraphics[width=23.5\unitlength]{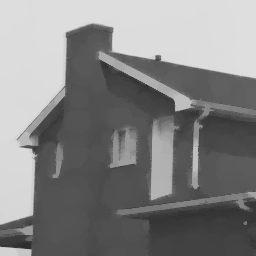}}
\put( 1.1,2){\colorbox{white}{\rule{0pt}{.7em}\hbox to.7em{\kern.1em\smash{a}}}}
\put(26.6,2){\colorbox{white}{\rule{0pt}{.7em}\hbox to.7em{\kern.1em\smash{b}}}}
\put(52.1,2){\colorbox{white}{\rule{0pt}{.7em}\hbox to.7em{\kern.1em\smash{c}}}}
\put(77.6,2){\colorbox{white}{\rule{0pt}{.7em}\hbox to.7em{\kern.1em\smash{d}}}}
\end{picture}
\caption{\label{fig:cloop}Non-adaptive and amoeba-based morphological 
closing and opening applied to the test image from Figure~\ref{fig:dilero}.
\textbf{(a)} Non-adaptive closing
with disk-shaped structuring element of radius $\varrho=5$. --
\textbf{(b)} Amoeba closing
with Euclidean amoeba metric, $\beta=0.1$, $\varrho=10$. --
\textbf{(c)} Non-adaptive opening
with structuring element as in (a). --
\textbf{(d)} Amoeba opening
with amoeba parameters as in (b).
}
\end{figure}
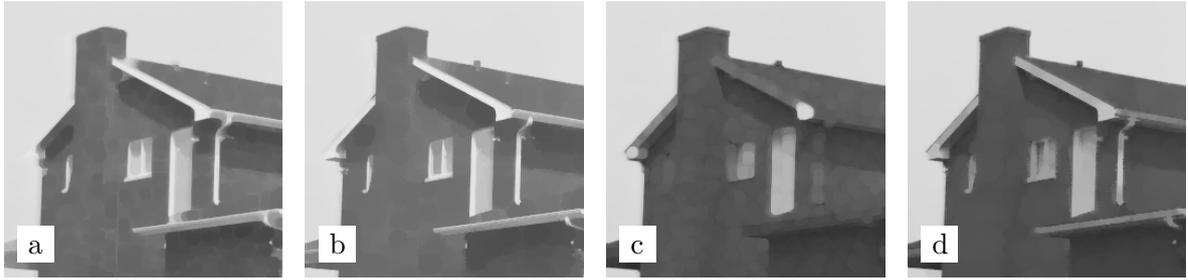

In Figure~\ref{fig:cloop} exemplary results of non-adaptive and amoeba-based
closing and opening of the test image from Figure~\ref{fig:morphinit}
are shown. Like its non-adaptive counterparts, amoeba-based closing and
opening remove small-scale dark or bright details, respectively. However,
the amoeba versions do this in a less aggressive way. Extended narrow
structures that are often removed partially by the non-adaptive filters
are more often preserved as a whole, with reduced contrast,
or removed completely by the amoeba filters, see e.g.\ the roof front
edge descending to the right from the chimney, and the acute roof corner
separating it from the sky.

\subsubsection{Opening and Closing Scale Spaces and PDEs}

The association between median, dilation and erosion filters and PDEs is
inherently related to the scale space structures of these filters,
compare \cite{Heijmans-JVCIR02}.
All of these filters form an additive semi-group in the sense that
iterative application of the same filter yields an increasing filter effect
that naturally adds up over iteration numbers.
In the case of dilation and erosion iteration numbers are also in linear
relation with increasing structuring element size, as dilating an initial
image $n$ times with (non-adaptive) structuring element radius $\varrho$ is
equivalent to dilating once with radius $n\varrho$.
Such an additive semi-group structure perfectly matches initial value problems
for PDEs in which, too, evolution times add up.

While opening and closing, too, have a scale space structure, their semi-group
operation is not additive but \emph{supremal} as it is based on taking the
maximum of parameters.
For example, repeating the same opening or closing operation on a given image
just reproduces the result of the first application of the filter (i.e.,
opening and closing operators are idempotent); and concatenating two openings
or two closings with structuring element radii $\varrho_1$, $\varrho_2$ gives
an opening or closing with radius $\max\{\varrho_1,\varrho_2\}$.

For this reason, also amoeba opening and closing are not associated with PDE
evolutions in the same way as the previous filters. Possible relations to
PDE-based filters may be considered in future research.

\section{Grey-Scale Segmentation}
\label{sec:seg}

Following established terminology, image segmentation denotes the
task to decompose a given image into regions that are in the one or other
way homogeneous in themselves but different from each other, with the intention
that these regions are meaningful in that they are associated to objects being
depicted. Intensity-based segmentation uses intensity as the main criterion
of homogeneity within and dissimilarity between segments. Specialising to
the case of two segments (foreground and background) with the additional
geometric hypothesis that segments are separated by sharp and smooth contours,
contour-based segmentation approaches based on curve or level set evolutions
lend themselves as tools for segmentation, with active contours as an
important representative. In this section we show how amoeba algorithms can
be made useful in this context.

Despite the fact that experiments on magnetic resonance data are used to
illustrate the concepts in this section, this is not meant to make a claim
that neither active contour nor the related active region methods (which are
not discussed further here) in their pure form could serve as a
state-of-the-art segmentation method for medical images. In fact,
competitive results in medical image segmentation are nowadays achieved by
complex frameworks that often include active contours and/or active regions
as a component but in combination with additional techniques that allow to
bring in anatomical knowledge such as shape and appearance models
\cite{Cootes-tr01}. An early representative of these frameworks is
\cite{Leventon-cvpr00}, which has been followed by many more since then.
Like classical geodesic active contours, the amoeba active contours presented
in the following could be integrated into this type of framework but this has
not been done so far.

\subsection{Amoeba Active Contours}
\label{subsec:aac}

The standard procedure of an \emph{active contour,} or \emph{snake,} method
starts with some initial
contour which may be obtained automatically from some previous knowledge
or heuristics regarding the position of a sought structure, or from human
operator input. Representing this contour either by a sampled curve or by
a level-set function, it is then evolved up to a given evolution time or
up to a steady state by the action of some parabolic PDE, which is often
derived as a gradient descent of a segmentation energy in the image plane.
An important representative are
geodesic active contours (GAC) \cite{Caselles-iccv95,Kichenassamy-iccv95}.
Their segmentation energy is essentially a curve length measure of the
contour in a modified metric on the image plane that favours placing the
contour in high-contrast locations. The PDE for GAC in level-set representation
reads
\begin{equation}
u_t = \lvert\bm{\nabla}u\rvert \,\mathrm{div}\left(
g\bigl(\lvert\bm{\nabla}f\rvert\bigr)\,
\frac{\bm{\nabla}u}{\lvert\bm{\nabla}u\rvert}\right)
\;.
\label{gac}
\end{equation}
Herein, $u$ is the evolving level-set function in the plane that represents
the actual evolving contour as one of its level sets (by default, the
zero-level set), and $f$ is the invariable image being segmented.
The similarity of \eqref{gac} to self-snakes \eqref{ssn} (which were
actually inspired from active contours, thus the name) together with the link
between amoeba median filtering and self snakes established by
Theorem~\ref{thm:iamfpde} suggest that an amoeba median approach could be
used to evolve the level set function $u$ instead of equation \eqref{gac}.

Introduced in \cite{Welk-ssvm11}, the resulting
\emph{amoeba active contour (AAC)} algorithm proceeds as follows:

\begin{enumerate}
\item Compute amoeba structuring elements based on the input image $f$.
\item Initialise the evolving level-set function $u$ to represent
the initial contour.
\item Evolve the image $u$ by median filtering with the amoebas from
Step~1 as structuring elements.
\end{enumerate}

Results from this algorithm look qualitatively fairly similar to those from
GAC, as will also be demonstrated later in this section.

\subsection{PDE Approximation}
\label{subsec:aacpde}

In order to study the relation between AAC and GAC, it makes sense again
to consider a space-continuous model and
to investigate the PDE approximated by AAC in the case of vanishing amoeba
radius. The following result was proven in \cite{Welk-JMIV15}. Note that
in this theorem the contrast scale parameter $\beta$ is fixed to $1$ for
simplicity, which, however, is no restriction of the result because in the
active contour setting in question, the case $\beta\ne1$ is easily mapped
to $\beta=1$ by just scaling the intensities of image $f$ by $\beta$.

\begin{theorem}[\cite{Welk-JMIV15}]
\label{thm:aacpde}
Let a smooth level-set function $u$ be filtered by amoeba median filtering,
where the amoebas are generated from a smooth image $f$. Assume that
the amoeba metric is given by
\eqref{ametcont}, \eqref{ametphi} with $\beta=1$.
One step of this filter for $u$ then approximates for $\varrho\to0$
a time step size of size $\tau=\varrho^2/6$
of an explicit time discretisation of the PDE
\begin{align}
u_t &= 
G \, u_{\bm{\xi\xi}} - \lvert\bm{\nabla}u\rvert\cdot
\bigl(H_1\, f_{\bm{\chi\chi}} + 2 H_2\, f_{\bm{\chi\eta}} 
+ H_3\, f_{\bm{\eta\eta}}\bigr) 
\label{aacpde}
\end{align}
with the coefficients given by
\begin{align}
&G \equiv G\bigl(\lvert\bm{\nabla}f\rvert,\alpha)
= \frac{1}{\nu\bigl(\lvert\bm{\nabla}f\rvert\sin\alpha\bigr)^2}\;,
\label{aacpdeG}
\\
&\!\!\begin{pmatrix}H_1&H_2\\H_2&H_3\end{pmatrix}
\equiv \begin{pmatrix}
H_1\bigl(\lvert\bm{\nabla}f\rvert,\alpha)~~&
H_2\bigl(\lvert\bm{\nabla}f\rvert,\alpha)\\
H_2\bigl(\lvert\bm{\nabla}f\rvert,\alpha)~~&
H_3\bigl(\lvert\bm{\nabla}f\rvert,\alpha)
\end{pmatrix}
\notag\\*
&~~{}= \frac32\,\nu\bigl(\lvert\bm{\nabla}f\rvert\sin\alpha\bigr)
\int\limits_{\alpha-\pi/2}^{\alpha+\pi/2}
\frac{\nu'\bigl(\lvert\bm{\nabla}f\rvert\sin\vartheta\bigr)}
{\nu\bigl(\lvert\bm{\nabla}f\rvert\sin\vartheta\bigr)^4}
\begin{pmatrix}\cos^2\vartheta&\sin\vartheta\cos\vartheta\\
\sin\vartheta\cos\vartheta&\cos^2\vartheta\end{pmatrix}
\,\mathrm{d}\vartheta \;.
\label{aacpdeH}
\end{align}
Here, $\bm{\eta}=\bm{\nabla}u/\lvert\bm{\nabla}u\rvert$ and
$\bm{\xi}\perp\bm{\eta}$ are unit vectors in gradient and level line
direction, respectively, for $u$, whereas
$\bm{\chi}=\bm{\nabla}f/\lvert\bm{\nabla}f\rvert$ and
$\bm{\zeta}\perp\bm{\chi}$ are the corresponding unit vectors for $f$,
and $\alpha=\angle(\bm{\eta},\bm{\chi})$ is the angle between
both gradient directions.
\end{theorem}

The proof of this result is found for the case of the Euclidean amoeba metric
in \cite{Welk-ssvm13}, and for general amoeba metric in \cite{Welk-JMIV15}.
It relies on Proof Strategy II from Section~\ref{subsubsec:iamfpdeproof}.

An attempt to analyse AAC using Proof Strategy I had been made in
\cite{Welk-ssvm11}, where, however, only a special case was successfully
treated: The theorem proven in \cite{Welk-ssvm11} states that AAC approximates
the GAC equation \eqref{gac} if image $f$ and level set function $u$ are
rotationally symmetric about the same centre.

In fact, the rotational symmetry hypothesis can be weakened; what is needed
for \eqref{aacpde}--\eqref{aacpdeH} to reduce to the exact GAC equation is
actually, whenever
$\alpha=0$ (thus,
$\eta=\chi$, $\xi=\zeta$),
$u_{\bm{\xi\eta}}=f_{\bm{\xi\eta}}=0$ and
$u_{\bm{\xi\xi}}/\lvert\bm{\nabla}u\rvert
=f_{\bm{\xi\xi}}/\lvert\bm{\nabla}f\rvert$
hold, \eqref{aacpde}--\eqref{aacpdeH} boil down to the GAC equation
\eqref{gac}.

At first glance, this is still a very artificial choice; however,
looking at the geometrical implications of this setting, one sees that
it means that the level lines of $u$ are
aligned to those of $f$, have the same curvature, and the image
contrast in both $f$ and $u$ does not change along these level lines.
Thereby the hypothesis of this special case is well approximated in the
near-convergence stage of a segmentation process when the
object---background contrast is more or less uniform along the contour.

As a consequence, the coincidence of AAC and GAC in this
case justifies that both approaches can expected
to yield very similar types of segmentations. The convergence behaviour
towards these segmentations may differ more; a closer comparison of
both PDEs in \cite{Welk-ssvm13,Welk-JMIV15} based on typical
geometric configurations indicates that the amoeba active contour PDE
drives contours toward image contours in a more pronounced way.

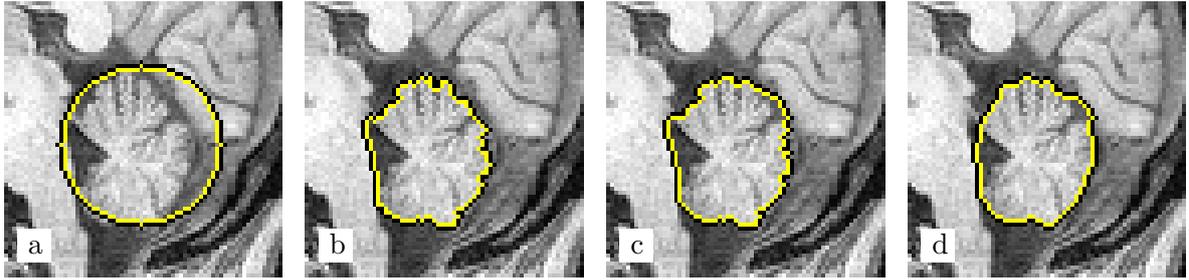
\begin{figure}[t]
\unitlength0.01\textwidth
\begin{picture}(100, 23.5)
\put( 0.0,0)
{\includegraphics[width=23.5\unitlength]{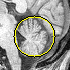}}
\put(25.5,0)
{\includegraphics[width=23.5\unitlength]{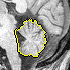}}
\put(51.0,0)
{\includegraphics[width=23.5\unitlength]{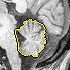}}
\put(76.5,0)
{\includegraphics[width=23.5\unitlength]{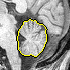}}
\put( 1.1,2){\colorbox{white}{\rule{0pt}{.6em}\hbox to.6em{\kern.1em\smash{a}}}}
\put(26.6,2){\colorbox{white}{\rule{0pt}{.6em}\hbox to.6em{\kern.1em\smash{b}}}}
\put(52.1,2){\colorbox{white}{\rule{0pt}{.6em}\hbox to.6em{\kern.1em\smash{c}}}}
\put(77.6,2){\colorbox{white}{\rule{0pt}{.6em}\hbox to.6em{\kern.1em\smash{d}}}}
\end{picture}
\caption{\label{fig:cerebellum}
Amoeba and geodesic active contour segmentation.
\textbf{(a)} Detail ($70\times70$ pixels)
from an MR slice of a human brain with
initial contour enclosing the cerebellum. --
\textbf{(b)} Amoeba active contours with Euclidean amoeba metric,
$\beta=0.1$, $\varrho=12$, 10 iterations. --
\textbf{(c)} Amoeba active contours with $L^1$ amoeba metric,
$\beta=0.1$, $\varrho=12$, 60 iterations. --
\textbf{(d)} Geodesic active contours with Perona-Malik edge-stopping
function, $\lambda=10$, 960 iterations of explicit scheme with
time step size $\tau=0.25$. --
From \cite{Welk-ssvm11,Welk-JMIV15}.
}
\end{figure}

Figure~\ref{fig:cerebellum} presents an example that confirms the
overall similarity between amoeba and geodesic active contours but
also the tendency of AAC to adapt more precise to very small-scaled
edge details. Frame (a) shows the original image with an initial contour
roughly enclosing the cerebellum. Frames (b) and (c) demonstrate
segmentation by AAC with Euclidean and $L^1$ amoeba metrics, respectively,
while Frame (d) shows a GAC result for comparison.

\subsection{Force Terms}
\label{sec:force}

Geodesic active contours in their basic form \eqref{gac}
suffer from some limitations. First of all, when initialised with a
contour enclosing a large area with one or several small objects inside,
the active contour process spends plenty of evolution time to slowly
move the contour inwards until it hits an object boundary, due to the
initially small curvature of the contour.
Secondly, for pronounced concave object geometries, the process tends to
lock in at undesired local minima that detect well some convex contour
parts but short-cut concave parts via straight line segments. Similar
problems can occur when segmenting multiple objects within one initial
contour, see the examples in \cite{Kichenassamy-ARMA96}. Thirdly,
as the basic curvature motion process involves only inward movement
of contours, it is generally not possible with \eqref{gac} to segment
objects from initial contours inside the object, which is sometimes
desirable in applications.
Due to their similarity to GAC, amoeba active contours share these problems.

A common remedy for these problems in the literature on
active contour segmentation is the introduction of a \emph{force
term}. Its typical form is $\pm\gamma\lvert\bm{\nabla}u\rvert$, i.e.\
essentially the right-hand side of a Hamilton-Jacobi PDE for
dilation or erosion, compare \eqref{hamjac-basecase}.
An erosion force accelerates the inward
motion of the contour; it allows to get past homogeneous
areas faster, and helps the contour to find concave object boundaries
and to separate multiple objects. By a dilation force it is possible
to push the contour evolution in outward direction, which makes it
possible to use initial contours inside objects.

In both cases, however, the force strength needs careful adjustment
because dilation or erosion may also push the contour evolution
across object boundaries, thereby preventing their detection.

In \cite{Cohen-CVGIPIU91} where this modification was proposed first
(by the name of ``balloon force''), $\gamma$ was chosen as constant,
but the possibility to steer it contrast-dependent, was mentioned.
This has been done in
\cite{Caselles-iccv95,Kichenassamy-ARMA96,Malladi-PAMI95} by
modulating the force term in a geodesic active contour model
with the same edge-stopping function $g$, such that the entire force
term reads as
$\pm \gamma\,g(\lvert\bm{\nabla}f\rvert)\,\lvert\bm{\nabla}u\rvert$
with constant $\gamma$.

The relation between amoeba quantile filters and
Hamilton-Jacobi PDEs mentioned in Section~\ref{subsec:dilero}
indicates how to achieve a similar modification in the amoeba
active contour algorithm: the median filter step should be
biased, basically by replacing the median with some quantile.
The most obvious way to do this is to use the $\alpha$-quantile
with a fixed $\alpha\ne1/2$. Within a discrete amoeba containing
$p$ pixels, this means to choose the value ranked
$\alpha p$ in the ordered sequence of intensities.
However, taking into account that the amoeba size $p$ (or the
amoeba area in the continuous setting) varies even for fixed
$\varrho$ with local image contrast, it is not
less natural to think of $\alpha$ as varying with the amoeba size.
If one chooses $\alpha-1/2$ inversely proportional to the amoeba
size, this comes down to modify the median with a fixed rank offset
$b$, such that in an amoeba of $p$ pixels one would choose the
intensity value with rank $p/2+b$. These two variants of the
AAC algorithm have been proposed in \cite{Welk-ssvm11}.
In \cite{Welk-JMIV15} a third variant (``quadratic bias'')
was introduced which
chooses from the rank order the element with index $p/2+r\,p^2$
with fixed $r$.
For these three scenarios, further analysis was provided in
\cite{Welk-JMIV15}, based on the Euclidean amoeba metric.
We summarise the results here.

\paragraph{Fixed offset bias.}

Choosing the entry at position
$p/2+b$ from the rank order approximates a force
term $+\gamma_b \,\lvert \bm{\nabla}u\rvert\,
\nu(\lvert\bm{\nabla}f\rvert\,\sin\alpha)$ with $\gamma_b\sim b$.
Note that in the symmetric case in which the PDE approximated by AAC
coincides with the GAC equation this becomes exactly the
``balloon force'' term with constant dilation/erosion weight from
\cite{Cohen-CVGIPIU91}.

\paragraph{Quantile bias.}

Choosing the element with index $p/2+q p$
from the rank order within each amoeba
approximates a force term
$+\gamma_q\,\lvert\bm{\nabla}u\rvert\,\sqrt{(1+\lvert\bm{\nabla}f
\rvert^2\sin^2\alpha)/(1+\lvert\bm{\nabla}f\rvert^2)}$
with $\gamma_q\sim q$. In the rotationally symmetric case this
term lies between the constant weight of \cite{Cohen-CVGIPIU91}
and the $g$-weight from \cite{Kichenassamy-ARMA96}.

\paragraph{Quadratic bias.}

Choosing the entry at index $p/2+r\,p^2$
from the rank order of intensities
yields an approximated force term $+\gamma_r\,\lvert\bm{\nabla}u\rvert
\,\nu(\lvert\bm{\nabla}f\rvert)/\nu(\lvert\bm{\nabla}f\rvert)^2$.
In the rotationally symmetric case this corresponds to the
$g$-weight from \cite{Kichenassamy-ARMA96}.

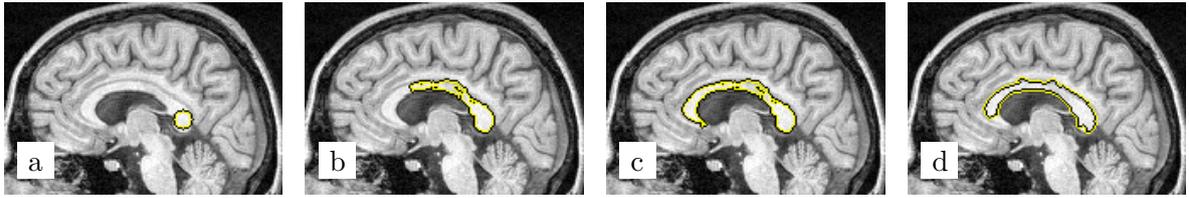
\begin{figure}[t!]
\unitlength0.01\textwidth
\begin{picture}(100, 16.4)
\put( 0.0,0)
{\includegraphics[width=23.5\unitlength]{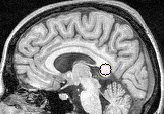}}
\put(25.5,0)
{\includegraphics[width=23.5\unitlength]{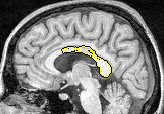}}
\put(51.0,0)
{\includegraphics[width=23.5\unitlength]{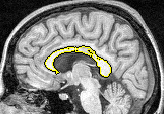}}
\put(76.5,0)
{\includegraphics[width=23.5\unitlength]{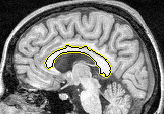}}
\put( 1.1,2){\colorbox{white}{\rule{0pt}{.7em}\hbox to.7em{\kern.1em\smash{a}}}}
\put(26.6,2){\colorbox{white}{\rule{0pt}{.7em}\hbox to.7em{\kern.1em\smash{b}}}}
\put(52.1,2){\colorbox{white}{\rule{0pt}{.7em}\hbox to.7em{\kern.1em\smash{c}}}}
\put(77.6,2){\colorbox{white}{\rule{0pt}{.7em}\hbox to.7em{\kern.1em\smash{d}}}}
\end{picture}
\caption{\label{fig:cc}Segmentation with initialisation inside the
sought object by amoeba and geodesic active contours with dilation
force. 
\textbf{(a)} Detail ($164\times114$ pixels)
from an MR slice of human brain with initial
contour placed inside the corpus callosum. --
\textbf{(b)} Amoeba active contour evolution with Euclidean amoeba
metric, $\beta=2$, $\varrho=20$, fixed offset bias $b=10$, and
20 iterations. --
\textbf{(c)} Same as in (b) but 35 iterations.
\textbf{(d)} Geodesic active contours with Perona-Malik edge-stopping
function, $\lambda=0.5$, dilation force $\gamma=-0.16$ (multiplied
with the edge-stopping function) and erosion force
$\gamma_{\mathrm{c}}=5\times10^{-4}$ (independent of the
edge-stopping function), explicit scheme with time step size
$\tau=0.25$, 18,960,000 iterations. --
From \cite{Welk-JMIV15}.
}
\end{figure}

To illustrate amoeba active contours with bias, Fig.~\ref{fig:cc}
presents an example (shortened from \cite{Welk-JMIV15}).
Frame (a) is a test image with initial contour inside a
mostly homogeneous object (the corpus callosum). Fig.~\ref{fig:cc}(b) and
(c) then show contours computed by amoeba active contours with fixed
offset bias for two different evolution times, one intermediate, one
displaying the final segmentation. For comparison, a segmentation with
geodesic active contours is shown in (d).

We remark that in the AAC examples, a few pixels within the corpus
callosum region are excluded from the segment, see the small isolated
contour loops there. This is not a numerical artifact but a result from
the precise adaption of amoebas to image structures even up to the
resolution limit (pixel precision) of the image -- the pixels not included
in the segment are noise pixels with intensities significantly deviating
from the neighbourhood, which are simply not included in any amoeba of
outside pixels.
Modifications like presmoothing input images can be applied to avoid this.
On the contrary, the absence of such difficulties in the
GAC example is a beneficial effect of the otherwise often undesirable
numerical blurring effect of the finite-difference scheme.

\section{Pre-Smoothing in Self-Snakes and Amoeba Filters}
\label{sec:presmoothing}

The approximation result of Theorem~\ref{thm:iamfpde} associates
iterated amoeba median filtering with the self-snakes equation \eqref{ssn}.
Unlike (mean) curvature motion \eqref{mcm}, self-snakes possess
edge-enhancing properties.
Rewriting \eqref{ssn} by the product rule, one can state the self-snakes
process as
\begin{equation}
u_t = 
g\bigl(\lvert\bm{\nabla}u\rvert\bigr)\, 
\lvert\bm{\nabla}u\rvert \,
\mathrm{div}\left(\frac{\bm{\nabla}u}{\lvert\bm{\nabla}u\rvert}\right)
+ \langle \bm{\nabla}g,\bm{\nabla}u\rangle
\label{ssn-rewrite}
\end{equation}
in which the
first summand is just a curvature motion process modulated by $g$,
whereas the second, advective, term is responsible for the edge-enhancing
behaviour. Unfortunately, this term has a shock-filter property
which makes not only its numerical treatment difficult -- in finite
difference schemes usually an upwind discretisation will be required to
approximate it -- but even entails ill-posedness of the PDE itself that
is reflected in a noticeable staircasing behaviour.
Indeed, as demonstrated by an experiment in \cite{Welk-JMIV11}, the result
of a numerical computation of a self-snakes evolution differs significantly
if the underlying grid resolution is changed.

A common remedy to this ill-posed behaviour is to use pre-smoothing in
the argument of the edge-stopping function, i.e.\ to replace
$g(\lvert\bm{\nabla}u\rvert)$ in \eqref{ssn} or \eqref{ssn-rewrite} by
$g(\lvert\bm{\nabla}u_\sigma\rvert)$ where $u_\sigma$ is the result of
convolving $u$ with a Gaussian of standard deviation $\sigma$. Thereby,
the ill-posedness of self-snakes is removed, and a stable filtering achieved,
at the cost of the additional smoothing-scale parameter $\sigma$.

In this section, we deal with the question whether this staircasing phenomenon
has also an analogue in the amoeba median filtering context, and what is an
appropriate counterpart for the pre-smoothing modification on the amoeba
side. This is done by quantitative analysis of a synthetic example, the
first part of which has been published before in
\cite{Welk-ssvm13,Welk-Aiep14}.

\subsection{Pre-Smoothing in Amoeba Median Filtering, and Amoeba Radius}

First of all, notice that a straightforward translation of the pre-smoothing
procedure to the amoeba median filtering context is to use $u_\sigma$
in place of $u$ when computing the structuring elements in an amoeba
median filtering step. This is actually an instance of the generalised
amoeba median filtering procedure of the amoeba active contour setting,
Sections~\ref{subsec:aac} and \ref{subsec:aacpde}, such that the PDE
approximation result from Theorem~\ref{thm:aacpde} can be applied to see
that it would approximate a PDE which is not identical to the standard
self-snakes with pre-smoothing, but closely related to it.

At second glance, however, it can be questioned whether the introduction
of the smoothing-scale parameter $\sigma$ into the amoeba median filter
is necessary.
Unlike finite-difference schemes for self-snakes, amoeba filtering by
construction already involves a very similar smoothing-scale parameter,
namely, the amoeba radius $\varrho$. One can conjecture that the
positive $\varrho$ necessarily used in any amoeba computation could already
provide a pre-smoothing effect similar to the Gaussian convolution in the
PDE setting. This conjecture will be investigated in the following.

\subsection{Perturbation Analysis of Test Cases}
\label{ssec:perturb}

The starting point for constructing the test cases is a simple slope
function that would be stationary under both self-snakes and amoeba
median filter evolutions, see Figure~\ref{fig:perturbschematic}(a).
From this slope, described by the function $u_0:\bbbr^2\to\bbbr$,
$u_0(x,y)=x$, test cases are derived by adding small
single-frequency oscillations such as
$\varepsilon\cos\langle\bm{k},\bm{x}\rangle$ with frequency vectors
$\bm{k}$.

Given the nonlinear nature of the filters under investigation, there
is no superposition property for the effects of different
perturbations of $u_0$. Nevertheless, interactions between
$u_0$ and the perturbations are always of higher order
$\mathcal{O}(\varepsilon^2)$, such that the analysis of the
first-order effects of perturbations still gives a useful intuition
about the behaviour of the filters.

\subsubsection{Test Case 1: Gradient-Aligned Oscillation}

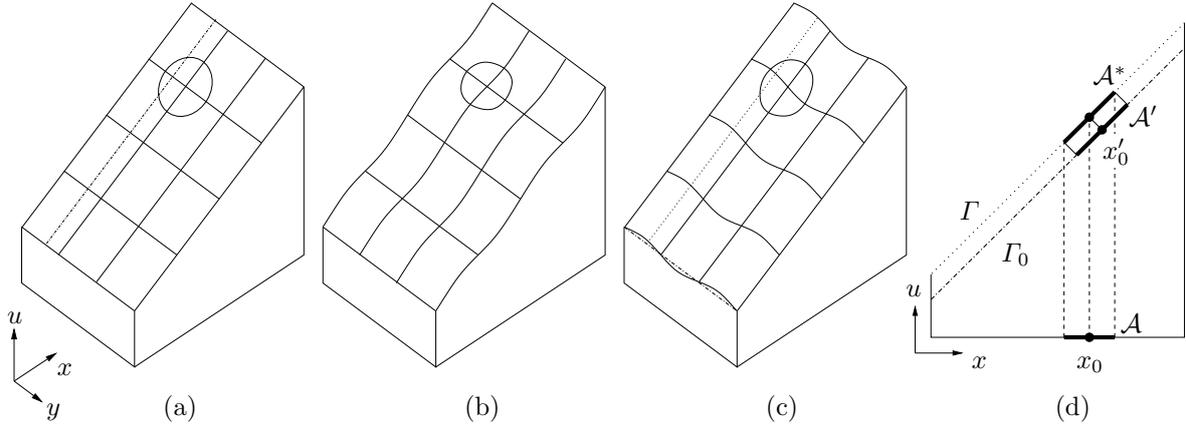
\begin{figure}[t!]
\unitlength0.001\textwidth
\begin{picture}(1000,360)
\put(  3, 10){\includegraphics[width=40\unitlength]{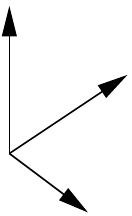}}
\put( 33,  0){\small$y$}
\put( 42, 34){\small$x$}
\put(  0, 78){\small$u$}
\put( 11, 40){\includegraphics[width=240\unitlength]{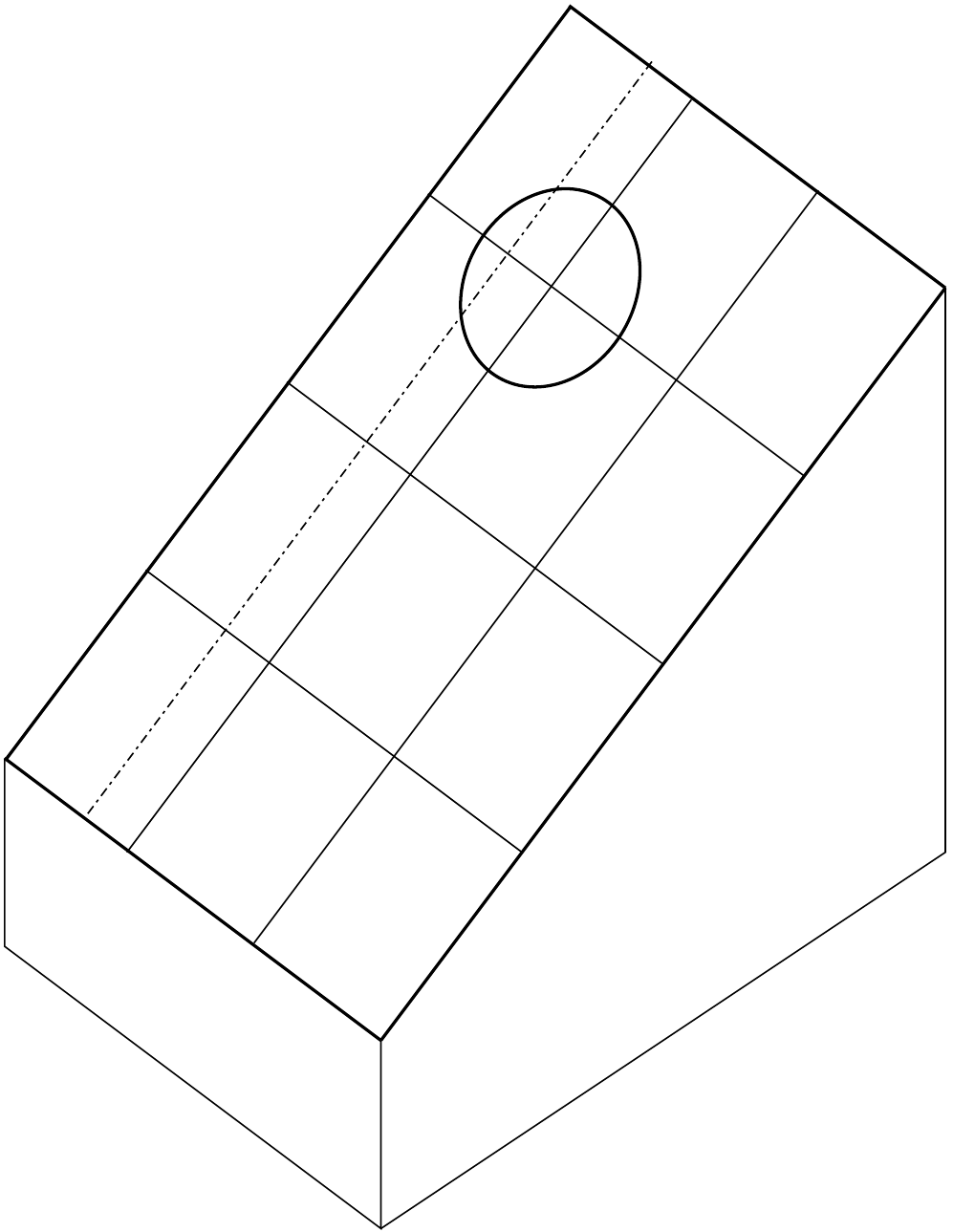}}
\put(131,  0){\small (a)}
\put(264, 40){\includegraphics[width=240\unitlength]{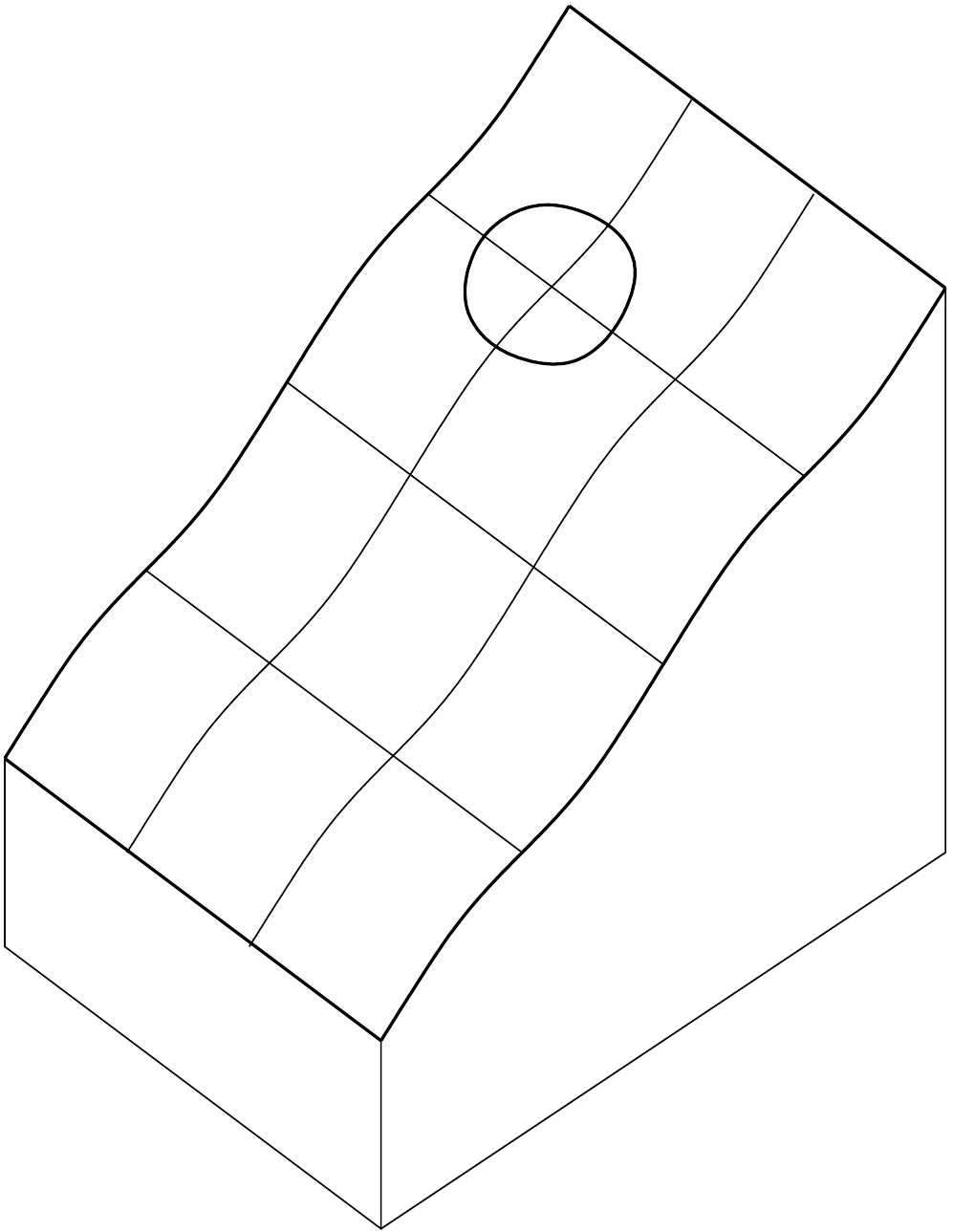}}
\put(384,  0){\small (b)}
\put(517, 40){\includegraphics[width=240\unitlength]{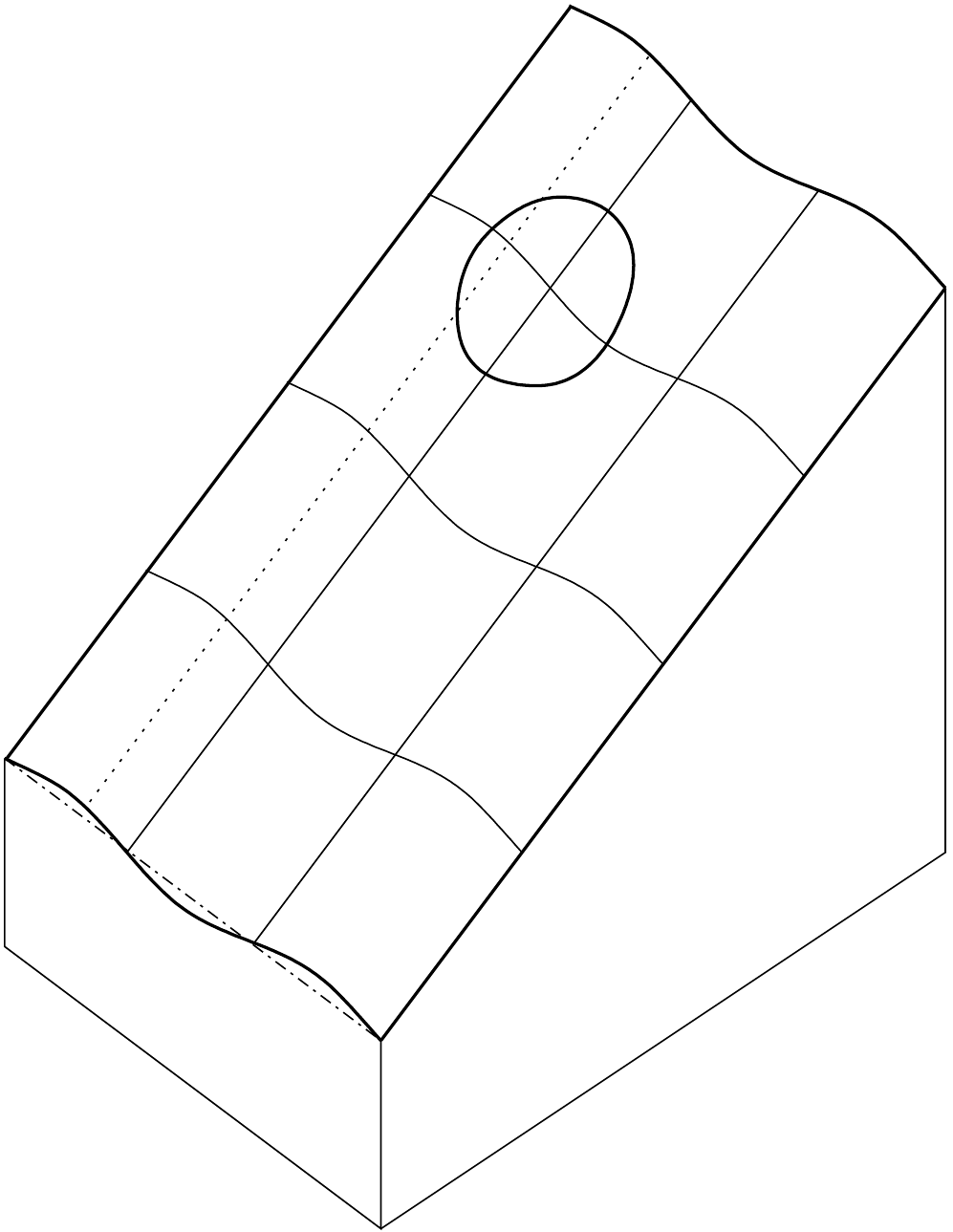}}
\put(637,  0){\small (c)}
\put(760, 50){\includegraphics[width=230\unitlength]{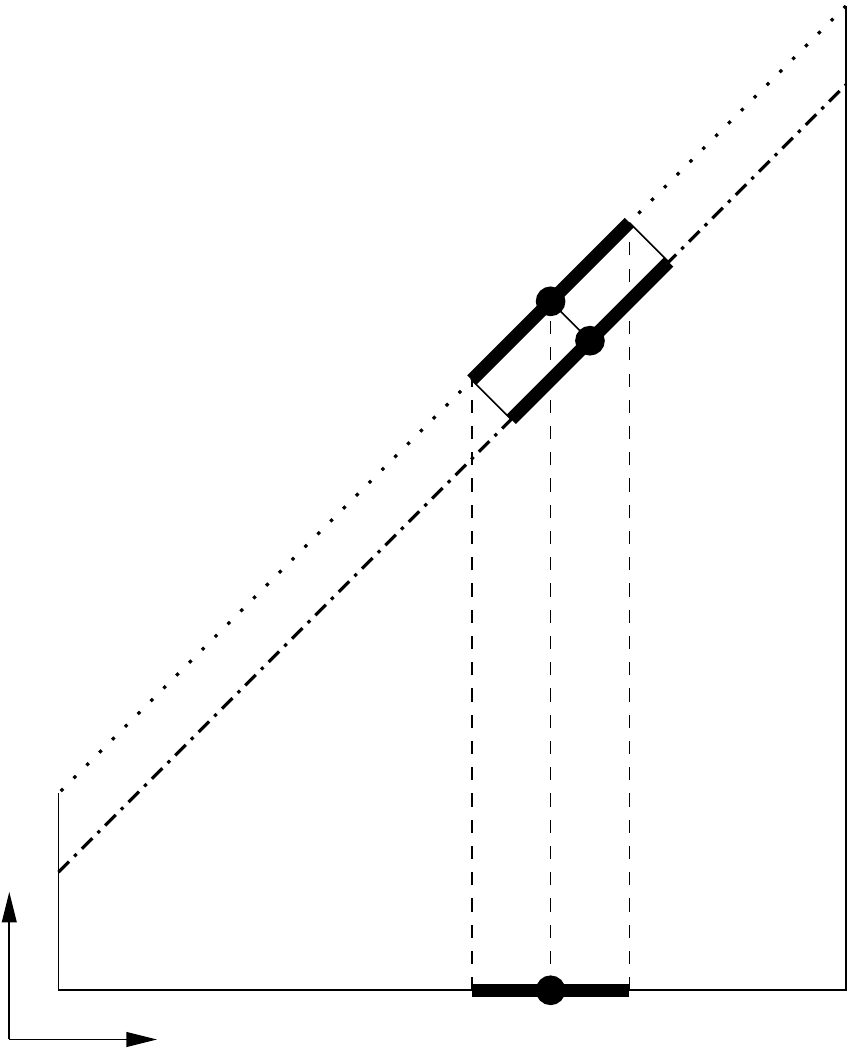}}
\put(810, 40){\small$x$}
\put(755,100){\small$u$}
\put(935, 70){\small$\mathcal{A}$}
\put(898, 40){\small$x_0$}
\put(912,280){\small$\mathcal{A}^*$}
\put(942,245){\small$\mathcal{A}'$}
\put(925,210){\color{white}{\rule{15\unitlength}{28\unitlength}}}
\put(920,217){\small$x_0'$}
\put(800,165){\small$\varGamma$}
\put(835,130){\small$\varGamma_0$}
\put(880,  0){\small (d)}
\end{picture}
\caption{\label{fig:perturbschematic}Schematic representation of
example functions used in the perturbation analysis,
Section~\ref{ssec:perturb}. \textbf{(a)} Graph $\varGamma_0$
of unperturbed function $u_0=x$, with a Euclidean $\varrho$-disk
whose projection to the $x$-$y$ plane yields an amoeba. --
\textbf{(b)} Graph $\varGamma$
of a function $u$ of type \eqref{perturb1} including
a gradient-aligned perturbation. --
\textbf{(c)} Graph $\varGamma$
of a function $u$ of type \eqref{perturb2} including
a level-line-aligned perturbation. --
\textbf{(d)} Cut in $x$ direction through the graph $\varGamma$
from (c) and the unperturbed graph $\varGamma_0$ from (a).
The sketch includes further
the amoeba $\mathcal{A}$ around $(x_0,y_0)$, the corresponding
Euclidean disk $\mathcal{A}^*$ on $\varGamma$ and the projection
$\mathcal{A}'$ of $\mathcal{A}^*$ to $\varGamma_0$ which is
centred at $(x_0',y_0)$.
}
\end{figure}

For the first test case, see \cite{Welk-ssvm13,Welk-Aiep14},
the perturbation frequency is aligned with the
gradient direction, $\bm{k}=(k,0)$, yielding the
input signal schematically depicted in
Figure~\ref{fig:perturbschematic}(b),
\begin{equation}
u(x,y) = x + \varepsilon\cos(kx)\;, \quad \varepsilon <\!\!< 1\;.
\label{perturb1}
\end{equation}

\paragraph{Self-Snakes Analysis.}

To determine the response of the self-snakes evolution
\eqref{ssn-rewrite} to the perturbed signal \eqref{perturb1}, notice
first that level lines
of \eqref{perturb1} are straight and parallel, such that one has
$\mathrm{div}(\bm{\nabla}u/\lvert\bm{\nabla}u\rvert)\equiv0$ and
$\langle\bm{\nabla}g,\bm{\nabla}u\rangle=g_xu_x$.
Further, one has $u_x=1-\varepsilon\,k\sin(kx)$ and
$g_x=\varepsilon\,k^2\cos(kx)/2+\mathcal{O}(\varepsilon^2)$, finally
turning \eqref{ssn-rewrite} into
\begin{equation}
u_t = g_xu_x = \frac12\,k^2\varepsilon\cos(kx)+\mathcal{O}(\varepsilon^2)\;.
\label{perturb1-ssn}
\end{equation}
From this it can be read off that a frequency response factor $k^2/2$ occurs
that grows indefinitely for high frequencies. Since the
nonlinearity of \eqref{ssn-rewrite} instantaneously spreads out the single
perturbation frequency $k$ to higher harmonics, arbitrarily high amplification
appears already within short evolution time, and the regularity of the
evolving function is lost. This explains the stair-casing behaviour of
self-snakes without pre-smoothing.

Using pre-smoothed $u_\sigma$ in the edge-stopping function argument, one
has instead $\partial_xu_\sigma=x+\varepsilon\,\exp(-k^2\sigma^2/2)\,\cos(kx)$,
$g_x=k^2\varepsilon\,\exp(-k^2\sigma^2/2)\,\cos(kx)/2$ and therefore
\begin{equation}
u_t = \frac12\,k^2\varepsilon\,\exp\left(-\frac{k^2\sigma^2}2\right)
\cos(kx) + \mathcal{O}(\varepsilon^2)\;,
\label{perturb1-ssnps}
\end{equation}
with the frequency response factor
$k^2\exp(-k^2\sigma^2/2)/2$ that is globally
bounded with its maximum at $k=\sqrt2/\sigma$. Therefore, pre-smoothing
ensures that the regularity of the evolving function is maintained.

\paragraph{Amoeba Filter Analysis.}

To analyse the effect of amoeba median filtering (with Euclidean amoeba
metric) on the function \eqref{perturb1}, consider
an amoeba of amoeba radius $\varrho$ around $(x_0,y_0)$, and assume that
the contrast scale is chosen as $\beta=1$.

The median of $u$ within that amoeba can be expressed via an integral
formula, see \cite{Welk-ssvm13,Welk-Aiep14}, which can be numerically
evaluated to be approximately equal to
$u(x_0,y_0)+\delta(k)\cdot\varepsilon\cos(kx_0)$ with a
frequency response factor $\delta(k)$.
In other words, one amoeba median filter step amplifies
the perturbation $u-u_0$ of \eqref{perturb1} versus $u_0(x,y)=x$
by the amplification factor $\lambda(k):=1+\delta(k)$.

\begin{figure}[t!]
\unitlength0.001\textwidth
\begin{picture}(1000,640)
\put( 67,376){\includegraphics[width=423\unitlength]
      {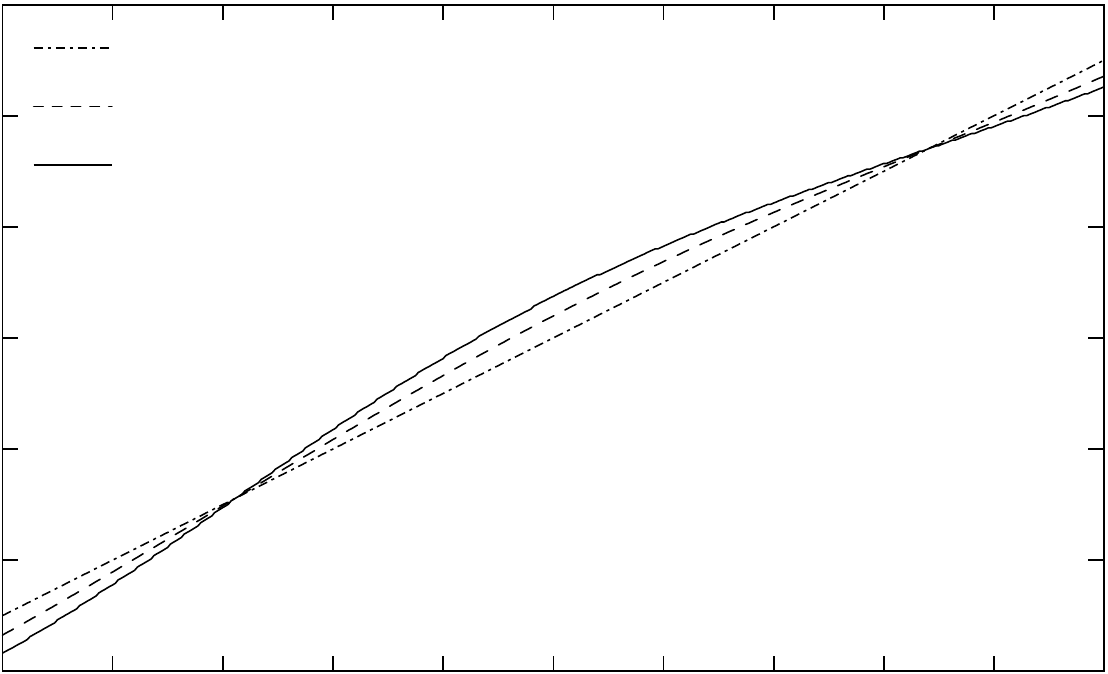}}
\put( 93,350){\small$4.6$}
\put(177,350){\small$4.8$}
\put(261,350){\small$5.0$}
\put(345,350){\small$5.2$}
\put(429,350){\small$5.4$}
\put(250,326){\small$k=5$}
\put( 30,368){\small$4.4$}
\put( 30,410){\small$4.6$}
\put( 30,452){\small$4.8$}
\put( 30,494){\small$5.0$}
\put( 30,536){\small$5.2$}
\put( 30,578){\small$5.4$}
\put( 30,620){\small$5.6$}
\put(120,609){\small$u_0(x)=x$}
\put(120,585){\small$u(x)=x+0.04\cos(5x)$}
\put(120,561){\small$v(x)$ (amoeba filtered)}
\put(577,376){\includegraphics[width=423\unitlength]
      {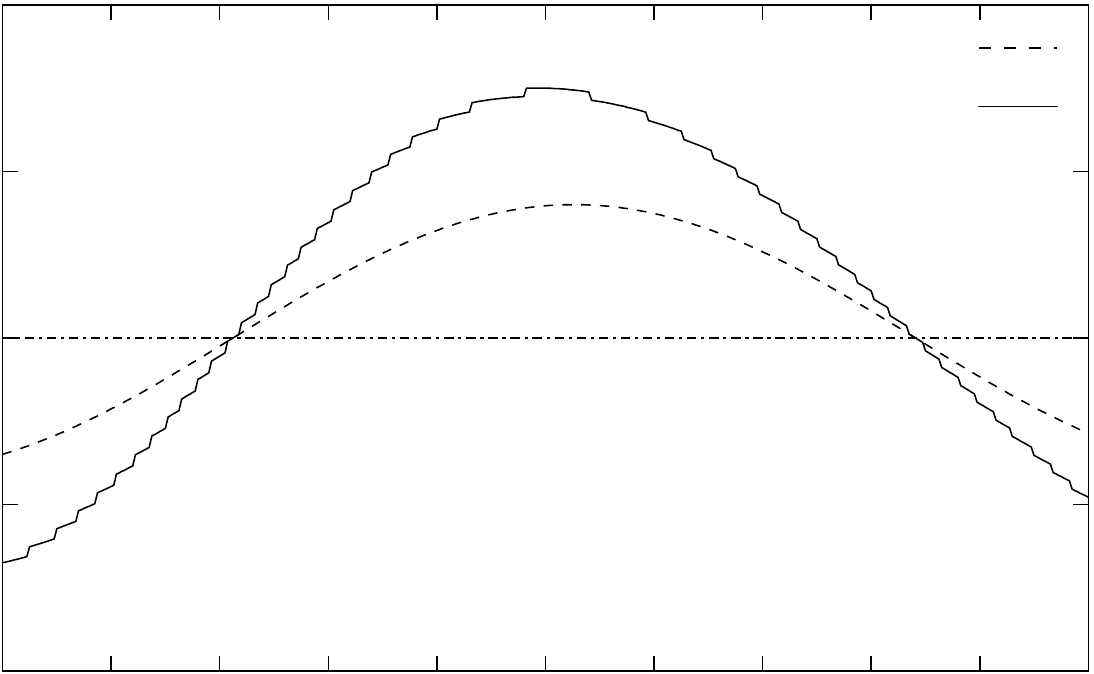}}
\put(603,350){\small$4.6$}
\put(687,350){\small$4.8$}
\put(771,350){\small$5.0$}
\put(855,350){\small$5.2$}
\put(939,350){\small$5.4$}
\put(627,326){\small$k=5$, amplification factor $1.85$}
\put(505,368){\small$-0.10$}
\put(505,431){\small$-0.05$}
\put(525,494){\small$0.00$}
\put(525,557){\small$0.05$}
\put(525,620){\small$0.10$}
\put(885,609){\small$u-u_0$}
\put(885,585){\small$v-u_0$}
\put( 67, 56){\includegraphics[width=423\unitlength]
      {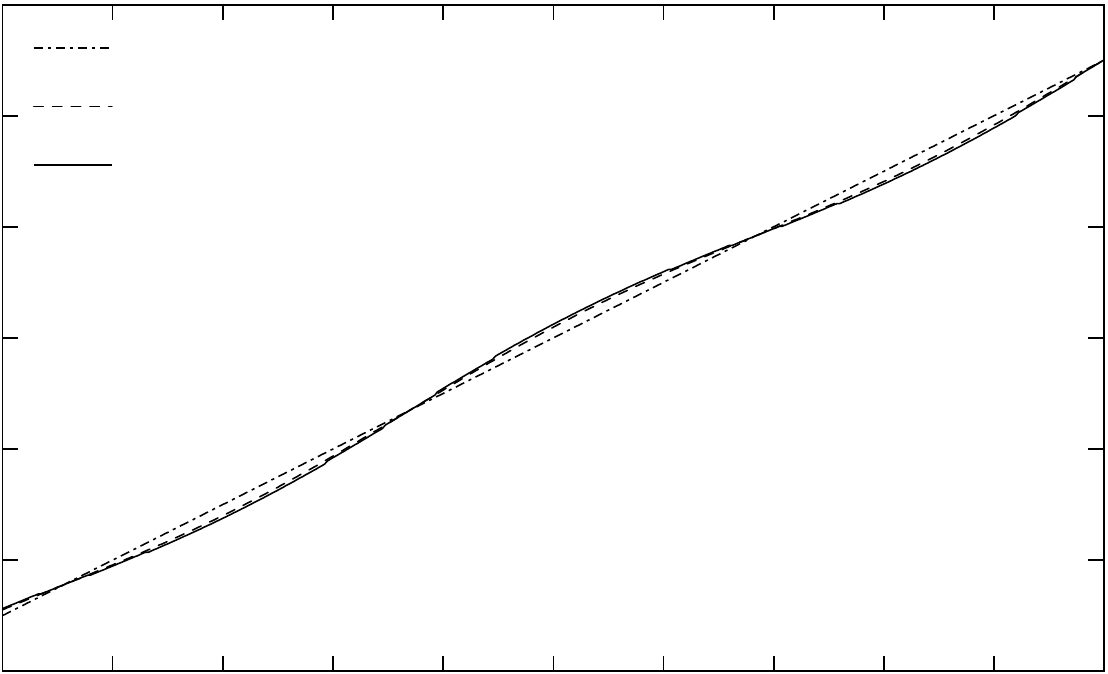}}
\put( 93, 30){\small$4.6$}
\put(177, 30){\small$4.8$}
\put(261, 30){\small$5.0$}
\put(345, 30){\small$5.2$}
\put(429, 30){\small$5.4$}
\put(245,  6){\small$k=10$}
\put( 30, 48){\small$4.4$}
\put( 30, 90){\small$4.6$}
\put( 30,132){\small$4.8$}
\put( 30,174){\small$5.0$}
\put( 30,216){\small$5.2$}
\put( 30,258){\small$5.4$}
\put( 30,300){\small$5.6$}
\put(120,289){\small$u_0(x)=x$}
\put(120,265){\small$u(x)=x+0.02\cos(10x)$}
\put(120,241){\small$v(x)$ (amoeba filtered)}
\put(577, 56){\includegraphics[width=423\unitlength]
      {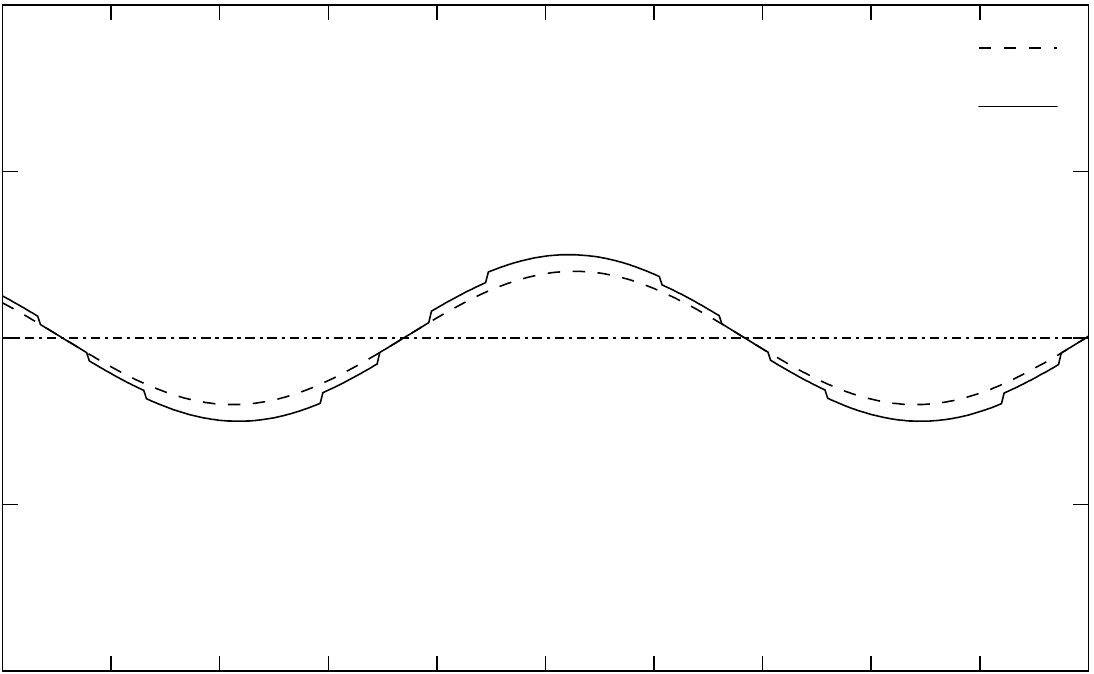}}
\put(603, 30){\small$4.6$}
\put(687, 30){\small$4.8$}
\put(771, 30){\small$5.0$}
\put(855, 30){\small$5.2$}
\put(939, 30){\small$5.4$}
\put(622,  6){\small$k=10$, amplification factor $1.27$}
\put(505, 48){\small$-0.10$}
\put(505,111){\small$-0.05$}
\put(525,174){\small$0.00$}
\put(525,237){\small$0.05$}
\put(525,300){\small$0.10$}
\put(885,289){\small$u-u_0$}
\put(885,265){\small$v-u_0$}
\end{picture}
\caption{\label{fig:perturb1singlefreq}
Numerical computation results for the amplification of a
gradient-aligned perturbation of a linear slope function by one
amoeba median filtering step. Top row shows $k=5$, bottom row $k=10$.
Graphs in left column show unperturbed function $u_0$,
perturbed input function $u$, and filter result $v$; graphs in right column
show perturbations $u-u_0$ and $v-u_0$. Horizontal axes represent $x$,
vertical axes represent function values.
Computations were carried out on a grid with mesh size $0.0025$.}
\end{figure}
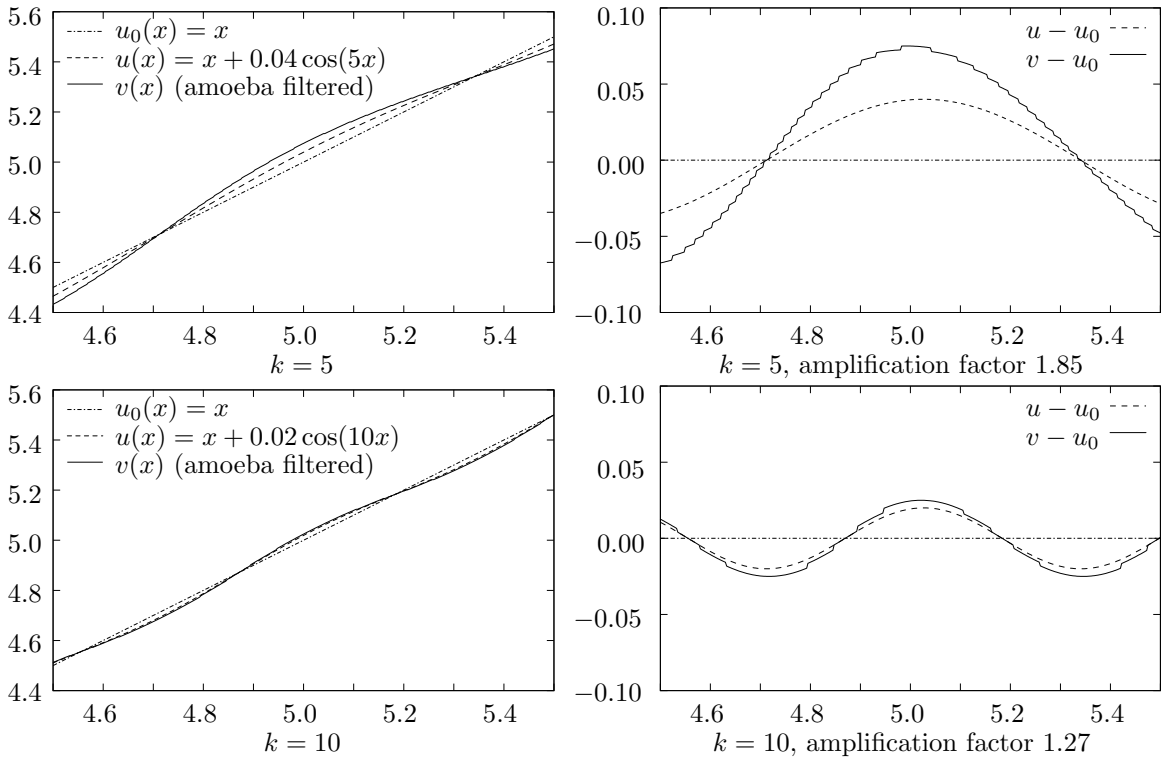

Figure~\ref{fig:perturb1singlefreq} shows results of numerical
approximation of one amoeba median filtering step with $\beta=1$,
$\varrho=1$, on test images of type \eqref{perturb1} with two
different frequencies $k$.
The numerical computation was carried out on a discrete grid
with mesh size $h=0.0025$. For best approximation to the
space-continuous case, amoeba distances between pixels were
computed by numerical integration instead of the Dijkstra search on
the pixel graph. Denoting the filtered image by $v$, numerical
amplification factors can be computed as
$\langle v-u_0,u-u_0\rangle/\langle u-u_0,u-u_0\rangle$ (with the
usual scalar product of functions on a suitable bounded interval);
these are in good accordance with the theoretical result.

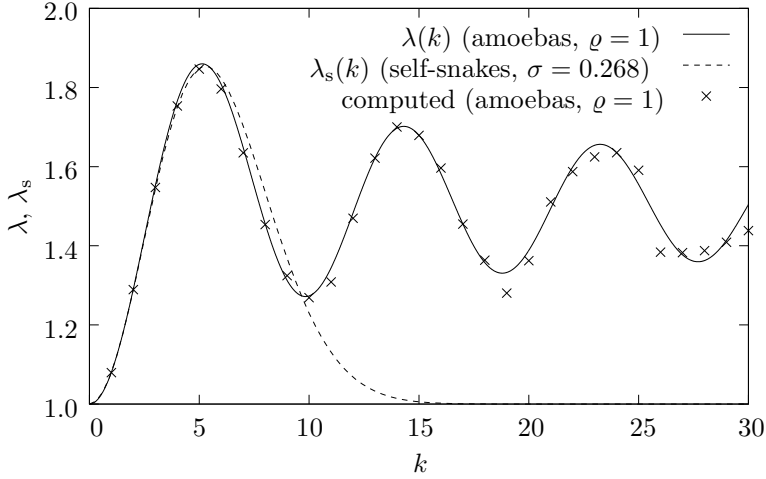
\begin{figure}[t!]
\unitlength0.001\textwidth
\begin{picture}(635,402)
\put( 67, 56){\includegraphics[width=560\unitlength]
      {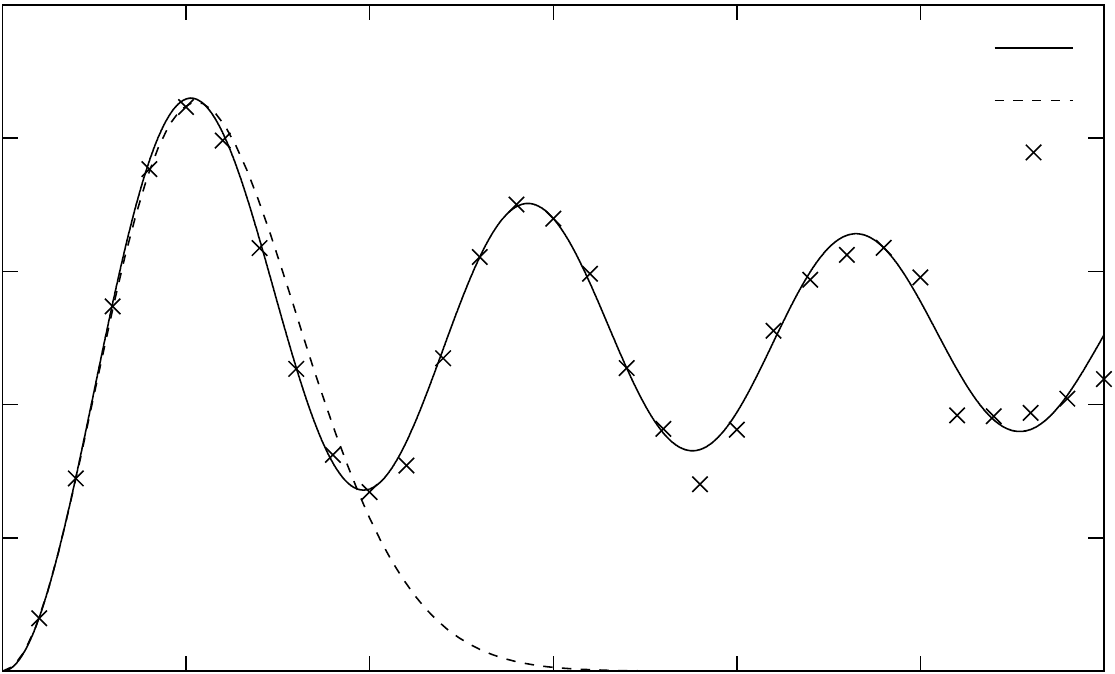}}
\put( 69, 30){\small$0$}
\put(155, 30){\small$5$}
\put(242, 30){\small$10$}
\put(335, 30){\small$15$}
\put(428, 30){\small$20$}
\put(522, 30){\small$25$}
\put(613, 30){\small$30$}
\put(340,  0){\small$k$}
\put( 30, 46){\small$1.0$}
\put( 30,113){\small$1.2$}
\put( 30,180){\small$1.4$}
\put( 30,248){\small$1.6$}
\put( 30,315){\small$1.8$}
\put( 30,382){\small$2.0$}
\put(  0,200){\rotatebox{90}{\small $\lambda$, $\lambda_{\mathrm{s}}$}}
\put(330,360){\small $\lambda(k)$ (amoebas, $\varrho=1$)}
\put(253,334){\small $\lambda_{\mathrm{s}}(k)$ (self-snakes, $\sigma=0.268$)}
\put(280,308){\small computed (amoebas, $\varrho=1$)}
\end{picture}
\caption{\label{fig:perturb1plot}
Amplification of a gradient-aligned perturbation of a linear
slope function by one amoeba median filtering step
(theoretical and numerical values) and a corresponding
time step of an explicit scheme for self-snakes with pre-smoothing.
Adapted and extended from \cite{Welk-ssvm13}.}
\end{figure}

Figure~\ref{fig:perturb1plot} shows the amplification function $\lambda(k)$
for $\varrho=1$
together with its counterpart
$\lambda_{\mathrm{s}}(k):=1+1/6\cdot k^2\exp(-k^2\sigma^2/2)/2$ for
one time step of self-snakes with pre-smoothing, with the time step
size $\varrho^2/6=1/6$ matching the amoeba radius according to
Theorem~\ref{thm:iamfpde}.
The figure also includes numerical
amplification factors for amoeba median filtering with the same
parameters for frequencies $k=1,2,\ldots,30$.
The parameter $\sigma=0.268$ in the self-snakes case has
been chosen for a good match to the first wave of $\delta(k)$.
With this parameter, the amplification behaviour for frequencies up to
approx.\ $10$ is very similar for the pre-smoothed self-snakes equation and
amoeba median filtering. However, for higher frequencies the amplification
factor of pre-smoothed self-snakes rapidly approaches one
(no amplification) whereas it oscillates
around $3/2$ for the amoeba filter.

As a result, oscillations with sufficiently high frequency
are just almost not amplified in the pre-smoothed self-snakes
evolution.
With amoeba median filtering, they are amplified by the globally bounded
factor $\lambda(k)$ in each iteration step.
Whatever $\varepsilon$ was in the initial image $u$ from
\eqref{perturb1}, after a finite number of iterations the oscillations grow
to a level for which the hypothesis $\varepsilon<\!\!<1$ of our analysis is
no longer valid. Even in the space-continuous setting under consideration,
oscillations cannot actually grow infinitely because the median operation
obeys the maximum--minimum principle.

In practice, amoeba filters are computed in a space-discrete setting such that
the effective range of spatial frequencies is limited by the
sampling theorem. For fixed amoeba radius $\varrho=1$ as in
Figure~\ref{fig:perturb1plot}, the relevant range of frequencies is determined
by the mesh size of the pixel grid. If this mesh size is not below approx.\
$\pi/10$, the higher lobes of the amplification
function $\lambda(k)$ that make up the difference to self-snakes with
pre-smoothing do not take effect at all. Translating this to a grid with
mesh size $1$, as common in image processing, this means that for amoeba
radius $\varrho$ up to approx.\ $10/\pi\approx3$ the frequency response
of amoeba median filtering does almost not differ from that of self-snakes
with pre-smoothing.

\subsubsection{Test Case 2: Level-Line-Aligned Oscillation}

To complement the perturbation analysis of gradient-aligned
oscillations, a second test case is considered in which the
perturbation frequency is aligned with the level line direction,
$\bm{k}=(0,k)$. The resulting input signal, compare the
schematic representation in Figure~\ref{fig:perturbschematic}(c),
reads
\begin{equation}
u(x,y) = x + \varepsilon\cos(ky)\;, \quad \varepsilon <\!\!< 1\;.
\label{perturb2}
\end{equation}
This test case was not presented in \cite{Welk-ssvm13,Welk-Aiep14}.
Given that self-snakes act smoothing along level line direction, it can be
expected that this kind of perturbation is dampened by their evolution.
This will be confirmed by the analysis, and the corresponding behaviour
of the amoeba median filter will be stated.

\paragraph{Self-Snakes Analysis}

Unlike for the first test case, gradient directions of $u$ now vary across
the image range, combining constant $u_x=1$ with
$u_y=-k\varepsilon\sin(ky)$. Accordingly, the edge-stopping function
takes the values
\begin{equation}
g(x,y) = \frac1{2+k^2\varepsilon^2\sin^2(ky)}
=\frac12\left(1-\frac{k^2\varepsilon^2}{2}\sin^2(ky)\right)
+\mathcal{O}(\varepsilon^3)
\end{equation}
and thereby $g_x(x,y)=\mathcal{O}(\varepsilon^3)$,
$g_y(x,y)=-k^3\varepsilon^2\sin(ky)\cos(ky)/2
+\mathcal{O}(\varepsilon^3)$.

This leads further to
\begin{align}
\lvert\bm{\nabla}u\rvert&=1+\frac{k^2\varepsilon^2}{2}\sin^2(ky)
+\mathcal{O}(\varepsilon^4)\;,\\
\mathrm{div}\left(\frac{\bm{\nabla}u}{\lvert\bm{\nabla}u\rvert}\right) &=
\partial_x\left(1-\frac{k^2\varepsilon^2}{2}\sin^2(ky)\right)
+\partial_y\bigl(-k\varepsilon\sin(ky)\bigr) +\mathcal{O}(\varepsilon^3)
\notag\\
&= -k^2\varepsilon\cos(ky) +\mathcal{O}(\varepsilon^2)\;,
\\
\langle\bm{\nabla}g,\bm{\nabla}u\rangle 
&= \mathcal{O}(\varepsilon^3)\;,
\end{align}
thus after inserting into \eqref{ssn-rewrite}
\begin{equation}
u_t = -\frac12k^2\varepsilon\cos(ky) +\mathcal{O}(\varepsilon^2)
\label{perturb2-ssn}
\end{equation}
which confirms by the negative sign of the frequency response factor
$-k^2/2$ that the perturbation is smoothed out by the self-snakes process.

Pre-smoothing here leads to
\begin{equation}
g(x,y)=\frac12\,\left(1-\frac{k^2\varepsilon^2}2\exp(-k^2\sigma^2)\sin^2(ky)
\right)+\mathcal{O}(\varepsilon^3)\;,
\end{equation}
which in the further course of the calculation only influences higher-order
terms, such that \eqref{perturb2-ssn} is replicated.

\emph{Remark on explicit time discretisations.}
A difference to the first test case to be noted here is that the
negative amplification factor does not depend on $\sigma$. This
implies a time step size limit for explicit time discretisations of
pre-smoothed self-snakes: With $k$ denoting the highest perturbation
frequency that can occur in the discretised image, given by the
Nyquist frequency of the grid ($k=\pi$ for spatial mesh size $h=1$),
the amplification factor $\lambda_{\mathrm{s}}(k):=1-\tau\, k^2/2$
within a single time step of size $\tau$ must not become $-1$
or lower, thus $\tau<4/k^2$ must be observed.

\paragraph{Amoeba Filter Analysis.}

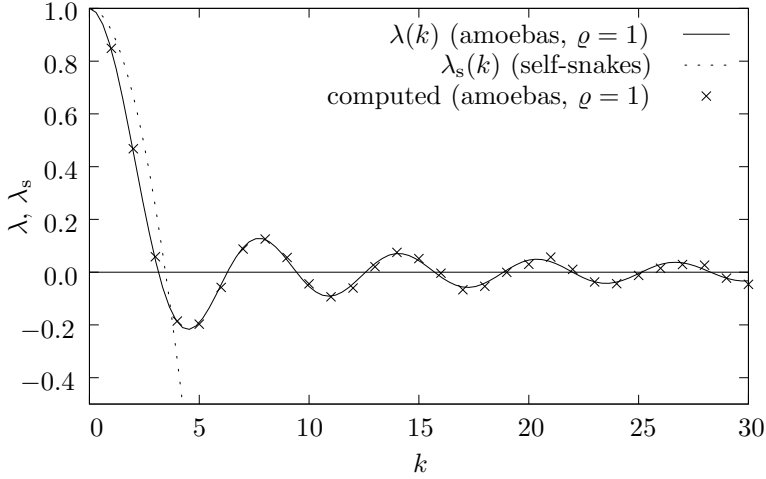
\begin{figure}[t!]
\unitlength0.001\textwidth
\begin{picture}(635,402)
\put( 67, 56){\includegraphics[width=560\unitlength]
      {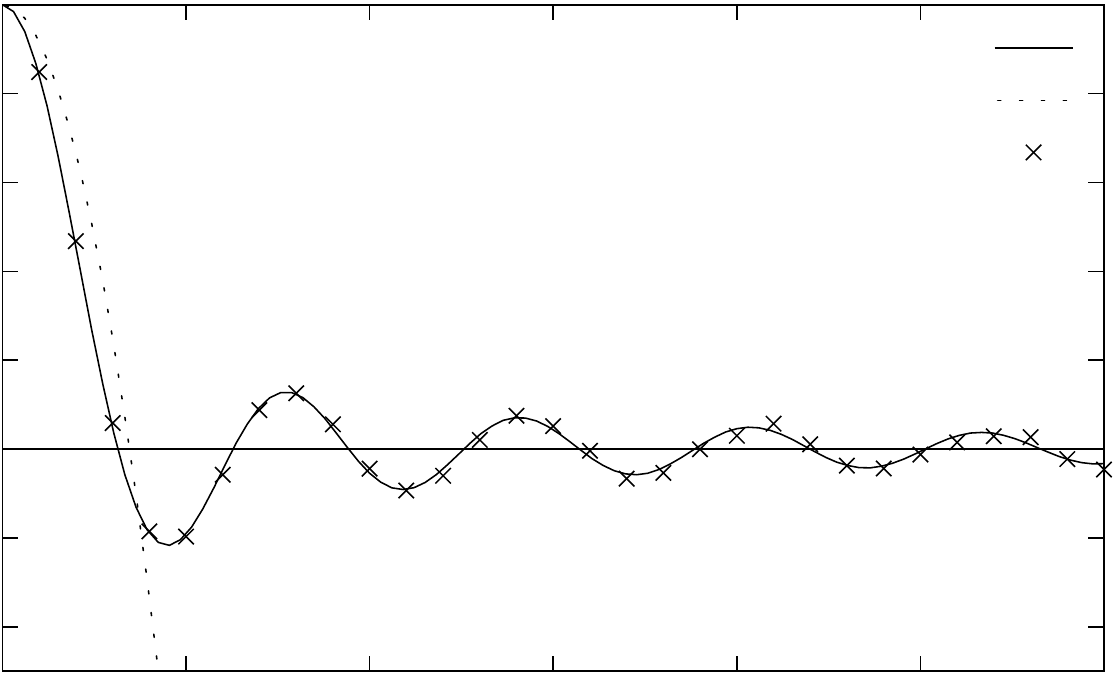}}
\put( 69, 30){\small$0$}
\put(155, 30){\small$5$}
\put(242, 30){\small$10$}
\put(335, 30){\small$15$}
\put(428, 30){\small$20$}
\put(522, 30){\small$25$}
\put(613, 30){\small$30$}
\put(340,  0){\small$k$}
\put( 10, 69){\small$-0.4$}
\put( 10,113){\small$-0.2$}
\put( 30,158){\small$0.0$}
\put( 30,203){\small$0.2$}
\put( 30,248){\small$0.4$}
\put( 30,293){\small$0.6$}
\put( 30,337){\small$0.8$}
\put( 30,382){\small$1.0$}
\put(  0,200){\rotatebox{90}{\small$\lambda$, $\lambda_{\mathrm{s}}$}}
\put(320,363){\small $\lambda(k)$ (amoebas, $\varrho=1$)}
\put(363,337){\small $\lambda_{\mathrm{s}}(k)$ (self-snakes)}
\put(268,311){\small computed (amoebas, $\varrho=1$)}
\end{picture}
\caption{\label{fig:perturb2plot}
Amplification of a level-line-aligned perturbation of a linear slope
function by one amoeba median filtering step (theoretical and
numerical values) and a corresponding time step of an explicit
scheme for self-snakes (with or without pre-smoothing).}
\end{figure}

To determine the response of an amoeba median filter step to the
perturbation \eqref{perturb2}, we consider again Euclidean amoeba
metric and $\beta=1$.
The image graph $\varGamma=\{(x,y,u(x,y))~|~(x,y)\in\bbbr^2\}$ of
\eqref{perturb2}, compare Section~\ref{subsec:amoebas-cont},
is a developable surface. The amoeba structuring element
$\mathcal{A}$ around $(x_0,y_0)$ then is the projection of a bent
Euclidean $\varrho$-disk $\mathcal{A}^*$
affixed to $\varGamma$ to the image plane, compare
Figure~\ref{fig:perturbschematic}(c).

The orthogonal projection $\mathcal{A}'$ of the same bent
$\varrho$-disk $\mathcal{A}^*$
not to the image plane but to the unperturbed image
graph $\varGamma_0=\{(x,y,x)~|~(x,y)\in\bbbr^2\}$,
compare Figure~\ref{fig:perturbschematic}(d),
is symmetric
w.r.t.\ the line $x=x_0':=x_0+\varepsilon\cos(kx_0)/\sqrt2$;
note that the point $(x_0,y_0,u(x_0,y_0))$ projects to
$(x_0',y_0,x_0')$.
Moreover, the projection from $\varGamma$ to $\varGamma_0$ changes
areas only by a factor $1+\mathcal{O}(\varepsilon^2)$. Similarly,
projection from $\varGamma$ to the image plane changes areas by a
factor $\sqrt2/2+\mathcal{O}(\varepsilon^2)$.

The amoeba median can therefore be computed up to
$\mathcal{O}(\varepsilon^2)$ from an area difference within
$\mathcal{A}'$ that solely results from the deviation of the
projected level line on $\varGamma$ from the line $x=x_0'$.

The level line of $u$ corresponding to $(x_0,y_0)$ is given by
$u(x,y)=u(x_0,y_0)$, thus
$x(y)=x_0+\varepsilon\cos(ky_0)-\varepsilon\cos(ky)$;
it projects on $\varGamma_0$ as
\begin{equation}
x(y) = 
x_0' +\frac12\bigl(\varepsilon\cos(ky_0)-\varepsilon\cos(ky)\bigr)
+\mathcal{O}(\varepsilon^2) \;.
\end{equation}
As the level line extends in $y$ direction from
$y_0-\varrho+\mathcal{O}(\varepsilon^2)$ to
$y_0+\varrho+\mathcal{O}(\varepsilon^2)$,
the resulting area difference on $\varGamma_0$ is compensated by a
level line shift of
\begin{align}
\varDelta x &= 
\frac{-2}{2\,\varrho}\int\limits_{y_0-\varrho}^{y_0+\varrho}
\frac{\varepsilon}2\bigl(\cos(ky_0)-\cos(ky)\bigr)\,\mathrm{d}y
+\mathcal{O}(\varepsilon^2)
\notag\\
&
= 
\left(\frac{\sin (k\varrho)}{k\varrho}-1\right) \varepsilon \cos(ky_0)
+\mathcal{O}(\varepsilon^2)
\;,
\end{align}
making $x_0+\varDelta x$ the sought median, and leading to a frequency
reponse factor $\delta(k):=\operatorname{sinc} (k\varrho)-1$ for the
increment of the perturbation.

As before, one amoeba median filter step changes the initial
perturbation $u-u_0$ of \eqref{perturb2} versus $u_0(x,y)=x$ by the
amplification factor $\lambda(k)=1+\delta(k)$, i.e.\
$\lambda(k)=\operatorname{sinc} (k\varrho)$.
Since $\lambda(k)$ is within $(-1,1)$ for all $k>0$, perturbations
of all frequencies are dampened.

Figure~\ref{fig:perturb2plot} shows the graphs of both amplification
functions, $\lambda(k)$ for amoeba median filtering with $\varrho=1$,
and
$\lambda_{\mathrm{s}}(k)=1+1/6 \cdot (-k^2/2)$ for the
corresponding time step of \eqref{perturb2-ssn} with time step size
$\varrho^2/6=1/6$, along with numerically computed amplification factors
for amoeba median filtering with the same parameters for
$k=1,2,\ldots,30$.

\section{Amoebas and Texture}
\label{sec:texture}

\newcommand{\GwA}{G_w^{\mathrm{A}}}
\newcommand{\TwA}{T_w^{\mathrm{A}}}
\newcommand{\TuA}{T_u^{\mathrm{A}}}
\newcommand{\GwP}{G_w^{\mathrm{E}}}
\newcommand{\TwP}{T_w^{\mathrm{E}}}
\newcommand{\TuP}{T_u^{\mathrm{E}}}
\newcommand{\MDE}{\Bar{I}_{\mathrm{D}}^{\kern0.1em\mathrm{E}}}
\newcommand{\IDW}{I_{\mathrm{D}}^{\mathrm{W}}}
\newcommand{\IfP}{I_{f^P}}
\newcommand{\IfV}{I_{f^V}}

As mentioned before, Dijkstra's shortest path algorithm on the
neighbourhood graph $G_w(f)$ or a subgraph thereof is used to compute
amoeba structuring elements. Whereas in image filtering, only the resulting
pixel set $\mathcal{A}_\varrho(i)$ around pixel $i$ is of interest,
the search tree created by Dijkstra's algorithm bears valuable information
in itself: its structure depends sensitively on the local structure of
contrasts in the image, thus, on its texture.
Building on work first presented in \cite{Welk-qgt14}, this section
discusses an approach directed at exploiting this information for texture
analysis.

\subsection{Six Graph Structures for Local Texture Analysis}
\label{subsec:sixgraphs}

\begin{figure}[t]
\unitlength0.011\textwidth
\begin{picture}(91,66)
\put(0,40){\rotatebox{90}{Amoeba}}
\put(0,5){\rotatebox{90}{Non-adaptive patch}}
\put( 4,63){Full edge-weighted graph}
\put(39,63){Dijkstra search tree,}\put(39,60){weighted}
\put(69,63){Dijkstra search tree,}\put(69,60){unweighted}
\put(11,33){\includegraphics[height=24\unitlength]{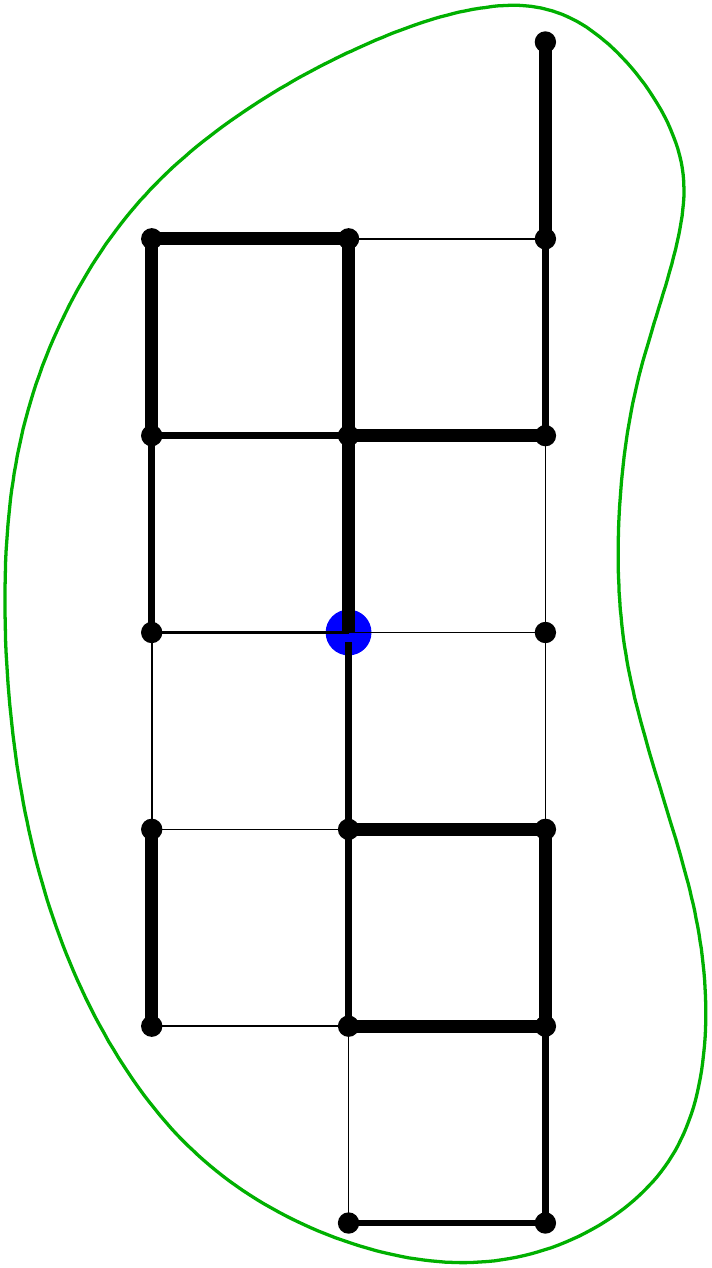}}
\put(27,33){\small$\GwA$}
\put(41,33){\includegraphics[height=24\unitlength]{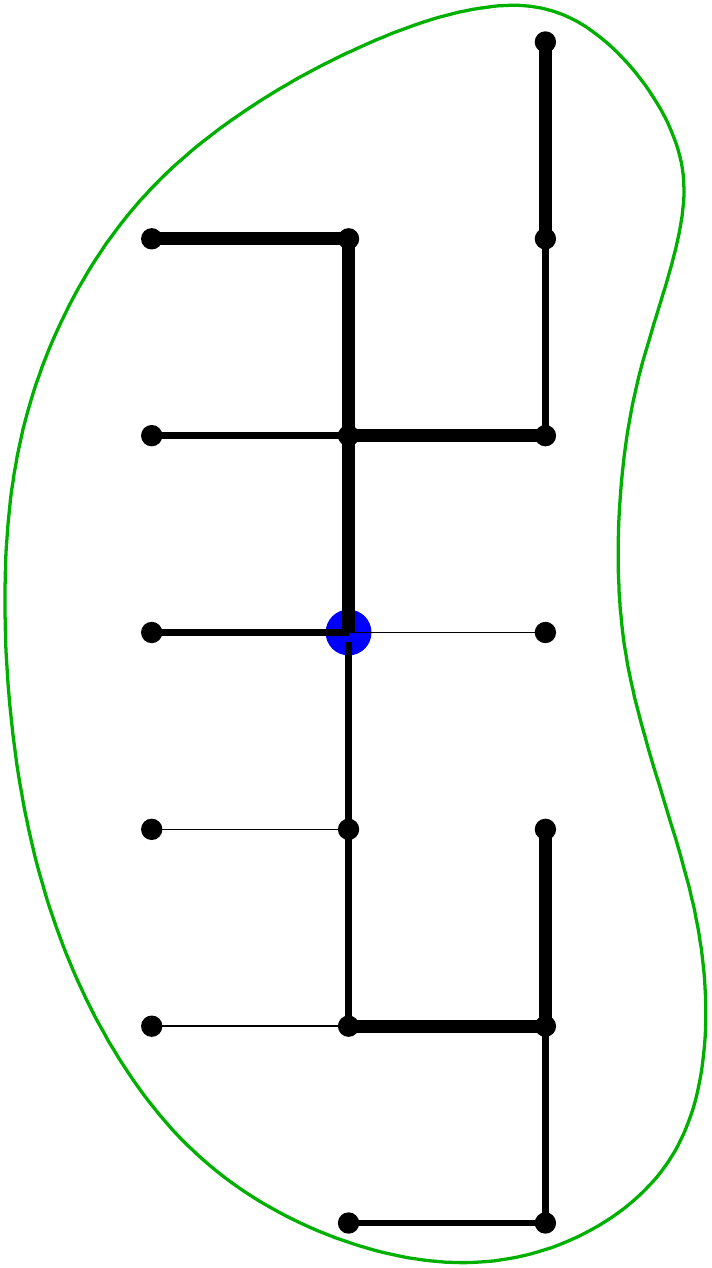}}
\put(57,33){\small$\TwA$}
\put(71,33){\includegraphics[height=24\unitlength]{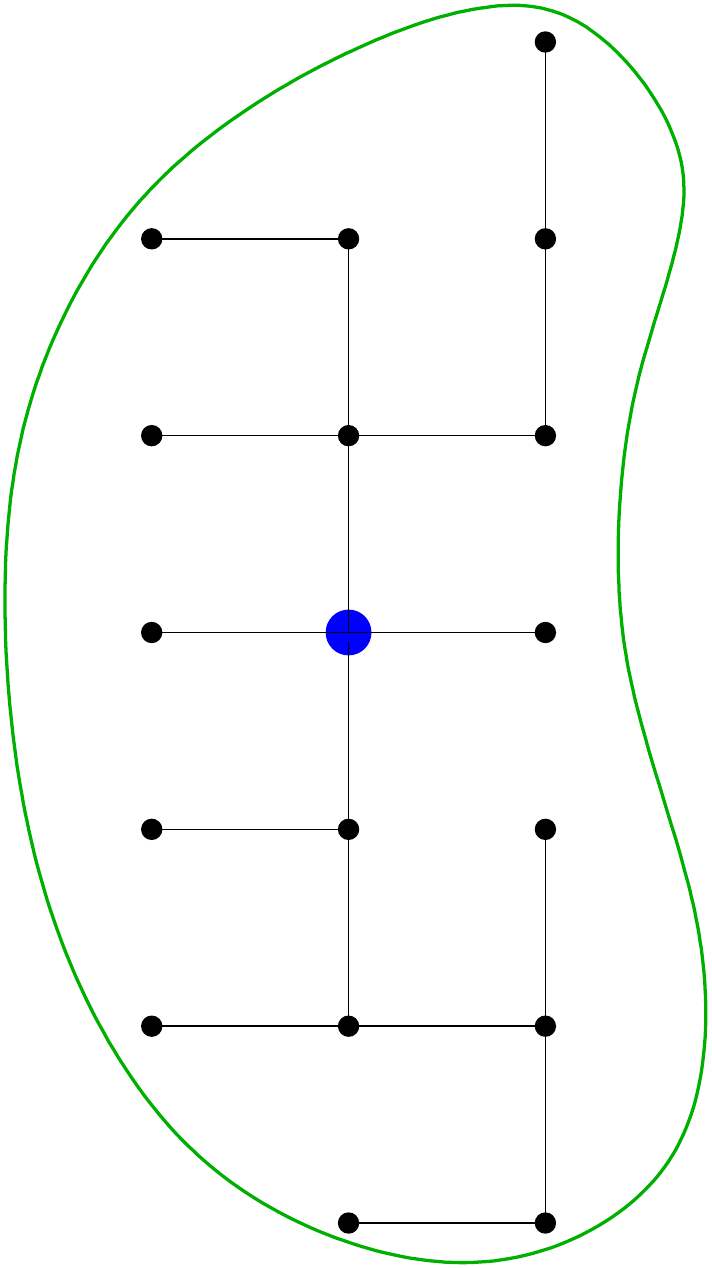}}
\put(87,33){\small$\TuA$}
\put( 7,3){\includegraphics[height=24\unitlength]{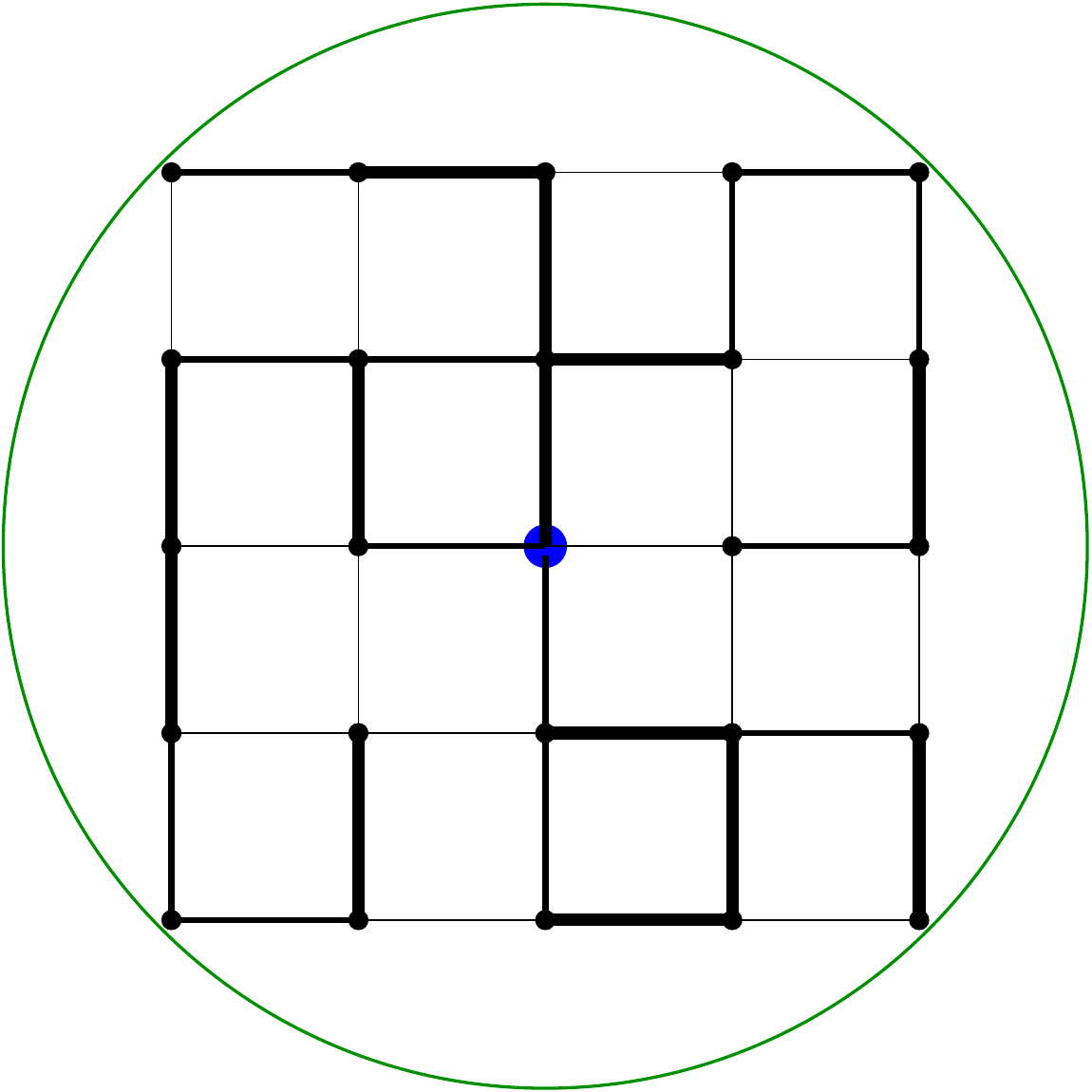}}
\put(27,2){\small$\GwP$}
\put(37,3){\includegraphics[height=24\unitlength]{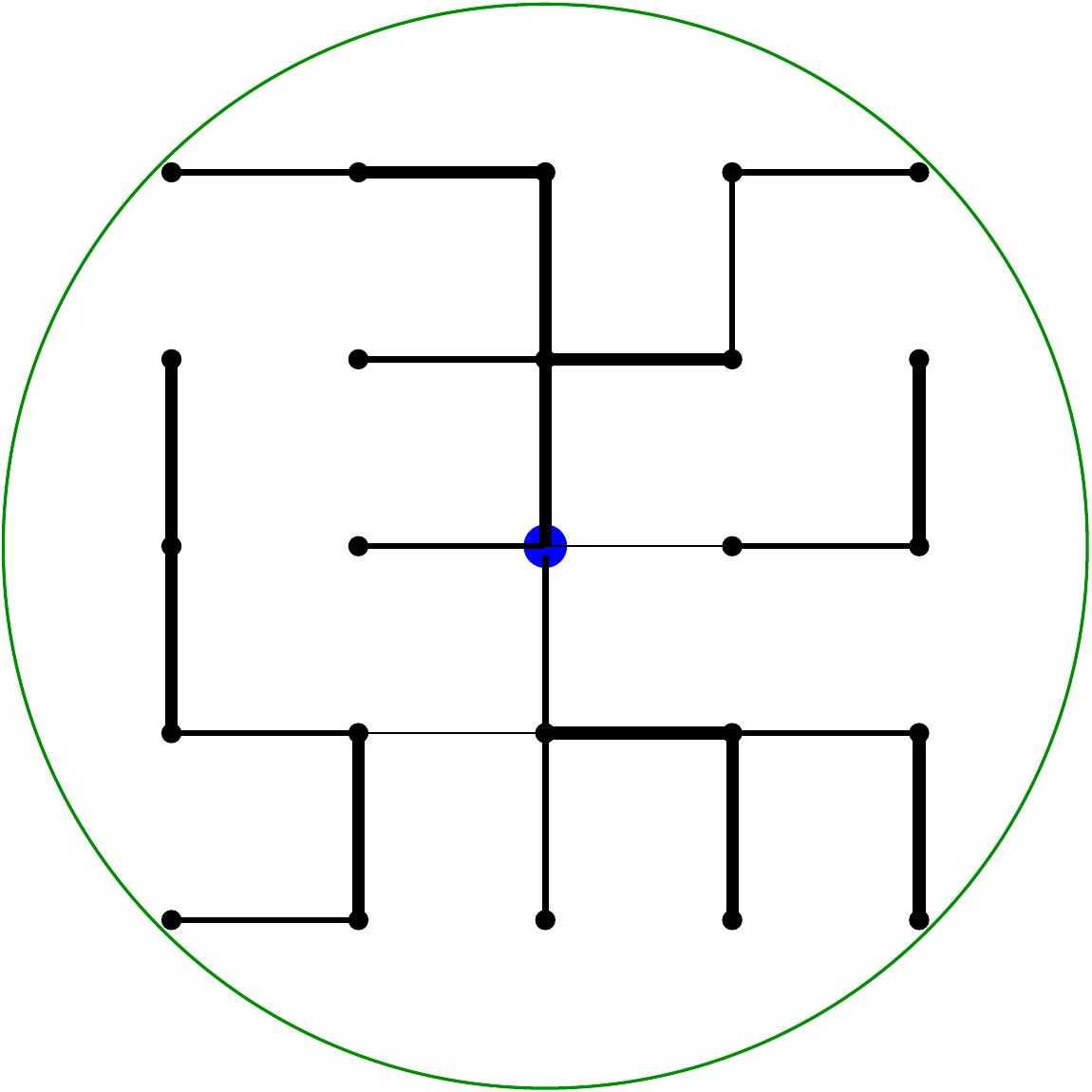}}
\put(57,2){\small$\TwP$}
\put(67,3){\includegraphics[height=24\unitlength]{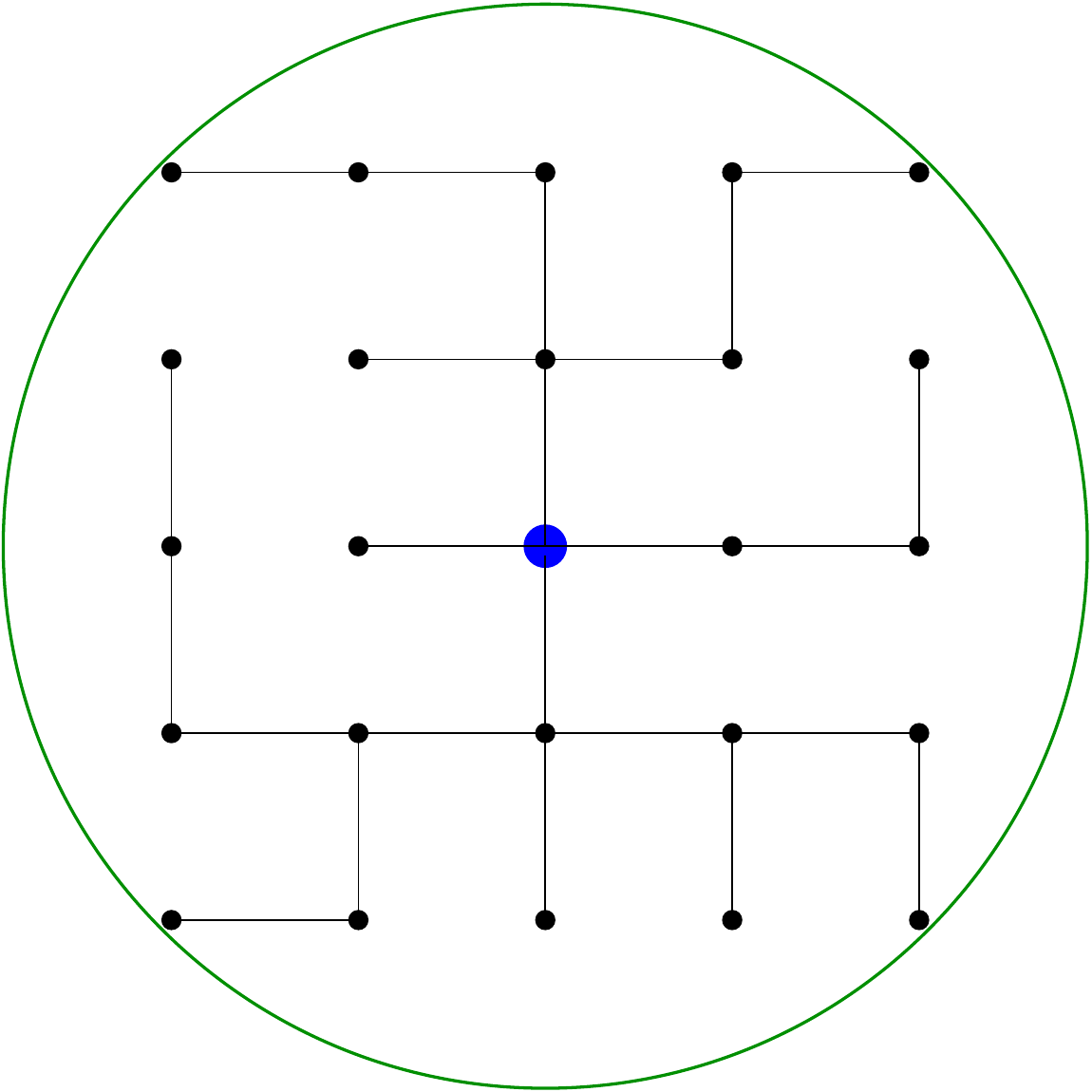}}
\put(87,2){\small$\TuP$}
\end{picture}
\caption{\label{fig:sixgraphs}Six graph setups for texture feature
construction from amoebas (schematic). For simplicity, graphs are
drawn based on 4-neighbourhood connectivity here. In the weighted graphs,
different line thicknesses symbolise edge weights.}
\end{figure}
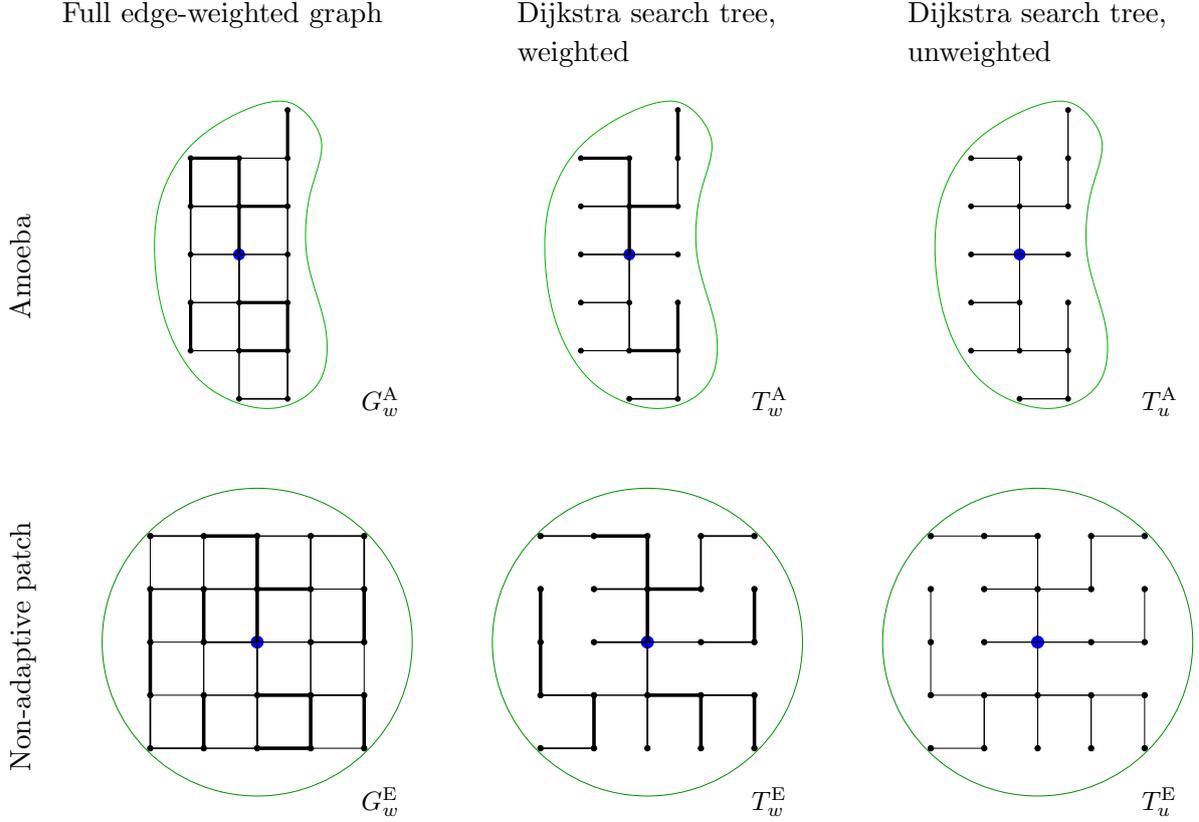

Looking at the amoeba construction in more detail, information about
local image texture is distributed to several features. The first aspect are
the amoeba distances between adjacent pixels themselves, i.e.\ the edge
weights of $G_w(f)$. A second source of information is the selected pixel
set of the amoeba $\mathcal{A}_\varrho(i)$. The third one is the connectivity
of the Dijkstra search tree.
This leads to six setups for graphs that encode these information
cues in different combinations.
Figure~\ref{fig:sixgraphs} illustrates these setups.

For the first group of three graphs, the pixels within
$\mathcal{A}_\varrho(i)$ serve as vertices. For these, one can consider
either the full weighted subgraph of $G_w(f)$, which will be denoted
by $\GwA$, the superscript $\mathrm{A}$ referring to the use of the amoeba
patch. Next, one can consider just the weighted Dijkstra tree, $\TwA$.
Third, deleting the edge weights from this tree yields an unweighted tree,
$\TuA$. Despite suspending the direct use of edge weights in this setting,
the connectivity structure derived thereof remains present.

The second group of three graphs is analogous to the first one but chooses
the pixels of a fixed window of Euclidean radius $\varrho$ around pixel $i$.
Again, one has the corresponding weighted subgraph of $G_w(f)$, which will
be denoted as $\GwP$, with the superscript $\mathrm{E}$ referring to the
Euclidean patch, the weighted Dijkstra tree $\TwP$ and the unweighted
Dijkstra tree $\TuP$.

\subsection{Quantitative Graph Theory: Graph Indices}
\label{subsec:gi}

We turn now to introduce exemplary graph descriptors that can be computed
from the previously mentioned graphs in order to obtain quantitative
texture descriptors. A larger set of graph descriptors is discussed
in the same context in \cite{Welk-qgt14}.

These graph descriptors are just samples from a
tremendous variety of more than 900 concepts \cite{Dehmer-PLO13}
that have been established over almost 70 years of research,
motivated from applications like the analysis of
molecule connectivity in computational chemistry, see e.g.\
\cite{Bonchev-JCP77,Hosoya-BCSJ71,Ivanciuc-JMC93,Plavsic-JMC93,Wiener-JACS47},
inexact graph matching \cite{Ferrer-bookchapter12,Sanfeliu-TSMC83},
or the quantitative analysis of (for instance, metabolistic)
networks, see e.g.\ \cite{Dehmer-AMC08,EmmertStreib-AMC07}.
In the recent decade, the systematic study of these measures has been
bundled in the field of \emph{quantitative graph theory}, see e.g.\
\cite{Dehmer-Book12,Dehmer-PLO12,EmmertStreib-AMC07}.

\subsubsection{Distance-Based Indices}

The historically first class of graph indices are computed directly from
the vertex distances within a graph.

Originally introduced for unweighted graphs $G$, the Wiener index
\cite{Wiener-JACS47} is obtained by just summing up the distances (path
lengths) between all pairs $\{i,j\}$ of vertices,
\begin{equation}
\label{gi-wiener}
W(G) := \sum_{\{i,j\}}d(i,j)\;.
\end{equation}
A modification is the Harary index introduced by Plav{\v s}i{\'c} et al.\
\cite{Plavsic-JMC93} that sums the reciprocals instead of
the distances themselves,
\begin{equation}
H(G) := \sum_{\{i,j\}}\frac1{d(i,j)}\;.
\end{equation}
It is straightforward to apply both indices also for weighted graphs,
replacing path lengths as distances by total path weights just as in
the amoeba definition.

\subsubsection{Information-Theoretic Indices}

Another important class of graph indices is based on entropy concepts.
Since Shannon's work \cite{Shannon-BSTJ48}, the entropy
\begin{equation}
H(p) := -\sum_{k=1}^np(k)\log_2 p(k)
\end{equation}
has been established as the fundamental measure of the information
content of a discrete probability measure $p$ on $\{1,\ldots,n\}$.

\paragraph{Bonchev-Trinajsti{\'c} Information Indices}

In \cite{Bonchev-JCP77}, entropy has been applied in several ways to
the distribution of distances within unweighted graphs to characterise
graph connectivity. We pick here two of them. We consider
a graph $G$ with vertices $1,\ldots,n$ and denote by $D(G)$ its diameter,
i.e.\ the largest path distance between two of its vertices. By $k_d$
we denote for $d=1,\ldots, D(G)$ the number of vertex pairs of exact
distance $d$,
\begin{equation}
k_d := \# \{ (i, j) ~|~ 1\le i<j\le n,~~d(i,j)=d \} \;.
\end{equation}
In \cite{Bonchev-JCP77}, the \emph{mean information on distances} $\MDE$
and the \emph{total information on the realised distances} $\IDW$ of $G$ are
defined, which (with a slight rewrite for $\IDW$) read as
\begin{align}
\MDE (G) &:= -\sum\limits_{d=1}^{D(G)} \frac{k_d}{\binom{n}{2}}
\log_2\frac{k_d}{\binom{n}{2}}\;,\\
\IDW (G) &:= W(G)\log_2 W(G) - \sum\limits_{1\le i<j\le n} 
d(i,j)\log_2 d(i,j)\;,
\end{align}
where $W(G)$ is the Wiener index \eqref{gi-wiener}.
Again, both definitions can formally be applied to weighted graphs by
performing the summation over the weighted path lengths $d$ occurring in $G$;
however, in non-degenerate cases all $k_d$ will equal $1$, turning the
mean information on distances $\MDE$ into a quantity that depends essentially
only on $n$, and does therefore not reveal much information about the graph.
In our texture analysis framework, $\MDE$ makes therefore sense only for
the unweighted graphs $\TuA$ and $\TuP$. In contrast, the total information
measure $\IDW$ makes perfect sense for weighted graphs and thus for all
six graph setups under consideration.

\paragraph{Dehmer Entropies}

While the Bonchev-Trinajsti{\'c} indices are based on entropies on the
set of distances in a graph, a class of entropy indices defined in
\cite{Dehmer-AMC08} works with distributions on the vertex set.
An arbitrary positive-valued function $f$ \emph{(information functional)}
on the vertices $1,\ldots,n$ of a graph $G$ is converted into a probability
density by normalising the sum of all values to $1$, such that the
individual probabilities $p(i)$ read as
\begin{equation}
p(i) := \frac {f(i)}{\sum_{j=1}^n f(j)}\;.
\end{equation}
The entropy
\begin{equation}
I_f(G) := H(p)
\end{equation}
is then a graph index based on the information functional $f$.

In \cite{Dehmer-AMC08}, two choices for $f$ have been considered
in the case of unweighted graphs, named $f^V$ and $f^P$.
For each of them, $f(i)$ is obtained from considering the set of
neighbourhoods of increasing radius around vertex $i$ in the path metric
of the graph. While $f^V(i)$ is the exponential of a weighted sum over
the cardinalities of such neighbourhoods, $f^P(i)$ is the exponential
of a weighted sum over the distance sums within these neighbourhoods
(i.e.\ the Wiener indices of the corresponding subgraphs). The weight
factors assigned to increasing neighbourhoods in both $f^P$ and $f^V$
can be chosen in different ways. Using what is called
\emph{exponential weighting scheme} in \cite{Dehmer-PLO12} and
measuring distances $d$ by total edge weights along paths
in edge-weighted graphs,
the resulting
information functionals can be stated as
\begin{align}
f^V(i) &:= \exp \left( M \sum_{j=1}^n q^{d(i,j)}\right)\;,\\
f^P(i) &:= \exp \left( M \sum_{j=1}^n q^{d(i,j)}d(i,j)\right)
\end{align}
with parameters $M>0$ and $q\in(0,1)$, see \cite{Welk-qgt14}
where it is also detailed how these expressions are derived from the
original definitions from \cite{Dehmer-AMC08}.

For the resulting entropy indices $I_{f^P}$ and $I_{f^V}$ as well as
for a third one, $I_{f^{\Delta}}$, which is not discussed here,
\cite{Dehmer-PLO12} demonstrated excellent discriminative power for
unweighted graphs, i.e.\ they are able to uniquely distinguish large
sets of different unweighted graphs. This finding lets appear
$I_{f^P}$ and $I_{f^V}$ also as outstanding candidates for texture analysis
tasks.

\subsection{Texture Discrimination}

As a first, yet simple, application of the framework that combines amoebas
and graph indices, texture discrimination is considered. In
\cite{Welk-qgt14}, a total of 42 candidate texture descriptors was
considered. These descriptors resulted from applying nine graph indices,
including those described in Section~\ref{subsec:gi} above,
to the six graph setups introduced in Section~\ref{subsec:sixgraphs},
using only those combinations that made sense (as e.g.\ some graph indices
cannot be used for weighted graphs). These graph indices were compared
to Haralick features \cite{Haralick-PIEEE79,Haralick-TSMC73}, a set of
region-based texture descriptors derived from several statistics
of co-occurrence matrices of intensities.
Despite their long history of more than fourty years, Haralick features
are still prominent in texture analysis; together with some more recent
modifications they continue to yield competitive results
\cite{Howarth-civr04,Huang-isbi04,Tesar-CMIG08}.

For the texture discrimination task, the experimental setup in
\cite{Welk-qgt14} was built to suit the region-based Haralick features
by aggregating the, actually local, amoeba-graph features regionwise.

\begin{figure}[t]
\includegraphics[width=0.5\textwidth]{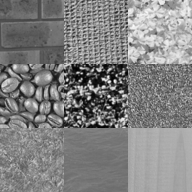}
\caption{\label{fig:vistex}Composite image containing patches of
nine different textures; top left to bottom right in rows:
\emph{brick, fabric, flowers, food, leaves, metal, stone, water, 
wood.}
Texture patches originate from the
\emph{VisTex} database, \cite{vistex}; they have been converted
to greyscale, downsampled and clipped. VisTex database
\copyright 1995 Massachusetts Institute of Technology.
Developed by Rosalind Picard, Chris Graczyk, Steve Mann, Josh Wachman,
Len Picard, and Lee Campbell at the Media Laboratory, MIT, Cambridge,
Massachusetts.}
\end{figure}

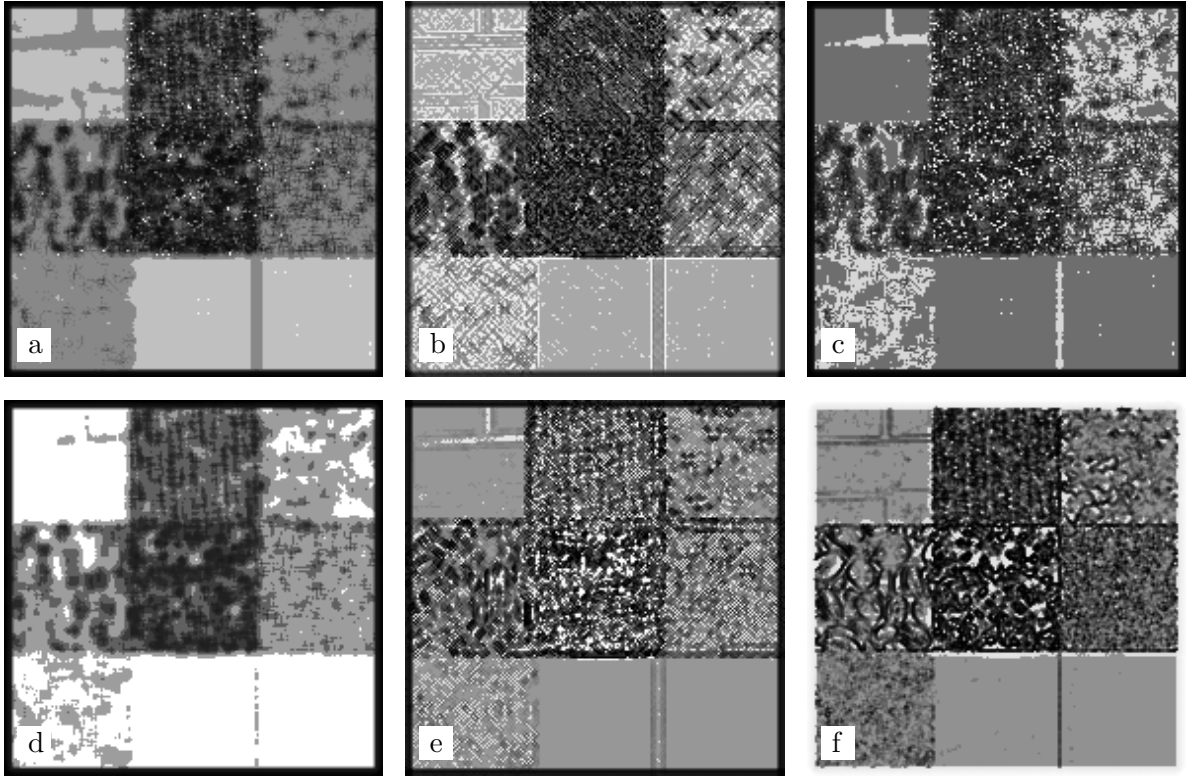
\begin{figure}[t]
\unitlength0.01\textwidth
\begin{picture}(100,66)
\put(0,34){\includegraphics[width=32\unitlength]
    {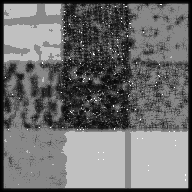}}
\put(34,34){\includegraphics[width=32\unitlength]
    {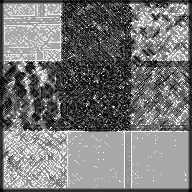}}
\put(68,34){\includegraphics[width=32\unitlength]
    {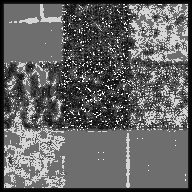}}
\put(0,0){\includegraphics[width=32\unitlength]
    {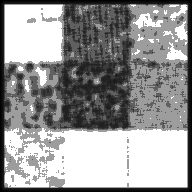}}
\put(34,0){\includegraphics[width=32\unitlength]
    {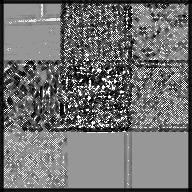}}
\put(68,0){\includegraphics[width=32\unitlength]
    {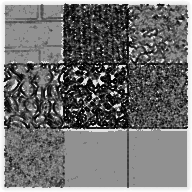}}
\put( 1.1,36){\colorbox{white}{\rule{0pt}{.6em}\hbox to.6em{\kern.1em\smash{a}}}}
\put(35.1,36){\colorbox{white}{\rule{0pt}{.6em}\hbox to.6em{\kern.1em\smash{b}}}}
\put(69.1,36){\colorbox{white}{\rule{0pt}{.6em}\hbox to.6em{\kern.1em\smash{c}}}}
\put( 1.1, 2){\colorbox{white}{\rule{0pt}{.6em}\hbox to.6em{\kern.1em\smash{d}}}}
\put(35.1, 2){\colorbox{white}{\rule{0pt}{.6em}\hbox to.6em{\kern.1em\smash{e}}}}
\put(69.1, 2){\colorbox{white}{\rule{0pt}{.6em}\hbox to.6em{\kern.1em\smash{f}}}}
\end{picture}
\caption{\label{fig:gi}Examples of graph-index-based feature
descriptors computed on the test image shown in
Figure~\ref{fig:vistex}. Graph indices have been computed from
amoebas with Euclidean amoeba metric, $\beta=0.1$ and $\varrho=5$.
All graph index images shown here are histogram equalised.
\textbf{(a)} Harary index on the weighted amoeba tree $\TwA$. --
\textbf{(b)} Dehmer entropy $\IfP$ on $\TwA$. --
\textbf{(c)} $\IDW$ on $\TwA$. --
\textbf{(d)} Harary index on the weighted tree in the Euclidean
neighbourhood $\TwP$. --
\textbf{(e)} Dehmer entropy $\IfP$ on $\TwP$. --
\textbf{(f)} Dehmer entropy $\IfV$ on $\TwP$.
}
\end{figure}

Amoeba-graph descriptors as well as Haralick features were computed
for a set of nine texture images from the \emph{VisTex} database,
\cite{vistex}.
Figure~\ref{fig:vistex} shows a
composite image made up of the nine textures used in \cite{Welk-qgt14}.
Figure~\ref{fig:gi} visualises selected amoeba-graph features on this
test image.
It can be seen that the different features respond with different degrees
of sensitivity and locality to the local structure of the textures.

For each descriptor and texture pair, a statistical
discrepancy measure $u:=\lvert\mu_1-\mu_2\rvert/\sigma$ was computed
from the mean values $\mu_1$, $\mu_2$ of the texture descriptor on both
textures and the joint standard deviation $\sigma$. Due to the variability
of each descriptor even within the same texture, thresholds for
discrimination were gauged from the measured discrepancies for different
patches of the same textures:
A higher threshold, $T_1$, was chosen as
double the maximum of the nine intra-texture discrepancies measured, and
a lower threshold, $T_2$, as the third-highest of the nine intra-texture
values. Texture pairs with discrepancy at least $T_1$ were considered
as ``certainly different'', and those with discrepancy at least $T_2$ as
``probably different''.

While not each texture descriptor could equally well distinguish each
pair of textures, it turns out that almost all texture pairs can be told
apart by at least some descriptors, with the overall discrimination capability
being well comparable with that achieved by the Haralick feature set under
consideration. Indeed, the pair \emph{water}/\emph{wood} (the last two
patches in the bottom row of Figure~\ref{fig:vistex} was the only one that
could not be distinguished with sufficient certainty, neither by the Haralick
nor the amoeba-graph feature set. The difficulty to distinguish these two
textures can also be seen in Figure~\ref{fig:gi}.

Given that different texture pairs are distinguished best with different
descriptors, it is of interest to study the
similarity and dissimilarity of different amoeba-graph texture descriptors
with regard to what texture pairs they can distinguish. In \cite{Welk-qgt14}
a metric on the set of texture descriptors has been established in this way.
In the further perspective, this is intended to
guide the selection of a subset of just a few descriptors that complement
each other well, which could therefore be a well-manageable feature set
for practical applications.

\subsection{Texture Segmentation}
\label{subsec:texseg}

Finally, we show a simple example that demonstrates the applicability of
amoeba-graph indices for texture segmentation. Here graph descriptors have
been used as input to a standard geodesic active contour method with an
outward force term $\gamma\,g\,\lvert\bm{\nabla}u\rvert$.

\begin{figure}[t]
\unitlength0.01\textwidth
\begin{picture}(100,100)
\put(0,68){%
\includegraphics[width=32\unitlength]{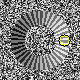}}
\put(34,68){%
\includegraphics[width=32\unitlength]{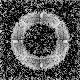}}
\put(68,68){%
\includegraphics[width=32\unitlength]{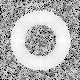}}
\put(0,34){%
\includegraphics[width=32\unitlength]{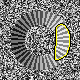}}
\put(34,34){%
\includegraphics[width=32\unitlength]{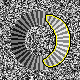}}
\put(68,34){%
\includegraphics[width=32\unitlength]{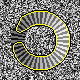}}
\put(0,0){%
\includegraphics[width=32\unitlength]{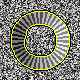}}
\put(34,0){%
\includegraphics[width=32\unitlength]{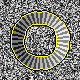}}
\put(68,0){%
\includegraphics[width=32\unitlength]{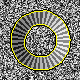}}
\put( 1.1,70){\colorbox{white}{\rule{0pt}{.6em}\hbox to.6em{\kern.1em\smash{a}}}}
\put(35.1,70){\colorbox{white}{\rule{0pt}{.6em}\hbox to.6em{\kern.1em\smash{b}}}}
\put(69.1,70){\colorbox{white}{\rule{0pt}{.6em}\hbox to.6em{\kern.1em\smash{c}}}}
\put( 1.1,36){\colorbox{white}{\rule{0pt}{.6em}\hbox to.6em{\kern.1em\smash{d}}}}
\put(35.1,36){\colorbox{white}{\rule{0pt}{.6em}\hbox to.6em{\kern.1em\smash{e}}}}
\put(69.1,36){\colorbox{white}{\rule{0pt}{.6em}\hbox to.6em{\kern.1em\smash{f}}}}
\put( 1.1, 2){\colorbox{white}{\rule{0pt}{.6em}\hbox to.6em{\kern.1em\smash{g}}}}
\put(35.1, 2){\colorbox{white}{\rule{0pt}{.6em}\hbox to.6em{\kern.1em\smash{h}}}}
\put(69.1, 2){\colorbox{white}{\rule{0pt}{.6em}\hbox to.6em{\kern.1em\smash{i}}}}
\end{picture}
\caption{\label{fig:texringseg}Texture segmentation by geodesic active contour
evolution based on amoeba/graph index texture features,
pre-smoothing $\sigma=3$, force term $\gamma=-2$, time step size $\tau=0.1$.
\textbf{(a)} Original image with initial contour. --
\textbf{(b)} Graph index $\IfP$ on weighted tree $\TwP$
(normalised from $[0,3.72]$ to $[0,255]$). --
\textbf{(c)} Graph index $\MDE$ on unweighted tree $\TuA$
(normalised from $[0,2.93]$ to $[0,255]$). --
\textbf{(d)} Contour after $500$ iterations
of GAC evolution using $\IfP$ on $\TwP$ and $\MDE$ on $\TuA$
each weighted $0.5$,
Perona-Malik threshold $\lambda=0.036$. --
\textbf{(e)} Same as (d) but $1000$ iterations. --
\textbf{(f)} Same as (d) but $2500$ iterations. --
\textbf{(g)} Steady state of the segmentation process from (d)--(f)
reached after $3300$ iterations. --
\textbf{(h)} Segmentation using only $\IfP$ on $\TwP$,
Perona-Malik threshold $0.48$,
steady state reached after $7500$ iterations. --
\textbf{(i)} Segmentation using only $\MDE$ on $\TuA$,
Perona-Malik threshold $0.4$,
steady state reached after $1200$ iterations.
}
\end{figure}

Figure~\ref{fig:texringseg}(a) shows a test image displaying a striped
ring in front of a noisy background. Figure~\ref{fig:texringseg}(b)
shows the field of graph indices $\IfP$ computed on weighted Dijkstra
trees in Euclidean patches, $\TwP$, while Figure~\ref{fig:texringseg}(c)
shows $\MDE$ on $\TuA$. It is evident from these examples that amoeba-graph
indices can turn the textured foreground
object into a more homogeneous region. Using just the two graph descriptors
as input channels for geodesic active contours one obtains a reasonable
segmentation, see Figure~\ref{fig:texringseg}(g).
One might ask whether one graph index alone does the job,
too. In the present example, this is indeed true; however, the results
in Figure~\ref{fig:texringseg}(h, i) are visibly less precise in locating the
contour separating foreground and background.

Note that this example is only a first proof of concept.
A deeper investigation of the potential of this approach to texture
segmentation as well as the study of parameter choice and comparison to
other texture segmentation methods are topics for future research.

\section{Outlook}

From the results reviewed in this paper it can be seen that
morphological amoebas provide a powerful framework for adaptive image
filtering with interesting cross-relations to other classes of filters.
They can also be applied fruitfully to related tasks such as image
segmentation. Combining amoeba procedures with ideas from quantitative graph
theory even allows to construct a new class of texture descriptors.

At the same time, there remain many questions for future research.
So far, the amoeba framework introduces adaptivity into local image filters
solely by modifying the first step of the filter procedure, i.e.\ the
selection stage. The aggregation step like median, maximum, or minimum is
left unchanged. Could further improvements of adaptivity be achieved by
envisioning also image-dependent modifications to the aggregation step?
How do modifications of selection and aggregation step interact?

Addressing the selection step itself, it would be possible to weaken
the dichotomy of including or not including neighbour locations, and to
consider unsharp or weighted neighbourhoods.

No amoeba filter for multi-channel (such as colour) images have been
studied in the present paper. In principle, there is little to prevent
one from applying amoeba procedures to multi-channel data. The amoeba
computation step generalises straightforwardly.
There are also generalisations of median filters
\cite{Austin-Met59,Spence-icip07,Weiszfeld-TMJ37,Welk-dagm03,Welk-SP07}
and supremum/infimum operations to multi-channel data
\cite{Burgeth-ismm13,Burgeth-SP07,Burgeth-eccv04} at hand.
The theoretical understanding of multi-channel amoeba filters, however,
lags behind that in the single-channel case.
A result in \cite{Welk-Aiep14} indicates that the median--PDE
relation even in its non-adaptive form, see Proposition~\ref{prop:imfpde},
has no equally simple multi-channel counterpart, thus
leaving little hope to derive manageable PDE equivalents of multi-channel
amoeba filters. New approaches to a deeper understanding of the properties
of multi-channel amoeba filters will have to be sought.

The field of texture analysis addressed in Section~\ref{sec:texture} still
is at an early stage of research. Ongoing research is directed at extending
the experimental evaluation of the newly introduced amoeba-graph texture
descriptors for texture discrimination to a broader body of data.
Another goal is the selection of a powerful set of a few amoeba-graph
descriptors with a high combined discrimination rate across multiple textures.
Tuning of the parameters of the descriptors has not been studied extensively
so far and will therefore be addressed in the future. Attempts are also
underway to analyse the effect of the amoeba-graph descriptors theoretically.

In the field of texture segmentation the combination of amoeba-graph
descriptors with other segmentation frameworks than the GAC considered in
Section~\ref{subsec:texseg} will be investigated. An integration with
an amoeba active contour procedure could lead to a texture segmentation
framework that uses the same sort of theoretically founded procedure for
both texture feature extraction and the actual segmentation step.
In many existing approaches, and also in the preliminary example from
Section~\ref{subsec:texseg}, these two steps are based on rather unrelated
approaches. With regard to the graph-theoretical roots of the texture
features under consideration, also graph-cut approaches for the segmentation
stage could be a candidate for further investigation.

\end{document}